\documentclass{dissertation}
\usepackage{subcaption}
\usepackage{graphicx}
\usepackage{float}
\usepackage{color, colortbl}
\usepackage{array}
\usepackage{xcolor}

\definecolor{DarkGrey}{rgb}{0.74,0.74,0.74}
\definecolor{LightGrey}{rgb}{0.87,0.87,0.87}
\definecolor{LightLightGrey}{rgb}{0.94,0.94,0.94}

\let\cleardoublepage\clearpage 

\newacronym{AI}{AI}{Artificial Intelligence}
\newacronym{GDL}{GDL}{Game Description Language}
\newacronym{GGP}{GGP}{General Game Playing}
\newacronym{GGS}{GGS}{General Game System}
\newacronym{GVGAI}{GVGAI}{General Video Game AI}
\newacronym{RBG}{RBG}{Regular Boardgames}

\usepackage{xspace}

\begin{document}

\frontmatter

\includepdf[pages=-]{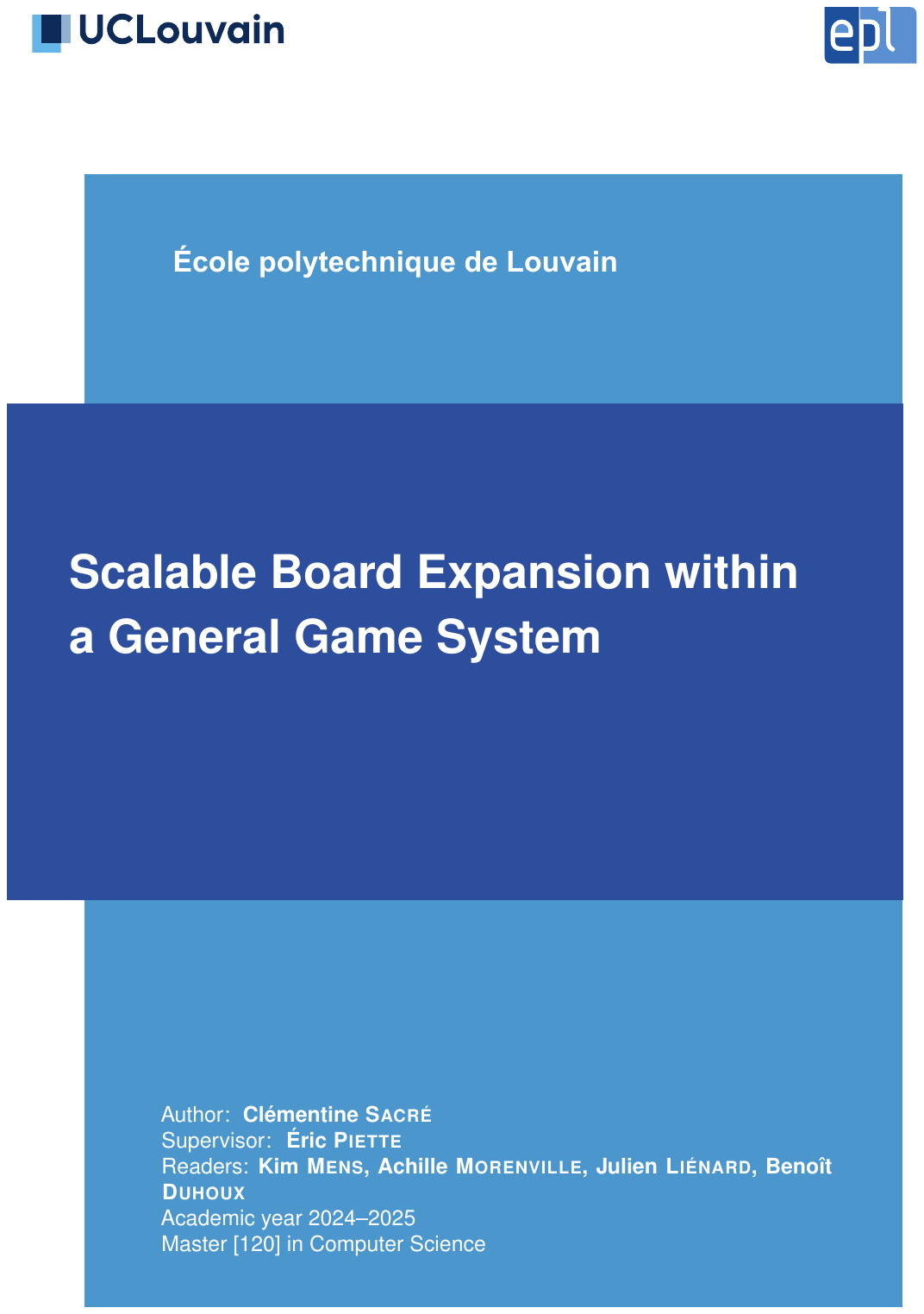}


\tableofcontents

\chapter*{Acknowledgments}
\addcontentsline{toc}{chapter}{Acknowledgments}
\setheader{Acknowledgments}

Writing this thesis has been a challenge and a very instructive journey at the same time, not only in terms of technical knowledge but also on a personal level, helping me push my own limits, and I would like to thank everyone who supported me throughout this experience.\\

First of all, I would like to sincerely thank my supervisor, Professor Eric Piette, for his support, advices, and availability throughout this project. His guidance has been incredibly helpful, and I truly appreciated his trust as well as the time he took to assist me along the way.\\


I am also thankful to Professor Kim Mens, Achille Morenville, Julien Liénard and Benoît Duhoux for accepting to be readers of this thesis.\\

Lastly, thanks to my family and friends for their support and motivation during this year, especially my dad.\\

The ChatGPT, Claude Sonnet, and Le Chat Mistral tools were used to assist me in translating, formulating and refining sentences, helping me maintain a formal yet pleasant tone throughout this report.

\mainmatter

\thumbtrue
\chapter{Introduction}
\label{introduction}

Unlike traditional board games, boardless tabletop games do not rely on a fixed playing surface. This absence of a fixed board allows for greater flexibility and creativity in the game. As a result, each game can evolve in many different ways, depending on the choices made by the players. Indeed, the course of the game is entirely shaped by individual decisions, guaranteeing a distinct experience at every session.\\

A well-known example of this type of game is \textit{Carcasonne}\footnote{\url{https://www.play-in.com/pdf/rules_games/carcassonne_jeu_de_base_-_nouvelle_edition_regles_fr.pdf}}. This tile-based strategy game calls on players to build a medieval landscape in the south of France. Players take turns drawing and placing tiles that depict cities, roads, monasteries and fields. Each square tile must be placed so that the elements on its edges align with the corresponding features on the adjoining tiles. On their turn, players may also place pawns on the tiles, to claim the different types of territory; these pawns score points when the territories they occupy are completed, or when they dominate specific areas. The game ends once all tiles have been placed, at which point points are scored according to the areas completed and the number of pawns covering them. The player with the highest score wins the game. \\
More elementary games, such as \textit{Andantino}, \textit{Trax} or \textit{Ringo}, also illustrate this boardless concept. \\

The application of Artificial Intelligence (AI) techniques to boardless tabletop games involves the development of algorithms capable of navigating these complex, dynamic environments without fixed structures. These AI systems engage in abstract decision-making and strategy formulation, often employing reinforcement learning to refine strategies based on player interactions. Such AI enhances the gaming experience by generating responsive and unpredictable scenarios, offering intelligent opposition in strategy-based games without predefined paths.\\

Current AI applications in this field are exemplified by the Ludii System, recognized as the forefront General Game Playing system in academic circles. However, the representation of state in these boardless games within Ludii is not yet optimal, reducing the efficiency of random play simulations and resulting in less proficient AI agents.\\

This master thesis aims to propose a novel, optimal, and scalable board expansion mechanism for the Ludii system. This new model will dynamically adjust the size of the board representation to minimize memory and computational resource usage for boardless games. In addition to these goals, it also seeks to provide a correct and coherent representation of boardless games.\\

An initial review will present a state-of-the-art analysis of existing approaches for representing board games in general, highlighting their respective strengths and limitations, as several other systems other than Ludii have also been developed for this purpose. This will be followed by an examination of Ludii’s implementation, with a particular focus on its current constraints regarding boardless games. This analysis will serve to introduce the main objective of this thesis.\\
Afterwards, a dedicated section will explore the internal structures of the Ludii system that are essential for understanding its concept of a game board, and how these can be extended to support dynamic boards, for boardless games. The next section will detail the improvements developed during this thesis, introducing a new mechanism for enabling scalable board expansion within Ludii.\\
To evaluate the effectiveness of these contributions, a series of experiments will be presented, demonstrating the benefits and performance of the proposed solutions. This will be followed by a discussion of the results and potential directions for future work. Finally, the thesis will conclude with a summary of key findings of this research.

\chapter{General Game Playing} 
\label{boardless}

For many years, games have served as valuable benchmarks in the field of AI research. They provide fast, cheap, and controlled environments that simulate aspects of the real world. As research in this domain progressed, games naturally became a means for comparing the performance of different AI systems. This, in turn, led to the creation of standardized competitions, fostering innovation and collaboration within the AI research community. Researchers began to explore not only how AI could master individual games, but whether it could generalize its reasoning across a variety of them~\cite{gvgaibook2019}.\\
This line of inquiry gave rise to the concept of General Game Playing.

\section{General Game Playing}
General Game Playing (GGP) is an area of AI which aims to create agents capable of playing a wide variety of games without being specifically programmed for each one. Unlike traditional game AIs, which are tailored to a single game such as \textit{Chess} or \textit{Go}, GGP agents must be able to interpret the rules of a new game at runtime and learn strategies to play it competently~\cite{pitrat1968realization}~\cite{genesereth2014general-ch1}. This makes GGP a powerful research area for studying general intelligence, learning, and decision-making under uncertainty.\\


The GGP paradigm is fundamentally characterized by three core pillars:
\begin{enumerate}
    \item \textbf{Generality}\\
    A defining feature of GGP systems is their ability to play a wide variety of games, ranging from simple to highly complex, regardless of the number of players, game mode (turn-based or simultaneous), or whether the player is in possession of all the information or part of it is hidden from him. This generality is what sets GGP apart from traditional game-specific AI.

    \item \textbf{Formal Game Structure}\\
    Games are formally specified using a Game Description Language (GDL), which ensures consistency and precision in rule representation. The language is designed to be both unambiguous and comprehensive, enabling machines to parse and understand the rules without human intervention.

    \item \textbf{Time Constraints}\\
    GGP agents must make decisions under strict time constraints. This requirement places significant demands on the efficiency of the reasoning and planning algorithms used.
\end{enumerate}

\section{General Game Systems}
General Game Systems (GGS) refer to platforms or software architectures designed to create, simulate, and analyze various types of games. Their purpose goes beyond simply playing games, as they are also used to study theoretical properties or game mechanics. Several such systems exist, and the most well-known will be detailed below~\cite{piette2019ludii}.\\

While GGS are primarily designed to model, generate, and analyze a wide range of games, GGP agents are intelligent agents developed to play such games without prior knowledge of their rules. In this context, a GGS can serve as an environment in which a GGP agent operates. The GGS defines and structures the games, often through a formal language, while the GGP involves intelligents agents that interpret this definition to reason, plan, and make decisions. Together, they form a complementary ecosystem for evaluating general intelligence in games and for exploring the design and dynamics of diverse game types.

\section{Overview of GGP Systems}
Over the years, several platforms and frameworks have been developed to support research in GGP. While they share the common goal of enabling AI agents to play a wide range of games without prior hardcoding, these systems differ in terms of design philosophy, supported game types, underlying languages, and target applications. The following subsections provide an overview of some of the most notable GGP platforms, outlining their key characteristics and roles.

\subsection{Zillions of games}
Zillions of Games~\cite{zillions-of-games-2} represents the first commercial application of GGP technology, launched in 1998. This pioneering system demonstrated the commercial viability of the GGP concept even before the term was widely recognized in academic research.\\

Developed by Jeff Mallett and Mark Lefler, the system enabled users to play a wide range of abstract strategy board games and puzzles defined in a specialized language. Zillions of Games paved the way for future GGP systems by showing that a single game engine could interpret the rules of diverse games and provide a competent AI opponent for human players. Zillions of Games has the potential to play a very large number of user-programmed games, as it comes with over 300 games and puzzles~\cite{enwiki:1236294678}.

\subsubsection{Limitations}
Despite its flexibility, Zillions of Games exhibits several significant limitations~\cite{zillions-of-games}:
\begin{itemize}
    \item Zillions of Games is exclusively tailored for games of perfect information, which severely restricts its applicability to games involving hidden or imperfect information, such as card games or modern board games. Instead, the program leverages complete knowledge of the game state, including opponents' cards and hidden elements, thus compromising fair play. While humans can still play these games manually, the system offers no mechanism for selectively revealing information to individual players.
    \item The scripting language used in Zillions of Games is highly constrained: it does not support arithmetic operations, functions, and variables beyond a limited set of Boolean flags. These limitations make it not really suited for implementing boardless games, which often require more complex logic.
    \item It also does not allow for multiple pieces to occupy the same location, nor can a single piece span multiple board positions. This limitation complicates the implementation of certain games, such as \textit{Ringo}, which involves stacked pieces.
    \item Lastly, the default engine of Zillions of Games struggles with games that exhibit a high branching factor. This limitation becomes especially problematic in the context of boardless games, as the branching factor increases significantly, as each new tile can potentially be placed in numerous locations and orientations, resulting in an exponential growth of possible game states.
\end{itemize}

\subsection{General Video Game AI}
General Video Game AI (GVGAI)~\cite{gvgaibook2019}, an extension of the GGP concept, is an open platform for testing AIs in general video games. It allows the creation of agents capable of playing different types of video games without requiring specific knowledge of each game. It is a flexible environment aimed at evaluating AI in a variety of games, which is part of a broader approach to AI in video games. Thus, GVGAI can be considered as a specific implementation or application within the larger field of GGP.\\

What makes GVGAI particularly interesting is its competitive approach. GVGAI competitions are regularly organized, where participants submit their AI controllers, which are then evaluated on an unknown set of games. These competitions stimulate innovation in the field, with various algorithmic approaches competing against each other~\cite{perez2016general}.

\subsubsection{Limitations}
However, in GVGAI, the game board (or environment) is generally generated at the beginning of the game and remains static or evolves according to fixed rules, but it is not constructed dynamically during the game, as in \textit{Dorfromantik} or other boardless games.

\subsection{GGP-Base}
One of the most widely adopted platforms for GGP research has been GGP-Base, which served as the standard for academic studies in the field from its introduction in 2005 until around 2016 – 2017~\cite{piette2019ludii}.\\

At the core of GGP-Base is the S-GDL (Stanford Game Description Language), a formal language based on first-order logic. Games described in S-GDL consist of first-order logic clauses that define the rules, legal actions, and goal conditions of a game. S-GDL is specifically designed for deterministic games with perfect information. However, it has since been extended to accommodate other types of games: S-GDL-II~\cite{schiffel2014representing} supports games with incomplete information, and S-GDL-III~\cite{thielscher2017gdl} allows the modeling of epistemic games, which involve reasoning about the knowledge and beliefs of other players~\cite{piette2019ludii}.\\

The development and advancement of GGP has been significantly driven by international competitions, particularly the International GGP Competition (IGGPC)~\cite{genesereth2013international} organized by Stanford University. These competitions have served as crucial benchmarks for evaluating the effectiveness of different AI approaches and have fostered innovation in the field by providing a standardized platform for comparing game-playing agents across diverse game types.\\

Several champion AI systems have emerged from these competitions~\cite{iggpc-winners}, some contributing significant advances to the field. These systems have progressively improved the state-of-the-art in GGP through innovations. \\

One early example is ClunePlayer~\cite{clune2007heuristic}, winner of the inaugural competition in 2005. ClunePlayer pioneered a novel methodology for automatically generating heuristic evaluation functions directly from the GDL representation of a game. Its core contribution lies in abstracting complex games into simplified models that capture essential dynamics, like the expected profit, control, and termination, using statistically stable features extracted from the game description. This approach produced interpretable and effective heuristics, allowing ClunePlayer to generalize successfully across a wide range of previously unseen games, thereby improving strategic reasoning in GGP environments.\\

Another prominent example is CadiaPlayer~\cite{bjornsson2009cadiaplayer}, having won the competition three times (2007, 2008, 2012). Unlike earlier systems that relied on minimax-based tree search guided by heuristics, CadiaPlayer introduced a paradigm shift by adopting Monte Carlo simulations for move selection. This innovation demonstrated the power and adaptability of simulation-based methods in general game environments and marked a significant step forward in the development of scalable, domain-independent decision-making algorithms.\\
Most notably, 
WoodStock~\cite{piette2017woodstock}~\cite{koriche2017constraint} achieved remarkable success by winning the 2016 IGGPC, the most recent edition of the tournament. WoodStock constitutes a major breakthrough in the field by introducing a constraint-based approach to game playing, fundamentally diverging from the more common Monte Carlo Tree Search (MCTS) strategies. By leveraging constraint satisfaction techniques for move selection and reasoning, WoodStock demonstrated the effectiveness of symbolic, logic-based reasoning in highly general and diverse game environments.

\subsubsection{Limitations}
Despite its strengths, S-GDL and the GGP-Base system face several limitations. Writing and debugging game descriptions in S-GDL requires a solid understanding of first-order logic, which can be a significant barrier for individuals without a background in computer science or mathematics.\\
Moreover, game components such as boards, decks, and even arithmetic operations must be defined explicitly for each game. As a result, creating or modifying game descriptions is often tedious and time-consuming. Even simple changes, like changing the size of a game board, can involve extensive code modifications~\cite{piette2019ludii}~\cite{piette2019empirical}.\\

Another major drawback is computational efficiency. Since S-GDL relies on logic resolution, parsing and processing game descriptions can be resource-intensive. This complexity limits the integration of S-GDL with external applications and hinders its potential use beyond AI game playing, such as in procedural content generation, game analysis, or educational tools~\cite{piette2019empirical}.\\


While GGP-Base allows for the representation of a wide variety of games, it is not well-suited for boardless games, which have evolving boards like \textit{Saboteur}, unless significant and complex workarounds are used. It is theoretically possible to simulate such games by predefining all possible board configurations and tile placements, however this approach is computationally expensive due to the combinatorial explosion of possibilities.

\subsection{Regular Boardgames}
Regular Boardgames (RBG)~\cite{kowalski2019regular} is a GGP language designed to formally describe a wide range of finite, deterministic, turn-based games with perfect information, especially board games. RBG is based on the theory of regular languages and aims to combine expressiveness, computational efficiency, and naturalness of game description in a single formalism.\\

RBG is capable of encoding a wide variety of board games, including those with complex rules and large branching factors, such as \textit{Amazons}, \textit{Arimaa}, large \textit{Chess} variants, \textit{Go}, international checkers, and paper soccer. RBG is designed for efficient computation and reasoning about game states and legal moves. Its use of regular languages allows for fast processing and avoids the computational overhead of logic-based languages like GDL. The language structure closely matches the intuitive structure of board games, making it easier for humans to write and understand game descriptions.

\subsubsection{Limitations}
RBG is fundamentally designed for games with a predefined, unchanging board structure. As mentioned in this source~\cite{kowalski2019regular}, [\textit{Board is a representation of the board ... . It is a static environment, which does not change during a play.}], this characteristic of RBG means that it cannot accommodate boardless games, or games in which the board evolves dynamically during gameplay, such as \textit{Carcassonne} among others. Since the board must remain static throughout the game, any mechanics that require the creation, removal, or modification of board elements as part of the gameplay are not supported within the RBG framework. Consequently, while RBG is well-suited for a wide variety of traditional board games with fixed layouts, its applicability is limited when it comes to games that feature a dynamically constructed or evolving play area.

\subsection{Ludii}
Ludii is a GGS developed as part of the Digital Ludeme Project\footnote{\url{http://ludeme.eu/}} (DLP). Ludii provides a unified framework for representing, modeling, playing and analyzing a wide range of board games from different cultures and historical periods. Built on the concept of \textit{ludemes}, it enables the compact and expressive specification of games using a domain-specific language, called L-GDL~\cite{piette2019ludii}~\cite{soemers2024ludii}. A ludeme is an element of play. It is a unit of structure or rules that can be used in various ways. A ludeme can be, among other things, a game mechanism, a type of decision, or a possible action. One of the main characteristics of a ludeme is that it can be shared between different games, whether or not they have the same root~\cite{parlett2016sa}.\\

Although Ludii was initially developed with the primary goal of building a comprehensive database of historical and modern board games, and to support research into their history, it also has the potential to be a flexible platform for various AI research topics and competitions~\cite{piette2019ludii}~\cite{stephenson2019ludii}.
The Ludii Games Database serves as a comprehensive resource for computational and cultural research on traditional board games~\cite{crist2024ludii}, establishing the foundations of digital arch{\ae}oludology as a new field of study~\cite{browne2019foundations}, which aims to analyse and reconstruct ancient games based on incomplete descriptions and archaeological evidence, using modern methods.\\
The system has been successfully applied to various research projects involving the analysis and reconstruction of traditional games. Notable examples include the analysis of the French Military Game~\cite{piette2021ludii_2} and computational approaches for recognising and reconstructing ancient games such as Ludus Latrunculorum~\cite{crist2023computational}. These studies demonstrate Ludii's capacity to contribute to both historical game research and cultural heritage preservation.\\
Although the DLP has concluded, the utilization of Ludii for game studies continues to expand into other domains through the GameTable research network, which aims to promote interdisciplinary collaboration in game research~\cite{piette2024gametable}.\\

Ludii distinguishes itself from earlier GGP systems by emphasizing human readability, efficiency, and support for a broader range of game types, including traditional board games, abstract strategy games, and even modern designer games. Ludii also supports logical puzzles, which can be solved using constraint programming techniques~\cite{piette2019ludii_2}. Its graphical interface and analytical tools make it accessible to both researchers and game designers, and its formal game descriptions facilitate comparative studies of gameplay mechanics and cultural evolution~\cite{soemersbridging}.\\

In recent years, the competitions based on Stanford’s GDL have been replaced by new competitions built around the Ludii system. The IGGPC, last held in 2016, faced growing limitations due to the low-level nature of Stanford GDL, which Ludii overcomes with a more expressive and structured game description language. The first competition using Ludii was organized in 2022, showcasing its potential as a new benchmark for GGP research and development~\cite{piette20232022}~\cite{stephenson2019ludii}.\\

Today, most research in GGP is conducted using Ludii. This includes work on enhancing MCTS through feature-based biasing techniques, demonstrating how domain-specific knowledge can be effectively incorporated into general algorithms~\cite{soemers2019biasing}. Other research focuses on the integration of self-play learning with policy gradient methods~\cite{soemers2019learning}, as well as the manipulation of experience distributions in expert iteration frameworks~\cite{soemers2020manipulating}. Further studies have explored the extraction and analysis of tactical patterns learned through self-play~\cite{soemers2023extracting}, and the use of spatial state-action feature representations to capture game-state information that generalizes across a wide variety of games~\cite{soemers2023spatial}. Research on general transfer learning approaches for policy-value networks also represents an important and active direction within the current GGP research~\cite{soemers2023towards}.\\

Ludii has been compared to both S-GDL and RBG through experimental studies, demonstrating significant advantages in usability and performance. In particular, Ludii offers more efficient reasoning across most evaluated games, and its game descriptions tend to be more compact and human-readable, thanks to its use of meaningful ludemes~\cite{piette2019empirical}.\\

Ludii is also used in educational contexts~\cite{stephenson2019ludii_2}, and supports the procedural generation of new board games using large language models (LLM)~\cite{todd2024gavel}, which can intelligently mutate and recombine existing game rules expressed in the Ludii game description language to produce novel and diverse games beyond those currently available in the system.

\subsubsection{Limitations}
While Ludii offers a powerful and flexible framework for GGP, it does present certain limitations when it comes to modeling and supporting boardless games, in which the playing area is dynamically constructed or expanded as players add tiles during gameplay. These limitations, along with the current implementation for such games in Ludii and the challenges it entails, will be discussed in the following section.

\chapter{Current implementation}
\label{current_implementation}
The primary objective of this thesis is to propose an innovative approach for managing scalable expansion of the board for baordless games.\\

In the current version of Ludii, for boardless games, the board is internally initialized with a fixed size at the beginning of the game. By default, a 41×41 board is generated for games based on square or triangular tiles, and a 21×21 board for those using hexagonal tiles.\\
This space, however, is hidden from the human user in order to create the illusion of a limitless playing area. In other words, although the dimensions are technically finite, the interface dissimulates the boundaries, allowing the player to perceive the board as potentially infinite. Unlike human users, AI agents have access to the entire board from the outset, which forces them to reason over a large, mostly empty space.\\ 

\begin{figure}[h!]
    \centering

    \begin{minipage}{0.48\textwidth}
        \centering
        \includegraphics[scale=0.25]{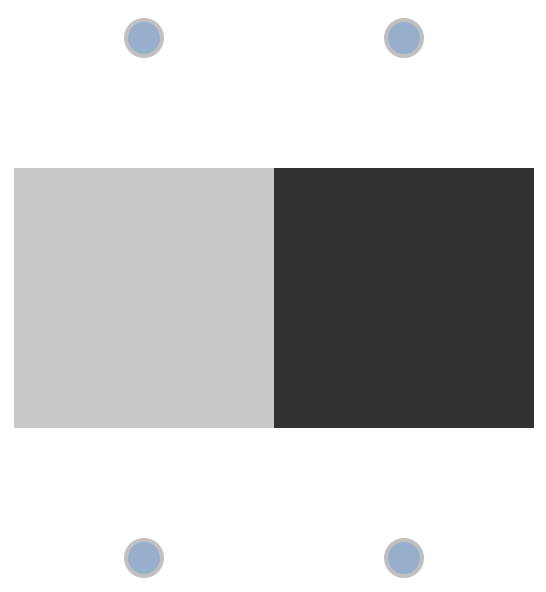}
    \end{minipage}
    \hfill
    \begin{minipage}{0.48\textwidth}
        \centering
        \includegraphics[scale=0.25]{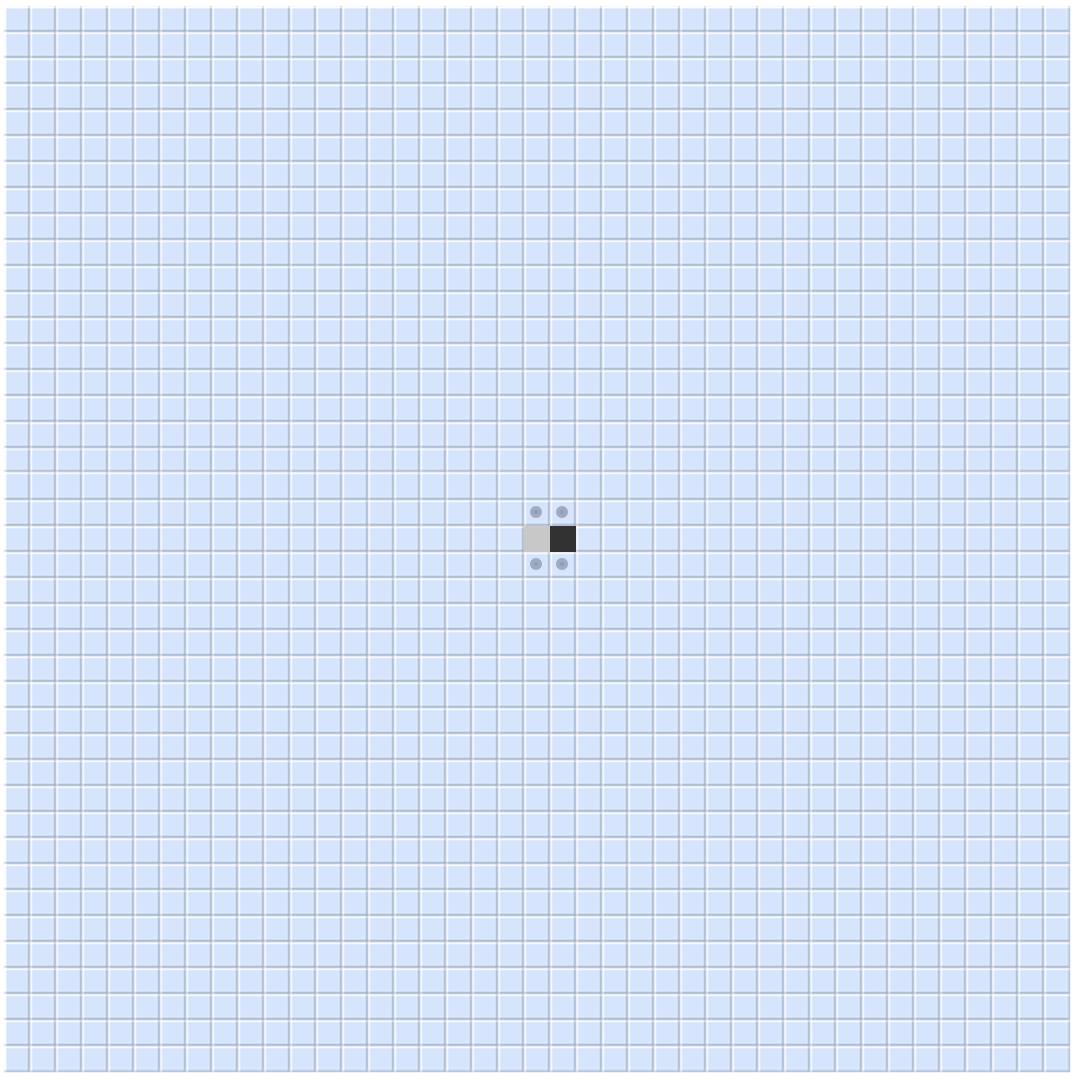}
    \end{minipage}
    
    \captionsetup{justification=centering}
    \caption*{(a) Square representation}

    \vspace{0.4cm}

    \begin{minipage}{0.48\textwidth}
        \centering
        \includegraphics[scale=0.3]{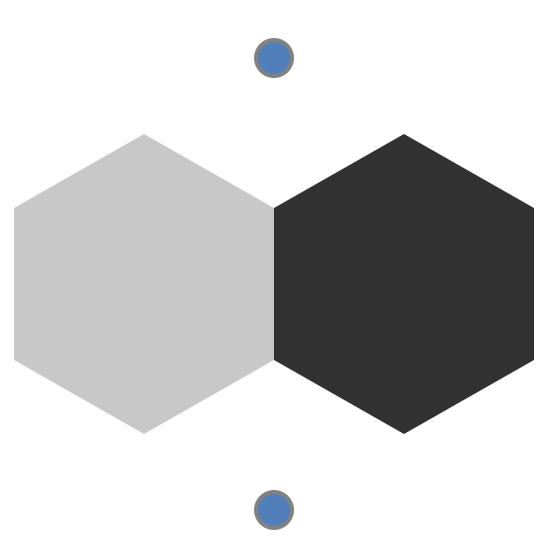}
    \end{minipage}
    \hfill
    \begin{minipage}{0.48\textwidth}
        \centering
        \includegraphics[scale=0.3]{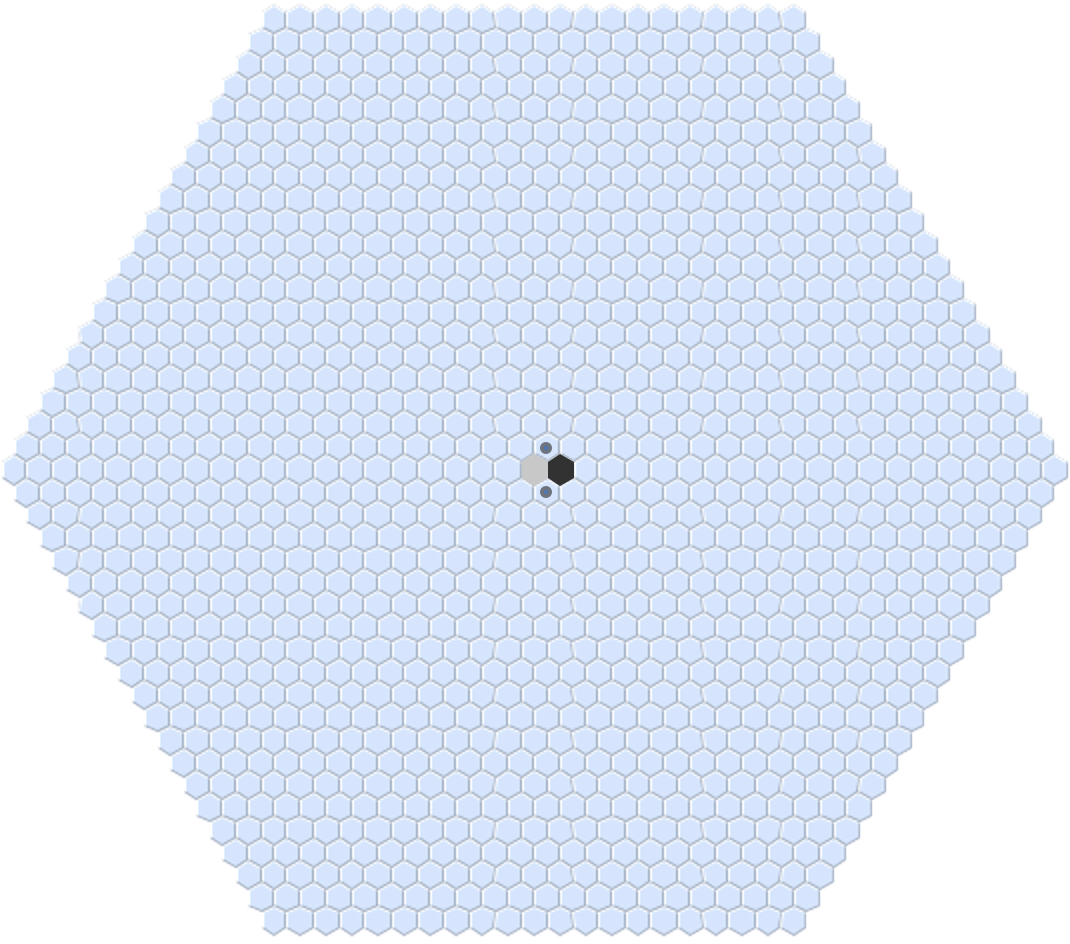}
    \end{minipage}

    \captionsetup{justification=centering}
    \caption*{(b) Hexagonal representation}

    \vspace{0.4cm}

    \begin{minipage}{0.48\textwidth}
        \centering
        \includegraphics[scale=0.3]{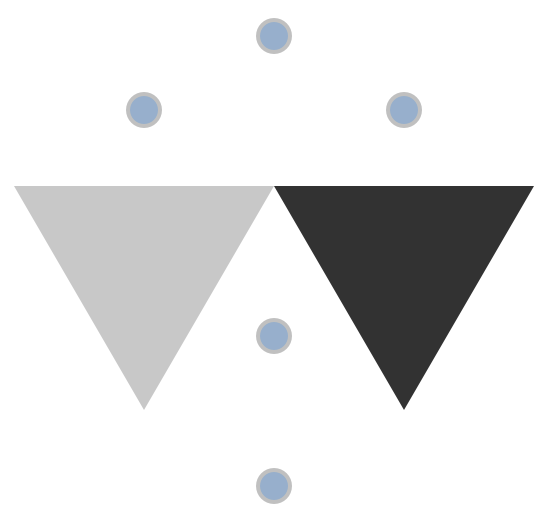}
    \end{minipage}
    \hfill
    \begin{minipage}{0.48\textwidth}
        \centering
        \includegraphics[scale=0.3]{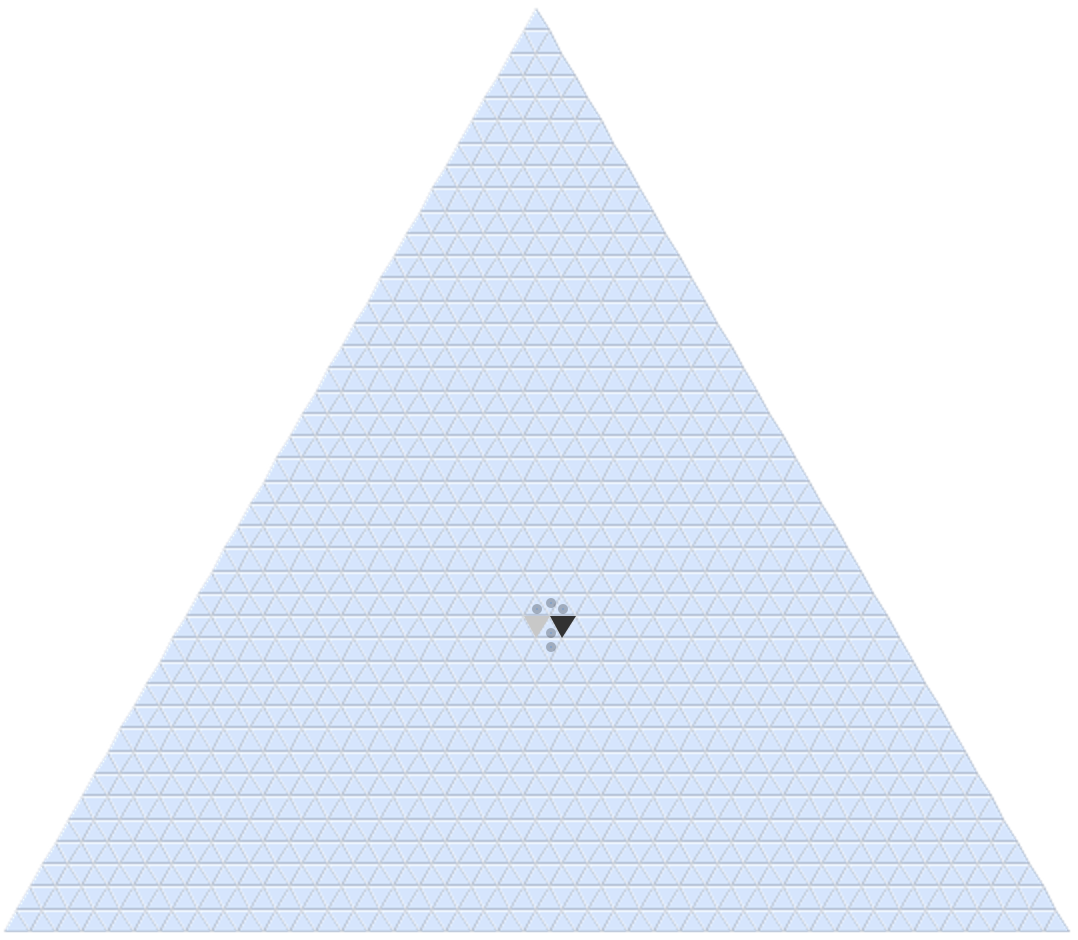}
    \end{minipage}

    \captionsetup{justification=centering}
    \caption*{(c) Triangular representation}

    \caption{Andantino game – visual (left) and logical (right) representations for different tile shapes}
    \label{fig:andantino_all}
\end{figure}

Figure \ref{fig:andantino_all} illustrates the board from a human user's perspective on the left column: only the playable areas are shown, represented here by the small blue dots indicating legal moves. In contrast, the right column of Figure \ref{fig:andantino_all} displays the board's internal logical representation, with a much larger and fully preallocated space.

\section{Limitations}
The current solution presents several significant limitations when applied to boardless games, in which the playing area is intended to expand indefinitely. There is a risk that players, by progressing in a particular direction, may eventually reach an invisible boundary, which would contradict the very rules of the game.\\

In addition to this gameplay constraint, the chosen implementation also has notable technical implications. Indeed, during game initialization, heavy precomputations are performed across the entire board to facilitate the internal logic of the game engine, particularly for AI models. This results in a significantly larger state representation than necessary, which really reduces the number of simulations (playouts) the AI can perform and negatively impacts its overall performance. These computations include, for instance, the determination of distances between all pairs of cells, the identification of possible connections, and the anticipation of certain occupation patterns. However, this step becomes extremely costly in terms of computation time and memory usage when the board is large, especially considering that a significant portion of these cells will likely never be used during the game. As a result, valuable resources are consumed to model a largely empty space, with little to no practical benefit.\\

Because of these precomputations, as well as the radial system, using a large fixed board for boardless games causes performance issues, in addition to result in an inaccurate representation of the game.\\
In Ludii, radials~\cite{browne2021general} are computational constructs used to represent straight, contiguous lines from a given position in specific directions, enabling efficient modeling of directional piece movements like sliding moves in chess. Each radial extends from a starting site along a fixed direction, connecting adjacent sites of the same type, which helps encode movement patterns. The current implementation for boardless games relies on a hardcoded 41x41 grid (for square and triangular tiles) and 21x21 for hexagonal tiles. This causes several key limitations related to radials:
\begin{itemize}
    \item \textit{Combinatorial Explosion}: With 1,681 positions on a 41x41 board for squares and triangles, and 1,141 positions on a 21x21 board for hexagons, each position can generate radials in up to 8 directions for squares, 6 for hexagons, and 3, 6, or 12 for various triangular tilings. This results in an enormous number of radials to compute and store, causing an exponential increase in complexity.
    \item \textit{Memory Consumption}: Precomputing all radials for such a large board requires significant memory, as the state space of legal positions expands rapidly.
    \item \textit{Preprocessing Time}: Radial generation becomes prohibitively slow because the system must analyze every position and direction before gameplay starts.
    \item \textit{Real-time Performance}: During play, move evaluation slows down as the system processes long radials, degrading the user experience and slowing down AIs.
\end{itemize}
These limitations show that having a huge fixed board is not suitable for boardless games.

\section{Goal of the thesis}
The central principle of this work is therefore to reduce the initial size of the board, in order to avoid allocating a disproportionate and largely under-utilized space at the beginning of the game. The idea is to start with a minimally sized board, just large enough to include the initial game elements, such as a starting tile or a small cluster of pre-placed tiles, and to allow for its progressive expansion based on the players' actions.\\

\begin{figure}[h!]
    \centering
    \includegraphics[width=1\linewidth]{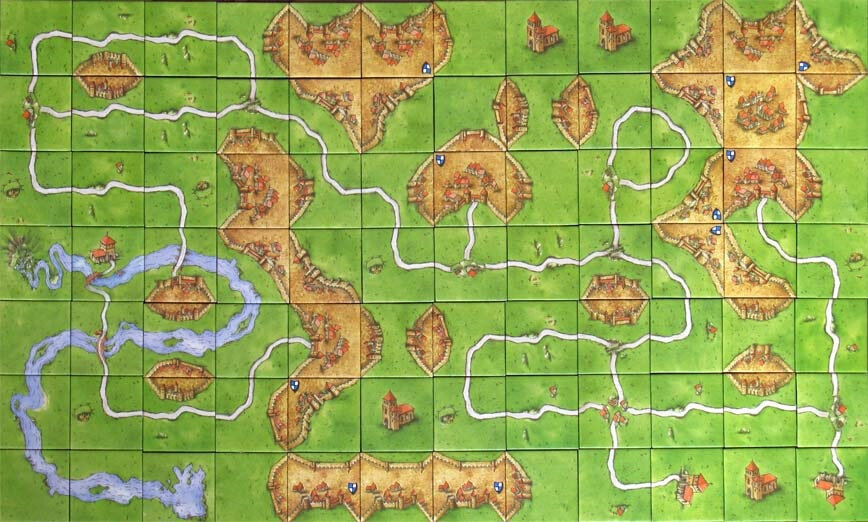}
    \captionsetup{justification=centering}
    \caption{Example of \textit{Carcassonne} play in an expanded game state~\cite{carcasonne}}
    \label{fig:carcasonne}
\end{figure}

Beyond memory and performance concerns, this approach also aims to provide a more faithful representation of boardless games, whose main characteristic is precisely the absence of predefined spatial boundaries.\\
In particular, many modern games, such as \textit{Carcassonne}, naturally lead to vast, sprawling play areas that far exceed the 41×41 limit currently imposed by Ludii’s default configuration, as shown on Figure \ref{fig:carcasonne}\footnote{Image taken from \url{https://ludilogiste.canalblog.com/pages/carcassonne/37993141.html}}. Relying on a fixed-size 'fake' board not only prevents such games from being accurately modelled, but also forces AI agents to reason over unnecessarily large search spaces, reducing their effectiveness.

By enabling dynamic board expansion, this thesis therefore seeks to overcome these limitations and open the door to more efficient AI strategies and more faithful representations of boardless games.

\section{Scope}
This thesis will focus more specifically on boards composed exclusively of regularly shaped tiles: square, hexagonal, or triangular. These shapes are widely used in board games and possess geometric properties that are particularly well-suited to dynamic expansion. Each shape entails a specific logic of growth due to its inherent geometry. Consequently, the method for dynamically expanding a board composed of square tiles differs fundamentally from that of a board made up of hexagonal or triangular tiles. Neighborhood rules, spatial coordinates, as well as the generation and placement methods for new cells, must be adapted to each shape, requiring the design of distinct mechanisms to manage board evolution depending on the chosen tile type.
Moreover, this thesis focuses exclusively on flat boards, those without stacks, thus limiting the scope to two-dimensional board expansions without vertical layering.

\chapter{Structures}
\label{structures}
To support the implementation of boardless games in Ludii, a set of core structures plays a central role in representing the various elements of a game. These classes are responsible for defining the board’s structure, managing the game state, handling player actions, and tracking ownership.\\

Although Ludii was originally designed for static boards, enabling dynamic behavior requires close interaction between the classes that handle the board’s structure (graph/topology), the evolving game state (state), the various containers that store not only the main board configuration but also other elements such as players’ hands, decks, and more (container, container state), the moves already executed (moves), and the ownership mechanisms (owned). Together, these existing structures form the foundation of a new approach that supports the dynamic expansion of the board. These structures must be adapted to allow for a scalable evolution of the board in response to players' actions. This section presents a detailed analysis of each relevant structure and explains how they currently work within the Ludii framework.

\section{State}
In Ludii, a game state represents all relevant information after $i$ moves played from the initial state $s_0$~\cite{piette2021ludii}. This state includes general information such as the player indices of the current, next, and previous mover, as well as the active players in the game (in some multiplayer games, players can be eliminated, thus becoming inactive). Additionally, it comprises information about the state of each game piece, with containers holding further state information, unlike components~\cite{browne2019ludii}.

\section{Container}
\label{sec:container}
Containers represent the physical structures that hold game components, such as boards, hands, decks, or any collection area where pieces can be placed. Each container can be modeled as a graph $G$ defined by a set $C$ of cells, a set $V$ of vertices and a set $E$ of edges. Each location $loc = \langle c_i, t_i, s_i, l_i \rangle$ is identified by its container $c = \langle C, V, E \rangle$, a type of site $t_i$ $\in$ {Cell, Vertex, Edge}, a site index $s_i$ $\geqslant$ 0 and a level $l_i$ $\geqslant$ 0 (typically 0 for most games, but higher values for stacking mechanics). Every location specifies a specific type of site in a specific container at a specific level~\cite{piette2021ludii}~\cite{browne2023ludii}. This hierarchical structure allows the game state to precisely encode the position and arrangement of components within each container.\\
A component refers to a basic game element, such as a piece, die, tile, or card, with fixed attributes (e.g., owner, size, value) and variable ones (e.g., state, orientation) that can change during gameplay~\cite{browne2019ludii}.

\subsection{Container state}
\label{sec:containerstate}

A container state $cs$ provides specialized state representations for each container in a game, each designed to minimize memory usage while optimizing data access during gameplay reasoning~\cite{piette2020ludii}. The system implements multiple container state types tailored to specific game mechanics, enabling Ludii to model a wide variety of different games~\cite{browne2023ludii}:
\begin{itemize}
    \item \textit{Flat state}: for games played on one single site type (without stacking).
    \item \textit{Graph state}: for games played on multiple site types (without stacking).
    \item \textit{Stack state}: for games played on one single site type (with stacking).
    \item \textit{Graph Stack state}: for games played on multiple site types (witch stacking).
    \item \textit{Deduction Puzzle state}: for puzzles corresponding to a Constraint Satisfaction Problem \cite{piette2019ludii_2}.
\end{itemize}

Each container state relies on a custom BitSet implementation, called ChunkSet, that compresses state information into minimal memory footprint based on the game's specific requirements~\cite{browne2019practical}~\cite{piette2020ludii}. A container state encodes information such as the placement of components on topological elements (cells, vertices, edges), as well as properties associated to each location, including piece orientation, stacking level, visibility, and other relevant attributes. 
The main chunks available in a flat state are described below, as this type of container state constitutes the primary focus of this master thesis~\cite{piette2021ludii}~\cite{browne2023ludii}:
\begin{itemize}
    \item \textbf{empty}: (1) if there is no component at the specific location id, else (0).
    \item \textbf{what}: The index of a component at a specific location, (0) if no component.
    \item \textbf{who}: The index of the owner of a component at a specific location, (0) if no component.
    \item \textbf{count}: The number of the same component at a specific location, (0) if no component.
    \item \textbf{state}: The local state of a component at a specific location, (0) if no component.
    \item \textbf{playable}: For boardless games, returning if a location is playable (1) or not (0).
\end{itemize}

\begin{figure}[h!]
    \centering
    \includegraphics[width=0.6\linewidth]{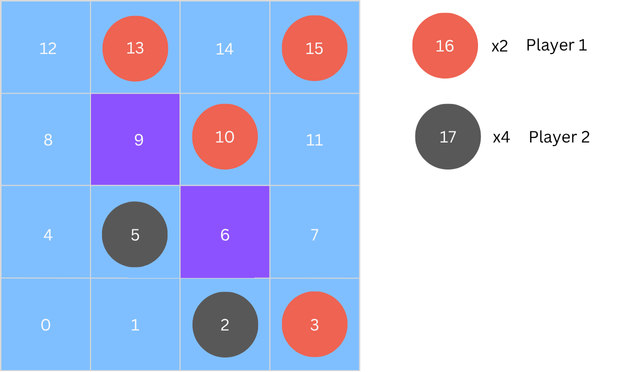}
    \captionsetup{justification=centering}
    \caption{Example of a board during a play}
    \label{fig:chunks_board}
\end{figure}

As an illustrative example, Figure \ref{fig:chunks_board} presents a game involving two players and a board. Player 1 has two chips (which is a specific component type) remaining in hand, with four already placed on the board, while player 2 has four chips in hand and two on the board. The purple tiles represent initial tiles, owned by the board, on which players are not allowed to play. The blue tiles represents the board itself, locations where player can may be able to play on, depending on the course of the game. The game is an instance of a flat state, as there is no stacking, and is played on a single site type (cells in this case). The board consists of 16 sites, numbered from 0 to 15, where players place their chips. In this context, Figure \ref{fig:chunks_board} can be interpreted as a visual representation of three containers: one for the main board and one for each player's hand. Accordingly, there are three distinct sets of chunks, each responsible for encoding the state of the sites within its corresponding container.\\

Table \ref{table:content_chunk_board} presents the different chunks associated with the container state representing the main board. The \textit{empty} chunk lists the indices of all cells that do not contain any component, which in this example corresponds to 10 different sites. The \textit{what} chunk specifies the indices of the occupied sites, each associated with the index of the component it contains. In this case, purple tiles are components assigned to index 1, red chips are components assigned to index 2, and black chips are components assigned to index 3. For instance, the entry \textbf{2:3} indicates that component 3 is placed on site 2, while \textbf{10:2} indicates that component 2 is on site 10. The \textit{who} chunk provides the indices of the occupied sites, each associated with the index of the player owning the respective component: the board is represented by ID 0, player 1 by ID 1, and player 2 by ID 2. For example, \textbf{2:2} means that the component on site 2 belongs to player 2, whereas \textbf{3:1} means that the component on site 3 belongs to player 1. The \textit{count} chunk records the number of components placed on each of the occupied sites; in this case, there is one component on sites \textbf{2, 3, 5, 6, 9, 10, 13,} and \textbf{15}, and no components on the remaining sites. The \textit{playable} chunk identifies the sites where players are still allowed to place components. Since components can only be placed on empty sites in this game, the values in the playable chunk are identical to those in the empty chunk.\\
\begin{table}[H]
    \centering
    \begin{tabular}{p{4.5cm}!{\color{white}\vrule}p{7.5cm}}
        \rowcolor{DarkGrey} Pre-generated sets & Description \\
        
        \rowcolor{LightGrey} empty &  0, 1, 4, 7, 8, 11, 12, 14 \\
        \rowcolor{LightLightGrey} what & 2:3, 3:2, 5:3, 6:1, 9:1, 10:2, 13:2, 15:2 \\
        \rowcolor{LightGrey} who & 2:2, 3:1, 5:2, 6:0, 9:0, 10:1, 13:1, 13:1  \\
        \rowcolor{LightLightGrey} count & 2:1, 3:1, 5:1, 6:1, 9:1, 10:1, 13:1, 15:1\\
        \rowcolor{LightGrey} state &  \\
        \rowcolor{LightLightGrey} playable & 0, 1, 4, 7, 8, 11, 12, 14  \\
    \end{tabular}
    \captionsetup{justification=centering}
    \caption{Example of chunks content for the container state of the game board}
    \label{table:content_chunk_board}
\end{table}

Table \ref{table:content_chunk_p1} presents the chunks associated with the container of the first player (in red). Since there are still at least one component in player's hand, there are no entries in the \textit{empty} chunks In this game, there is no possibilities to place a chip back into a player's hand, leading to the fact that the \textit{playable} chunk is also empty. The player's hand is represented by a single site, with index 0. In the \textit{what} chunk, the value corresponds to the index of the component held, which matches the component index used on the main board (see Table \ref{table:content_chunk_board}), in this case, index 2 for red chips. The \textit{who} chunk confirms that the component belongs to player 1 (with ID 1). Finally, the \textit{count} chunk indicates that two components are present on site 0, meaning player 1 still has two chips in hand.\\
\begin{table}[H]
    \centering
    \begin{tabular}{p{4.5cm}!{\color{white}\vrule}p{7.5cm}}
        \rowcolor{DarkGrey} Pre-generated sets & Description \\
        
        \rowcolor{LightGrey} empty &  \\
        \rowcolor{LightLightGrey} what & 0 : 2 \\
        \rowcolor{LightGrey} who & 0 : 1  \\
        \rowcolor{LightLightGrey} count & 0 : 2\\
        \rowcolor{LightGrey} state &  \\
        \rowcolor{LightLightGrey} playable & \\
    \end{tabular}
    \captionsetup{justification=centering}
    \caption{Example of chunks content for the container state of the first player}
    \label{table:content_chunk_p1}
\end{table}

It is important to note that, in the context of global site indexing, this local index 0 does not correspond to the global index 0. In this example, the site representing player 1's hand corresponds to global index 16, while the site for player 2's hand corresponds to index 17. As chunk indexing is local to each container, it always starts at 0, whereas global indexing ensures a unique identifier for every site across all containers.\\

Table \ref{table:content_chunk_p2} presents the chunks associated with the container of the second player (in gray). Similar to player 1, there are no \textit{empty} or \textit{playable} entries, and the hand is represented by a single site, also with index 0. The \textit{what} chunk shows that the held component is of index 3, corresponding to gray chips on the main board. The \textit{who} chunk indicates that the component belongs to player 2 (with ID 2). According to the \textit{count} chunk, four components are present on site 0, which means player 2 still has four chips in hand.
\begin{table}[H]
    \centering
    \begin{tabular}{p{4.5cm}!{\color{white}\vrule}p{7.5cm}}
        \rowcolor{DarkGrey} Pre-generated sets & Description \\
        
        \rowcolor{LightGrey} empty &  \\
        \rowcolor{LightLightGrey} what & 0 : 3 \\
        \rowcolor{LightGrey} who & 0 : 2  \\
        \rowcolor{LightLightGrey} count & 0 : 4\\
        \rowcolor{LightGrey} state &  \\
        \rowcolor{LightLightGrey} playable & \\
    \end{tabular}
    \captionsetup{justification=centering}
    \caption{Example of chunks content for the container state of the second player}
    \label{table:content_chunk_p2}
\end{table}

\section{Graph}
The graph constitutes the core of the board representation in Ludii. A board is first defined as a graph, a flexible and powerful data structure particularly well-suited for modeling simple, complex or irregular topologies, as encountered in board games. A graph is composed of several fundamental elements: a set $C$ of cells (also referred to as faces), a set $V$ of vertices, and a set $E$ of edges. These components are interconnected: each vertex can be linked to one or more other vertices via edges. When multiple vertices are connected in such a way that they form a loop, this enclosed space defines a cell.\\

At the code level in Ludii, each graph is then transformed into a Topology object, which serves as an internal representation of the graph's structure. This topological approach allows Ludii to represent boards as mathematical graphs where vertices, edges, and cells serve as playable sites with defined spatial relationships and directional properties. The topology system performs extensive precomputations at the start of a game. This includes calculating adjacency relationships, supported directions, as well as identifying which elements are on the perimeter, inside, to the left, to the right, and many other spatial properties. All of these data structures are built once when the game is initialized, ensuring that this information can be accessed extremely quickly during gameplay. This approach optimizes performance, as complex spatial queries are resolved in advance and do not need to be recalculated repeatedly during the game.\\
As mentioned in Section~\ref{sec:container}, each container in Ludii is represented as a graph, not only the main board. Consequently, each container is also translated into a Topology object.

\subsection{Square}
\begin{figure}[H]
    \centering
    \includegraphics[width=0.9\linewidth]{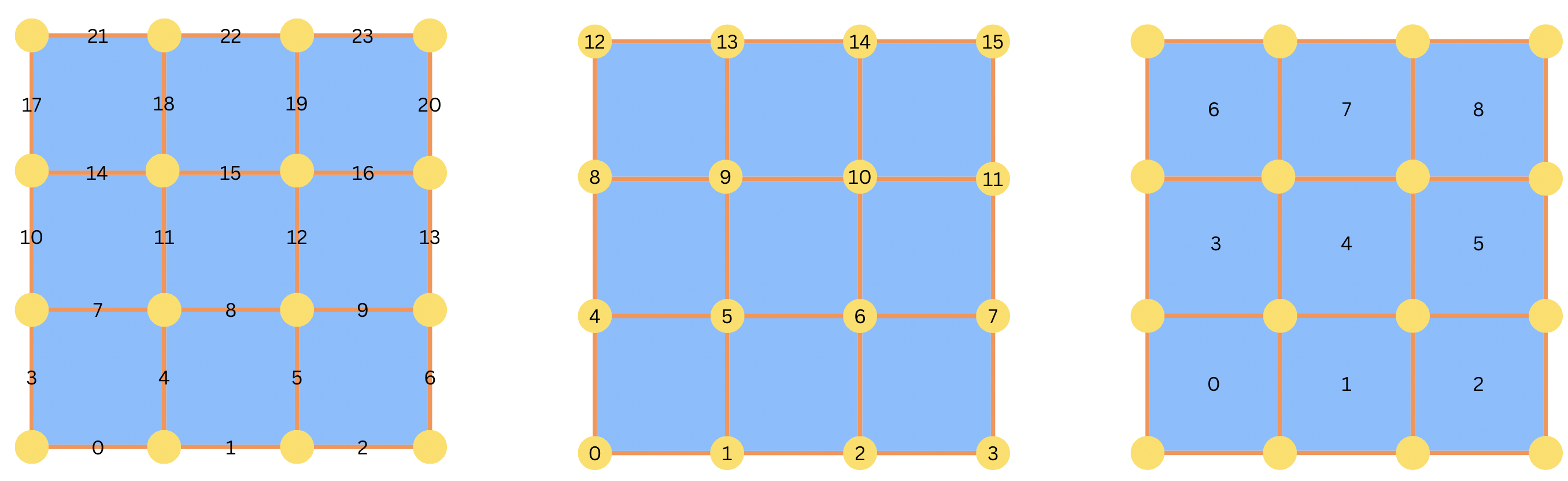}
    \captionsetup{justification=centering}
    \caption{Graph of 3x3  with square tiles}
    \label{fig:graph_square}
\end{figure}

In the case of square tiles, each Face is associated with four vertices and four edges. This organization is illustrated in Figure \ref{fig:graph_square}, where the image shows how each element is identified and connected within a graph $G$, enabling easy tracking and management of these components throughout the game. Indices start at 0.

On Figure \ref{fig:graph_square}, the blue surfaces represent the cells, the yellow dots represent the vertices, and the orange lines represent the edges.

\subsection{Hexagonal}
\begin{figure}[h!]
    \centering
    \includegraphics[width=1\linewidth]{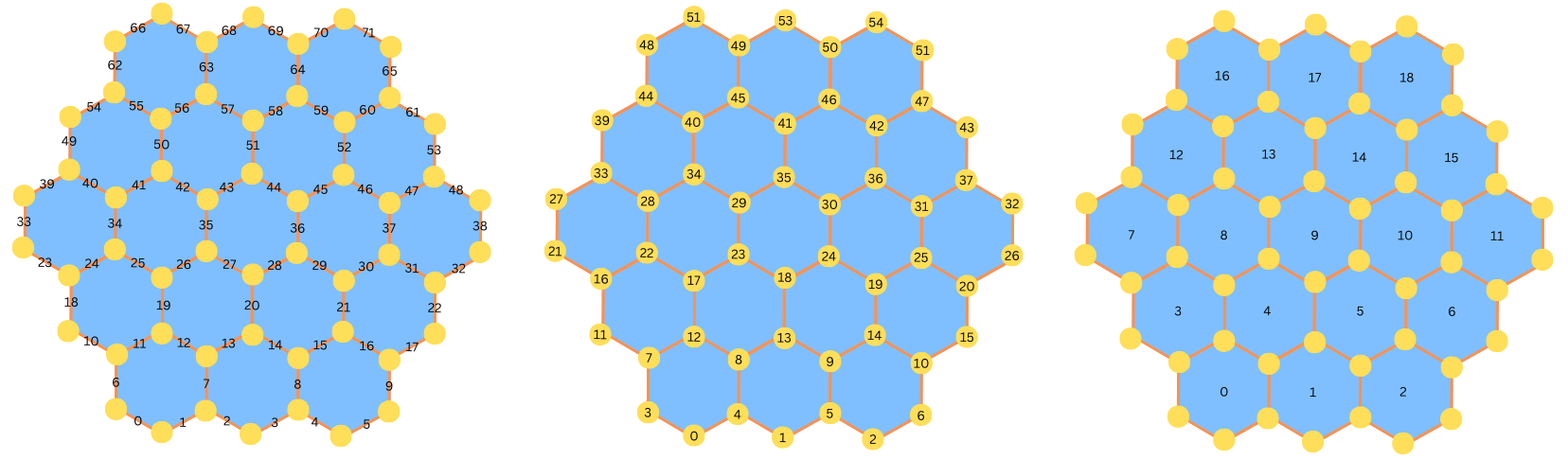}
    \captionsetup{justification=centering}
    \caption{Graph of 3x3 with hexagonal tiles}
    \label{fig:graph_hexagonal}
\end{figure}

In the case of hexagonal tiles, each Face is associated with six vertices and six edges. This organization is illustrated in Figure \ref{fig:graph_hexagonal}.

\subsection{Triangular}
\begin{figure}[h!]
    \centering
    \includegraphics[width=0.9\linewidth]{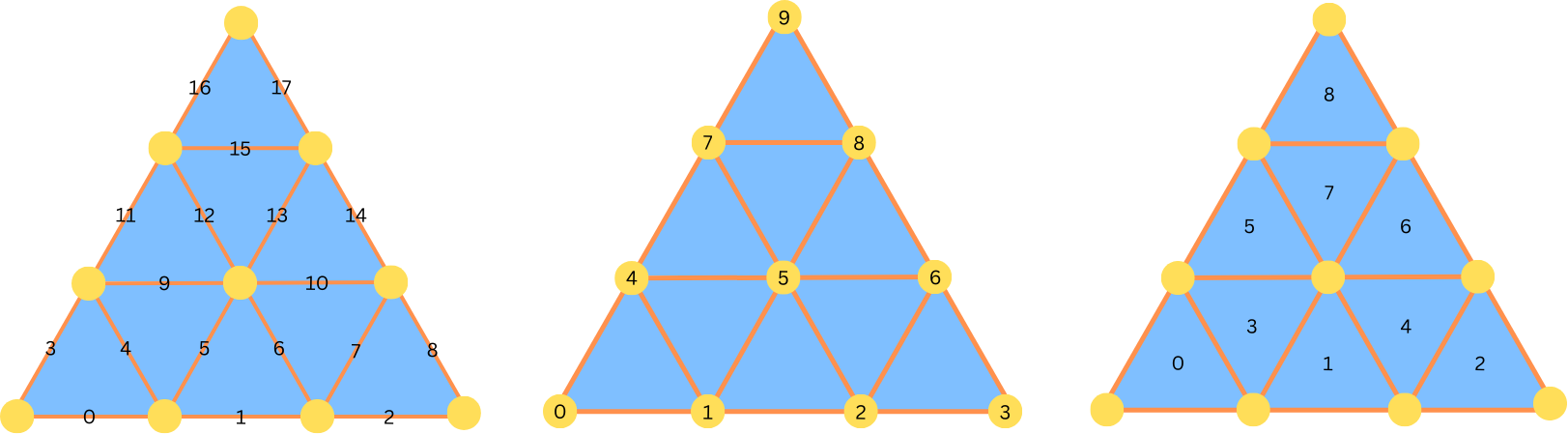}
    \captionsetup{justification=centering}
    \caption{Graph of 3x3  with triangular tiles}
    \label{fig:graph_triangular}
\end{figure}

In the case of triangular tiles, each Face is associated with three vertices and three edges. This organization is illustrated in Figure \ref{fig:graph_triangular}.

\section{Owned}
The Owned class plays a central role in managing the elements present on the board or in the players' hand. It represents an optimized data structure that allows for direct and low-complexity access to the location of each game component (pieces, tiles, etc.), based on their owner and component type. More specifically, Owned establishes a quick link between each component, its position in the game space, and the container to who it belongs.\\
Therefore, it will be necessary to update the owned structure, as it retains the identifiers of the sites occupied by game components. In the context of a boardless game, these identifiers are likely to change throughout the game, especially when they need to be re-contextualized according to the transformation of the board (whether the board is expanding or reducing in size). It is therefore essential for the structure to account for this variability in order to ensure the consistency and integrity of the game state at all times.\\

For each container present in the current game, a corresponding sublist is maintained in the owned structure. Each of these sublists is then subdivided into sets, with each set representing a specific type of component belonging to that container.\\
In the case of a container associated with a player (its hand), these sets contain the components that the player owns throughout the game. For the container of the main board, the sets include the elements shared between all players, as well as the components that make up the initial structure of the board itself.\\
For example, in the game \textit{Scrabble}, the letter tiles placed on the board are shared components accessible to all players. Another example, that includes initial tiles, is \textit{Carcassonne}, in which a starting tile is placed on the board to initiate the game. \\

Each set stores the unique identifiers of the sites on which the corresponding components are located, thus allowing for precise tracking of the distribution of pieces on the board and in the reserves. Based on the example of the board shown in Figure \ref{fig:chunks_board}, here is what the content of the owned structure would look like for this particular configuration:
\begin{table}[H]
    \centering
    \begin{tabular}{p{4.5cm}!{\color{white}\vrule}p{7.5cm}}
        \rowcolor{DarkGrey} Pre-generated sets & Description \\
        
        \rowcolor{LightGrey} Board & [\{6, 9\}] \\
        \rowcolor{LightLightGrey} Player 1 & [\{3, 10, 13, 15, 16\}] \\
        \rowcolor{LightGrey} Player 2 & [\{2, 5, 17\}] \\
    \end{tabular}
    \captionsetup{justification=centering}
    \caption{Example of Owned content}
    \label{table:content_owned}
\end{table}
Indeed, Players 1 and 2 each have a single set, which can be explained by the fact that each possesses only one type of component, chips. The board also contains one set, as it holds only one type of component, namely the initial tiles.\\
Player 1’s set for its only component includes the indices 3, 10, 13, and 15, as shown in Figure \ref{fig:chunks_board}, where red chips are located at these indexed sites. It is also important to note that the set includes site 16, despite it not being present on the main board. The site 16 corresponds to the site in player 1’s hand, and including this value facilitates understanding that this component can also be found in the player 1’s hand.

\section{Moves}
In Ludii, a move refers to an action performed by a player that modifies the current state of the game. It is the fundamental unit of transition between two successive game states: each move encapsulates a specific transformation, whether simple or complex, applied to the elements of the board, the players' reserves, or any other structure within the system.

A move can cover a wide variety of operations, including but not limited to:
\begin{itemize}
    \item Adding a piece to the board, such as when a player places a tile or a piece during their turn.
    \item Moving a piece from their hand to the board, in the case where players have a personal reserve.
    \item Moving a piece already on the board, as in a game where elements can be repositioned (e.g., chess, where a piece can move from one square to another).
    \item Removing a piece, in the case of a capture or a game effect that removes an element from the board.
\end{itemize}

Each move encapsulates a set of essential information, allowing the action to be correctly applied and ensuring its traceability. Among these details, some are particularly important within the scope of this thesis, including:
\begin{itemize}
    \item \textit{from}: The starting index of a move. It may be null if the piece is being added.
    \item \textit{to}: The destination index of a move.
    \item \textit{edge}: If the move was made on an edge of the board.
\end{itemize}

In the specific context of a boardless game, the move becomes a central point of analysis, as it can also be responsible for expanding the board itself. For example, when a player adds a new tile on an edge of the existing board, this move involves a structural update of the game space.\\
Indeed, it is the moves that will, in some cases, trigger the processes for updating the board. They act as signals indicating that an expansion of the game's spatial structure is necessary to reflect the new configurations introduced by the players' actions.

\chapter{Improvement} 
\label{amelioration}

This section addresses the strategies developed during this thesis to enable the dynamic expansion of the game board in Ludii. Rather than relying on a large, fixed-size board as the current implementation does, the objective is to begin with a minimal initial structure and allow the board to expand progressively in response to the players' actions. To achieve this, two main techniques for managing board expansion are explored. These techniques are then analyzed in the context of their integration within Ludii’s existing architecture, where two principal methods for updating the game state to accommodate the increasing board size have been identified. The section concludes with a discussion on the method used to determine the initial size of the board before any expansion occurs.

\section{Contour}
The first approach involves expanding the perimeter surrounding the existing board, thereby creating a contour prepared to accommodate new tiles. This method ensures that the board always has free space for any future expansion, regardless of the direction chosen by the players. For each tiling considered in this master’s thesis, this is illustrated in Figures \ref{fig:edgemove_square_contour}, \ref{fig:edgemove_hexagonal_contour}, and \ref{fig:edgemove_triangular_contour}.

\subsection{Square}
To determine the number of tiles added in the context of a board using square tiles, the following formula can be used:

\begin{equation}
nbAddedCell_{s} = (4 * (prevDim_{s} + 2)) - 4
\end{equation}
where $nbAddedCell_{s}$ represents the number of added celle at step $s$, and $prevDim_{s}$ denotes the dimension (either width or height, both are equal) of the previous board at step $s$.\\

\begin{figure}[H]
    \centering
    \includegraphics[width=0.5\linewidth]{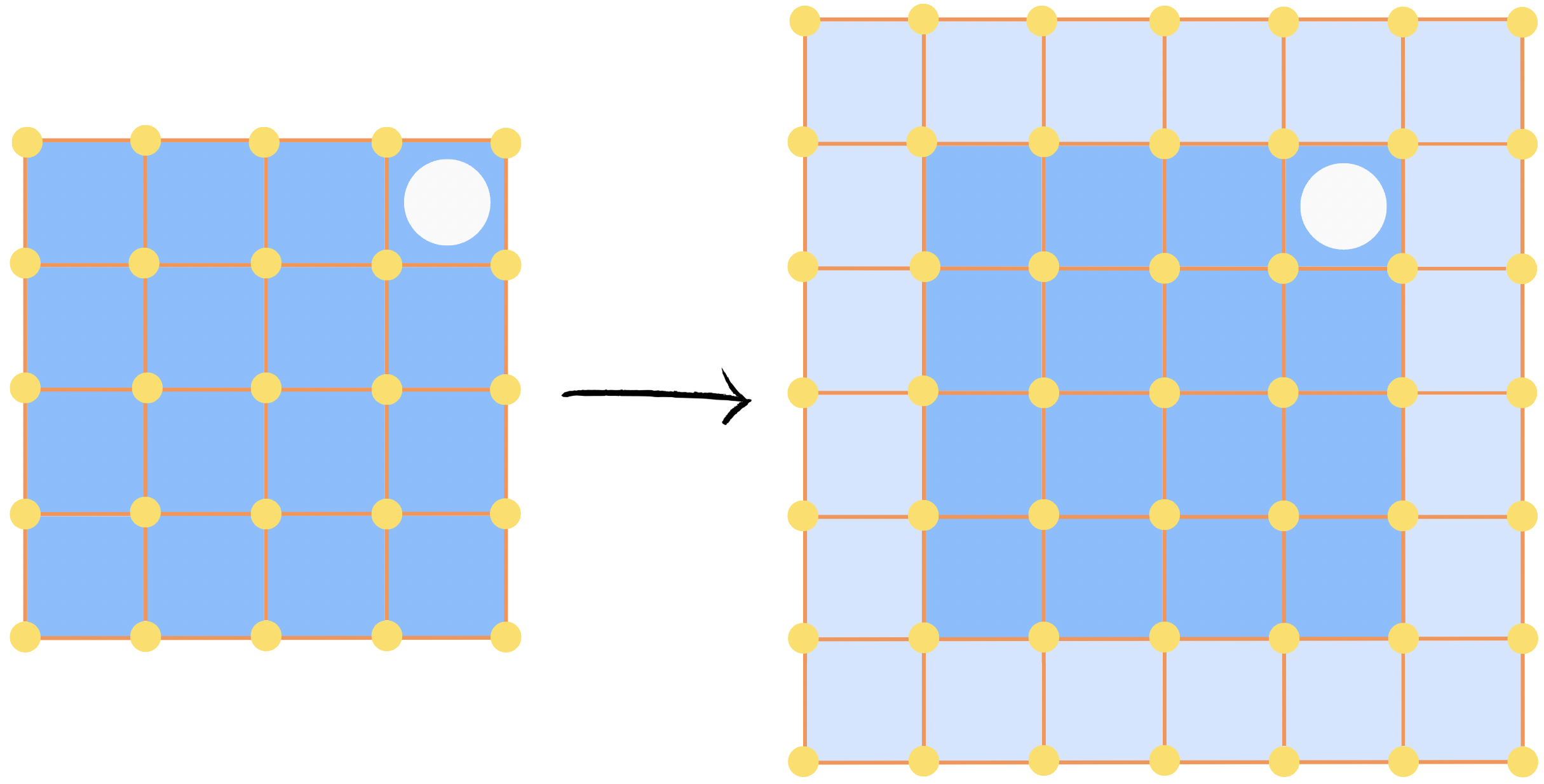}
    \captionsetup{justification=centering}
    \caption{Perimeter strategy when placing a tile on a square board}
    \label{fig:edgemove_square_contour}
\end{figure}

As an example, consider Figure \ref{fig:edgemove_square_contour}. When expanding the board from size 4 to size 6, the number of new cells added is 20:
\begin{equation*}
    (4 * (4 + 2)) - 4 = 20
\end{equation*}

\begin{figure}[h!]
    \centering
    \includegraphics[width=0.6\linewidth]{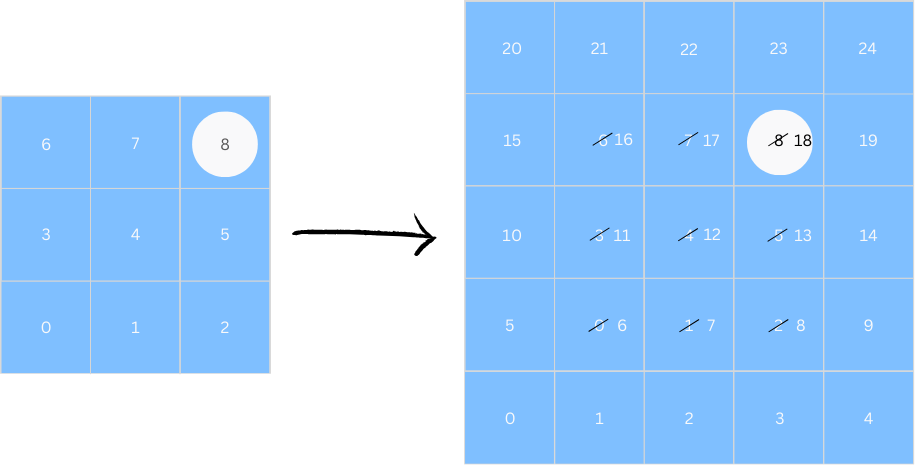}
    \captionsetup{justification=centering}
    \caption{Mapping between a square board of size 3 to size 5}
    \label{fig:mapping_square_1}
\end{figure}

\paragraph{Mapping}
As the board expands, it becomes necessary to identify how the indices from the previous board map to those on the new, larger board. This is achieved through a \textbf{mapping} process. It would be incorrect to assume that index 0 on the previous board will necessarily correspond to index 0 on the expanded board. For example, as illustrated on the left side of Figure \ref{fig:mapping_square_1}, if a player plays on the site with index 0, the addition of a perimeter around the board causes the original index 0 to become index 6 in the new board. This mechanism ensures that players can continuously play along the edges of the board.\\
The mapping technique shown in Figure \ref{fig:mapping_square_1} applies specifically to boards composed of square tiles.

In this context, each cell \textit{i} is assigned a new identifier, denoted $newCellID_{i}$, which can be calculated using the following formula:
\begin{equation}
    newCellID_{i} = (prevDim_{s} + 1) + (2 * (r_{i} + 1)) + prevCellID_{i}
\end{equation}

where $prevDim_{s}$ denotes the dimension (width or height, both being equal) of the previous board at step \textit{s}, $r_i$ is the row index of cell $i$ (starting from 0, from bottom to top), and $prevCellID_{i}$ is the identifier of that cell on the previous board.\\
As an example, consider Figure \ref{fig:mapping_square_1}. For site 7, the new identifier is calculated as follows:
\begin{equation*}
    (3+1)+(2*(2+1))+7=17
\end{equation*}
Here, the previous board has a dimension of 3, and site 7 is located on row 2. Hence, its mapped value on the new board is 17.

\subsection{Hexagonal}
To determine the number of tiles added in the context of a board using hexagonal tiles, the following formula can be used:

\begin{equation}
nbAddedCell_{s} = (6 * (prevDim_{s} + 1)) - 6
\end{equation}

\begin{figure}[h!]
    \centering
    \includegraphics[width=0.6\linewidth]{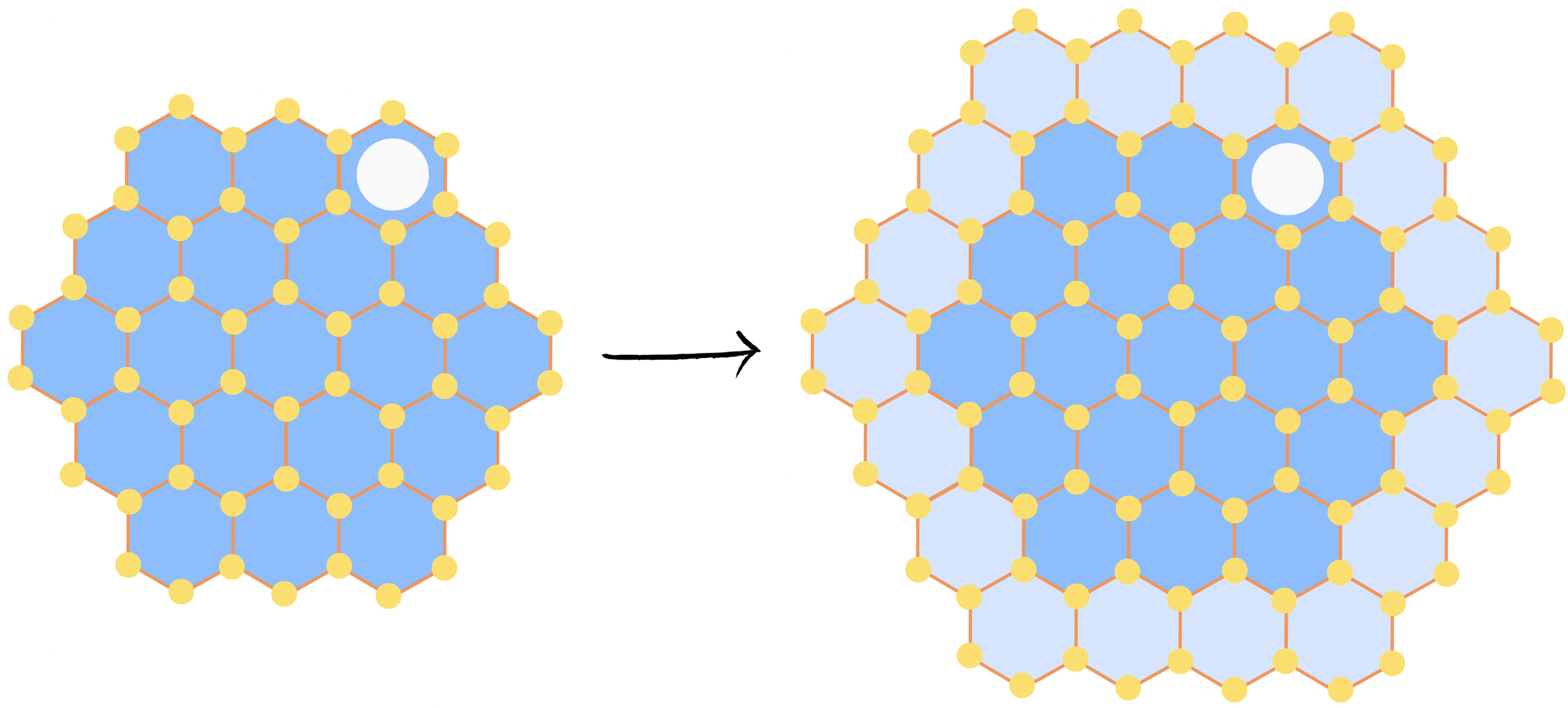}
    \captionsetup{justification=centering}
    \caption{Perimeter strategy when placing a tile on a hexagonal board}
    \label{fig:edgemove_hexagonal_contour}
\end{figure}

In the case of a hexagonal board, illustrated in Figure \ref{fig:edgemove_hexagonal_contour}, increasing the board dimension from 3 to 4 adds 18 new cells, determined by:
\begin{equation*}
    (6 * (3 + 1)) - 6=18
\end{equation*}

\begin{figure}[h!]
    \centering
    \includegraphics[width=0.7\linewidth]{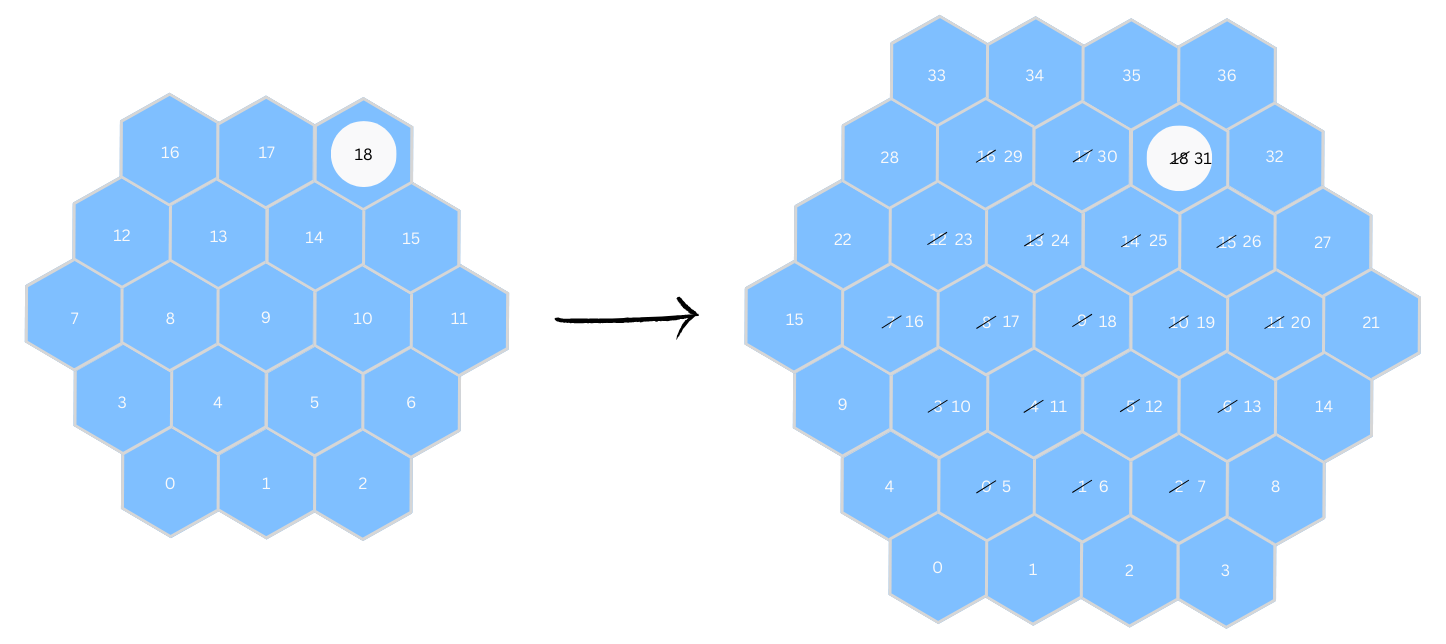}
    \captionsetup{justification=centering}
    \caption{Mapping between a hexagonal board of size 3 to size 4}
    \label{fig:mapping_hexagonal_1}
\end{figure}

\paragraph{Mapping}
The mapping technique illustrated in Figure \ref{fig:mapping_hexagonal_1} applies to boards composed of hexagonal tiles.

In this context, each cell \textit{i} is assigned a new identifier, which can be calculated using the following formula:
\begin{equation}
newCellID_{i} = prevDim_{s} + (2 * (r_{i} + 1)) + prevCellID_{i}
\end{equation}

Figure \ref{fig:mapping_hexagonal_1} illustrates this in practice: applying the formula to site 14, row 3, on a board of dimension 3 results in index 25:
\begin{equation*}
    3+(2*(3+1))+14=25
\end{equation*}

\subsection{Triangular}
To determine the number of tiles added in the context of a board using triangular tiles, the following formula can be used:

\begin{equation}
    nbAddedCell_{s} = (3 * (((prevDim_{s} + 3) * 2) - 1)) - 6
\end{equation}

\begin{figure}[h!]
    \centering
    \includegraphics[width=0.5\linewidth]{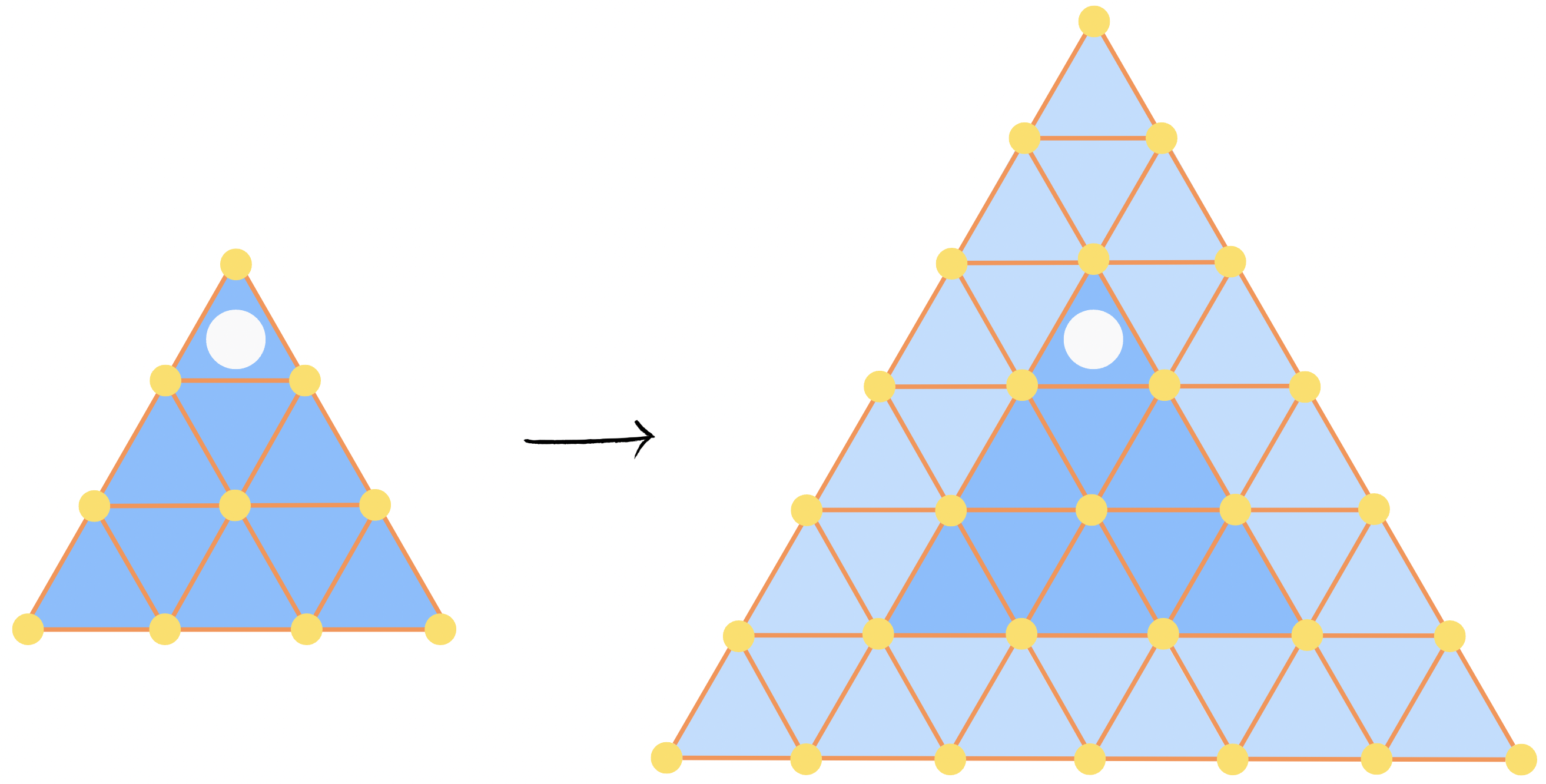}
    \captionsetup{justification=centering}
    \caption{Perimeter strategy when placing a tile on a triangular board}
    \label{fig:edgemove_triangular_contour}
\end{figure}

Using Figure \ref{fig:edgemove_triangular_contour} as a reference, the expansion of the triangular board from size 3 to size 6 results in 27 new cells being added, as shown by the following calculation:
\begin{equation*}
    (3 * (((3 + 3) * 2) - 1)) - 6=27
\end{equation*}

\begin{figure}[h!]
    \centering
    \includegraphics[width=0.7\linewidth]{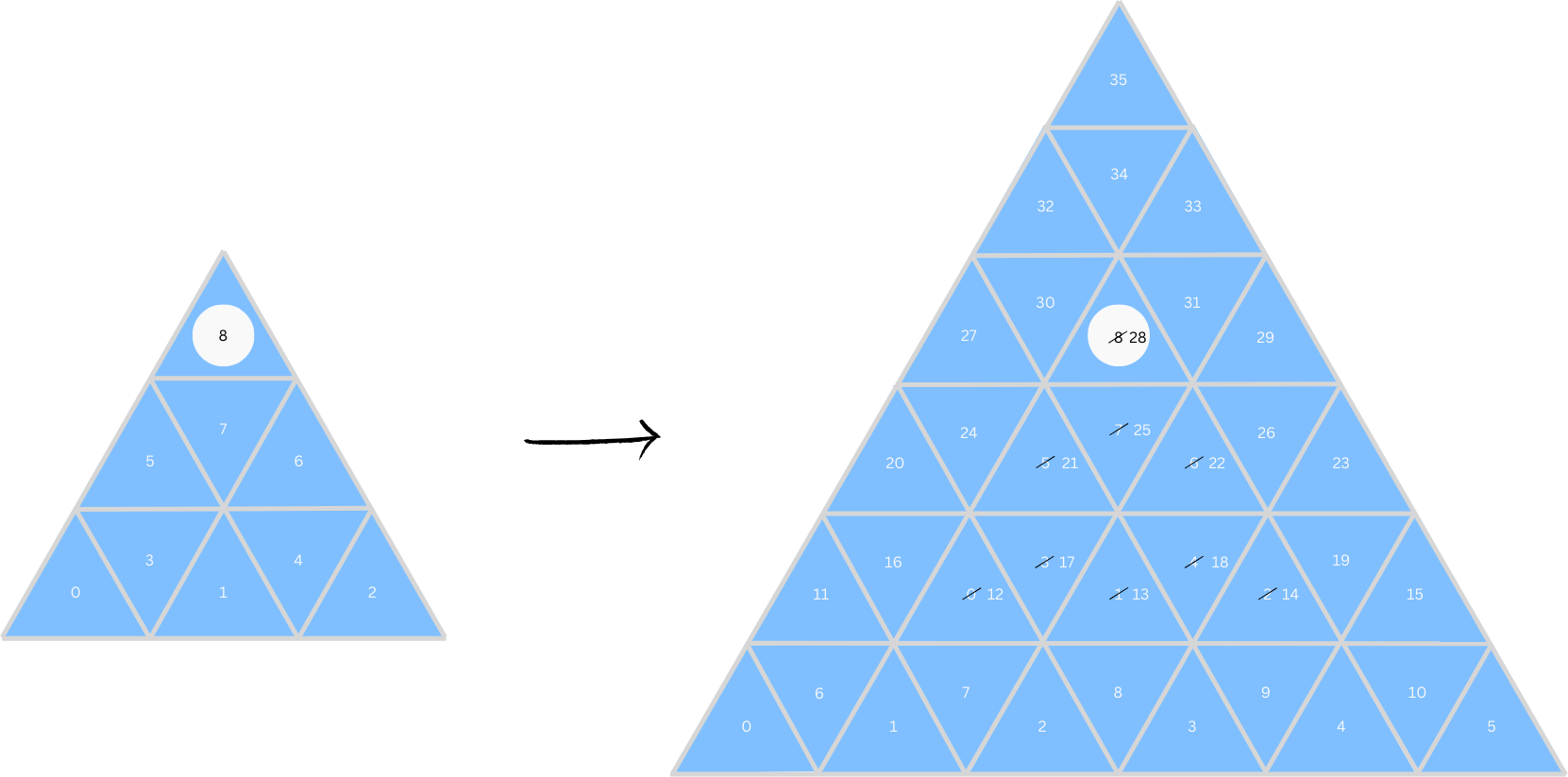}
    \captionsetup{justification=centering}
    \caption{Mapping between a triangular board of size 3 to size 6}
    \label{fig:mapping_triangular_1}
\end{figure}

\paragraph{Mapping}
The mapping technique depicted in Figure \ref{fig:mapping_triangular_1} applies to boards composed of triangular tiles.

In this context, each cell \textit{i} is assigned a new identifier, which can be calculated using the following formula:
\begin{equation}
newCellID_{i} = ((prevDim_{s} + 3) * 2) - 2 + (2 * (r_{i} + 1)) + prevCellID_{i}
\end{equation}

For instance, in Figure \ref{fig:mapping_triangular_1}, site 7 on a triangular board of dimension 3 and located on row 3 is mapped to index 25, as shown by the calculation:
\begin{equation*}
    ((3+3)*2)-2+(2*(3+1)) + 7=25
\end{equation*}

Row and column indexing differs when dealing with a triangular board. On the right board in Figure \ref{fig:mapping_triangular_1}, triangles pointing upwards and triangles pointing downwards do not belong to the same columns. As a result, sites 0 to 5 are on row 0, while sites 6 to 10 are on row 1. For example, site 35 is located on row 10.

\section{Zone}
\label{sec:zone}
\begin{figure}[h!]
    \centering
    \includegraphics[width=0.5\linewidth]{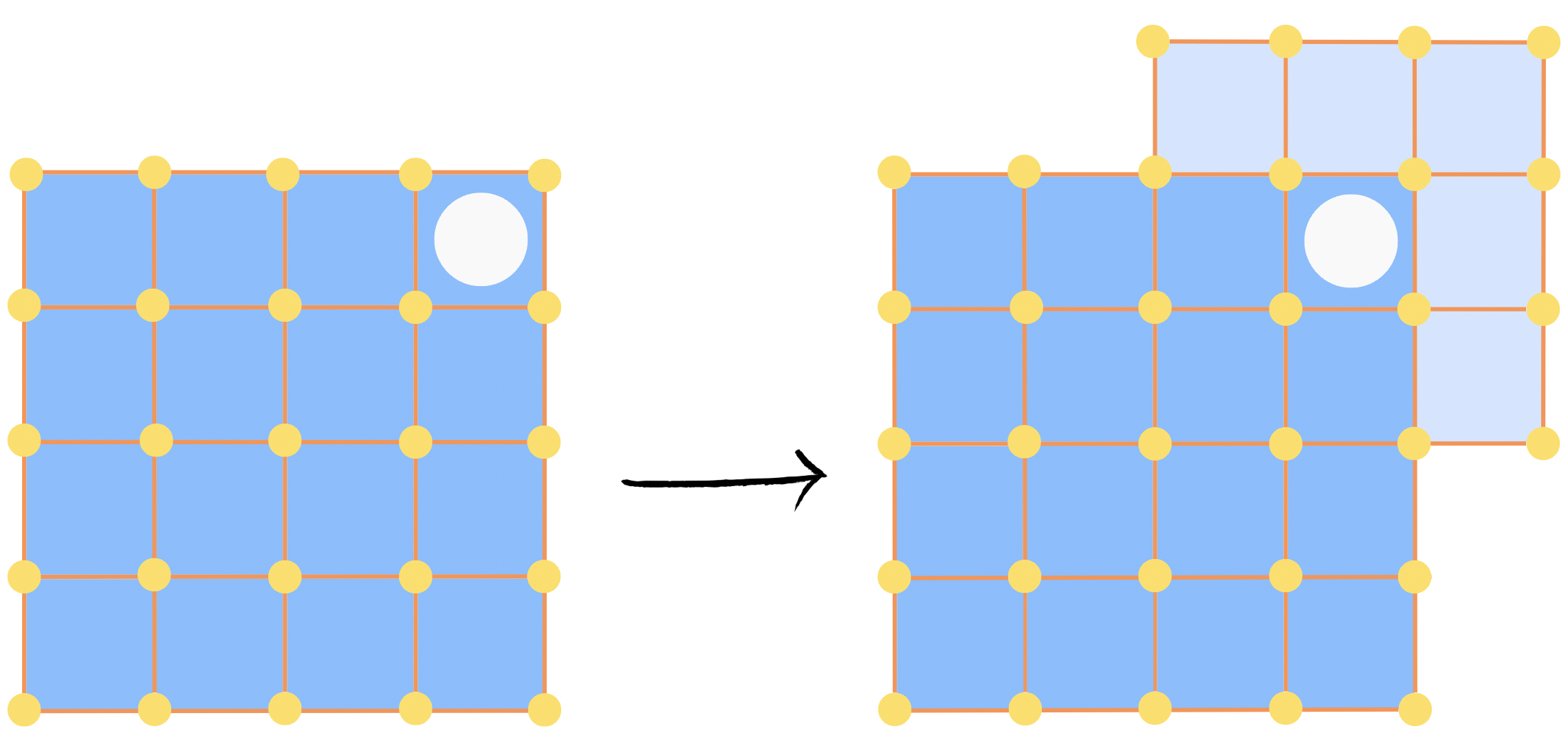}
    \captionsetup{justification=centering}
    \caption{Zone strategy when placing a tile on a square board}
    \label{fig:edgemove_square_zone}
\end{figure}

\begin{figure}[h!]
    \centering
    \includegraphics[width=0.6\linewidth]{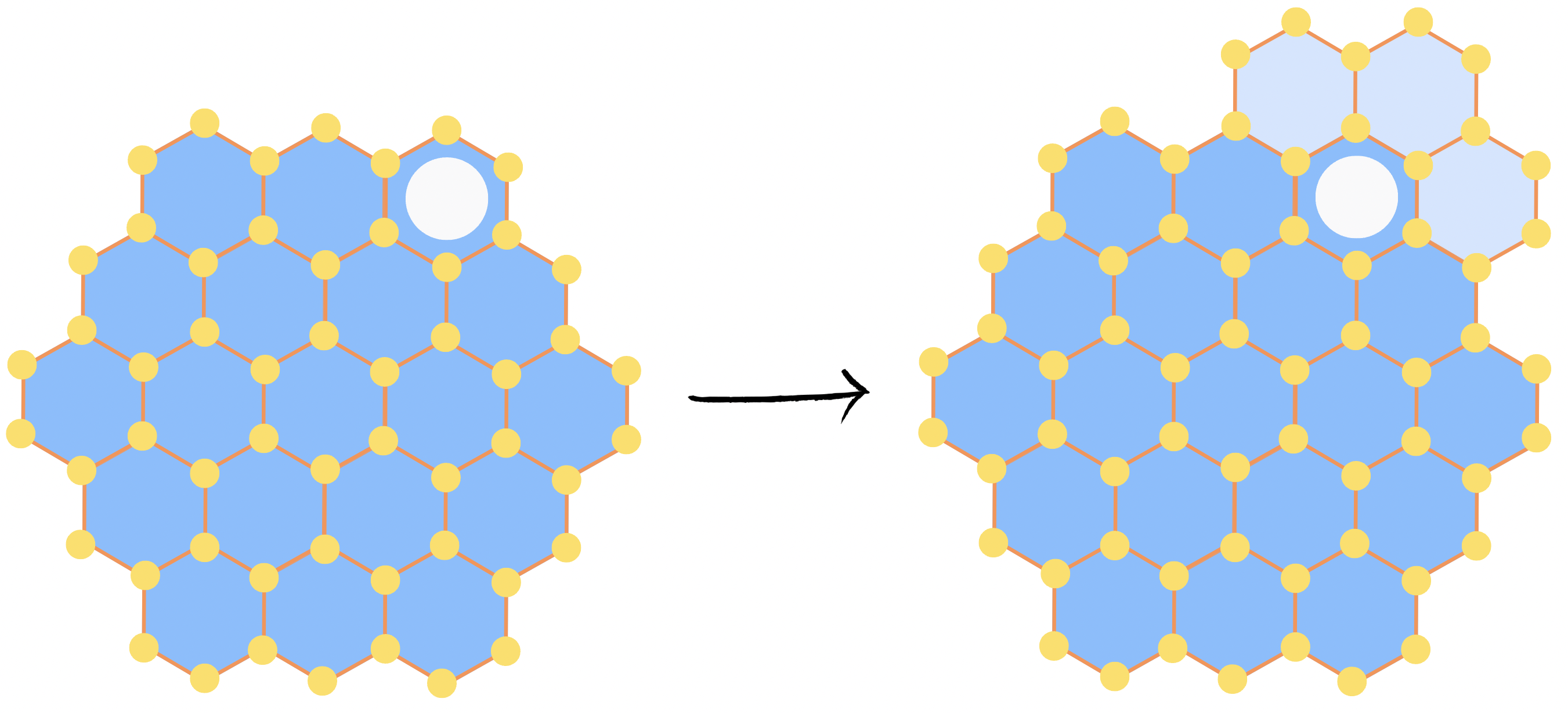}
    \captionsetup{justification=centering}
    \caption{Zone strategy when placing a tile on a hexagonal board}
    \label{fig:edgemove_hexagonal_zone}
\end{figure}

\begin{figure}[h!]
   \centering
   \includegraphics[width=0.5\linewidth]{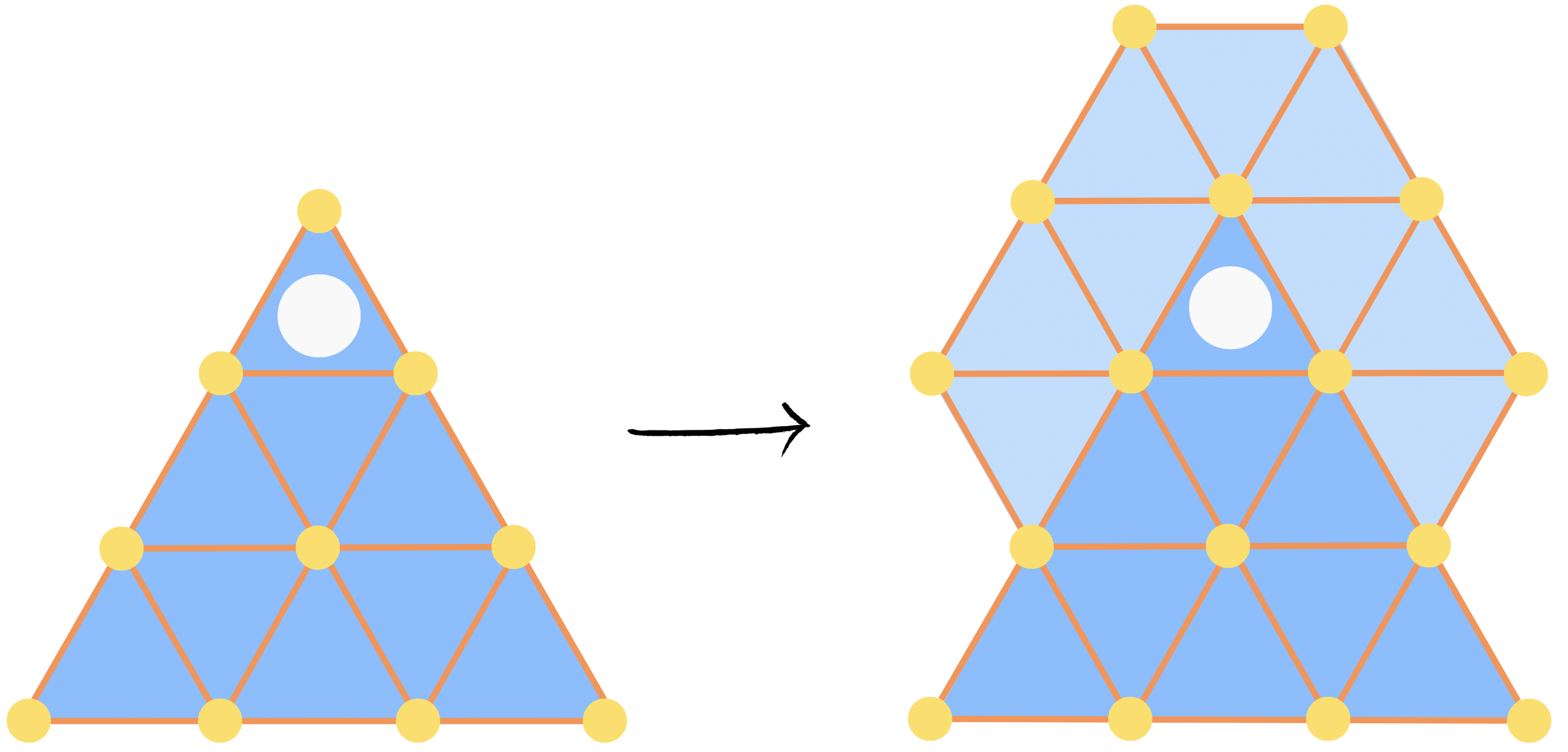}
   \captionsetup{justification=centering}
   \caption{Zone strategy when placing a tile on a triangular board}
   \label{fig:edgemove_triangular_zone}
\end{figure}

The second approach is more targeted, as it consists of adding new cells only around the area where a tile has just been placed, as shown in Figures \ref{fig:edgemove_square_zone}, \ref{fig:edgemove_hexagonal_zone}, and \ref{fig:edgemove_triangular_zone}. This strategy helps save memory and aligns better with the actual progression of the game, as the board only expands where necessary, unlike the previously discussed solution, which generates many cells that may never be used.\\

In this second approach, all cells that share at least one vertex with the cell where a tile has just been placed are ensured to exist in the board's topology. If some of these cells are already present, only the missing ones are created. The number and arrangement of cells added to the board depend directly on the type of tiles used in the game.

\subsection{Square}
In the specific case of square-shaped tiles, the system examines all eight immediately adjacent positions: the four orthogonal cells (above, below, left, right) as well as the four diagonal cells. If any of these positions lack a site, one is automatically generated and incorporated into the board, making it available for occupation in subsequent turns. This process creates the necessary cells, vertices, and edges.

\begin{figure}[h!]
    \centering
    \includegraphics[width=0.6\linewidth]{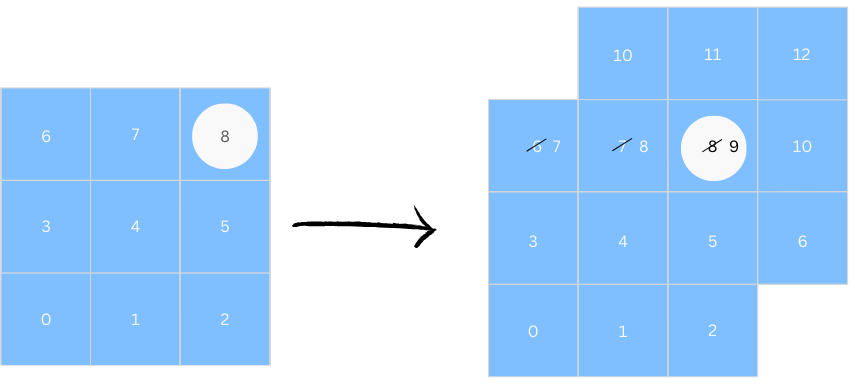}
    \captionsetup{justification=centering}
    \caption{Mapping of a growing square board - example 1}
    \label{fig:mapping_square_2}
\end{figure}

\begin{figure}[h!]
    \centering
    \includegraphics[width=0.6\linewidth]{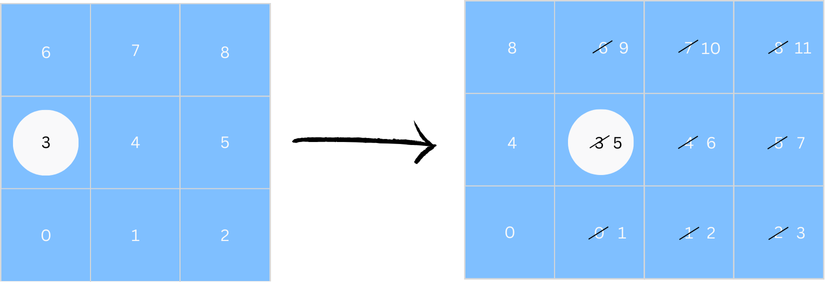}
    \captionsetup{justification=centering}
    \caption{Mapping of a growing square board - example 2}
    \label{fig:mapping_square_3}
\end{figure}

\paragraph{Mapping}
The mapping technique shown in Figures \ref{fig:mapping_square_2} and \ref{fig:mapping_square_3} applies to boards composed of square tiles, as part of the strategy for expanding a specific area of the board when a move is made on its edge.\\

In this context, this operation cannot be reduced to the systematic application of a formula for recalculating the identifiers of the cells, as the impact of this expansion varies depending on the position of the added area. The further the area is positioned to the left and at the bottom of the board, the greater the impact on the cells.\\

Indeed, the more the expansion occurs toward the lower-left corner of the board (that is, toward lower coordinate values) the more it affects the indexing of existing cells. Conversely, expansions toward the upper or right edges of the board, corresponding to higher coordinates, typically result in the addition of new cells, meaning that most existing cells are still mapped, but simply to their original indices.\\
The number of affected cells depends on the specific location where the new area is added. In Figure \ref{fig:mapping_square_2}, five new cells are introduced, whereas in Figure \ref{fig:mapping_square_3}, only three new cells are added.\\

Instead of using a simple formula to determine how previous indices should be mapped to their new counterparts, the process involves traversing the new board sequentially. Each time a newly added cell is encountered, a counter tracking the number of new cells is incremented. For every existing cell, its new index is computed by adding its previous index to the current value of this counter.\\
For instance, in Figure \ref{fig:mapping_square_3}, consider the board on the right. The traversal begins from the bottom-left cell, which is identified as a newly added cell. The counter, initially at 0, is incremented to 1. The next cell to the right corresponds to what was previously site 0; since this cell already existed, its new index is calculated as $0+1=1$. The following cell, previously site 1, is mapped to $1+1=2$, and so forth. Upon moving to the row above, another new cell is encountered, and the counter is incremented again. When reaching the cell that had the previous index 3, its new index becomes $3+2=5$, as two new cells have been added prior to it during the traversal.

\subsection{Hexagonal}
For hexagonal tiles, the system examines all six immediate neighboring positions, corresponding to the cardinal directions specific to a hexagonal grid (northeast, east, southeast, southwest, west, northwest). Thus, when a move is made at the edge of the board, only these six adjacent cells are considered in the check. Any empty position among them results in the creation of a new site, ensuring the smooth spatial development of the board and continuous accessibility for future player actions.

\begin{figure}[h!]
    \centering
    \includegraphics[width=0.7\linewidth]{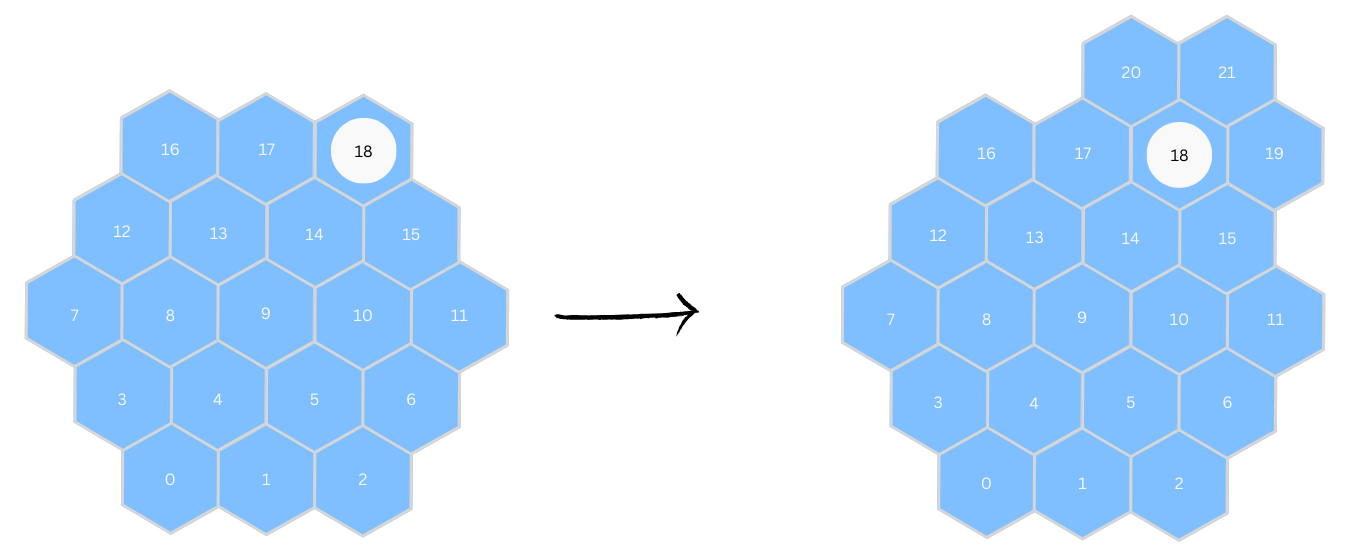}
    \captionsetup{justification=centering}
    \caption{Mapping of a growing hexagonal board - example 1}
    \label{fig:mapping_hexagonal_2}
\end{figure}

\begin{figure}[h!]
    \centering
    \includegraphics[width=0.7\linewidth]{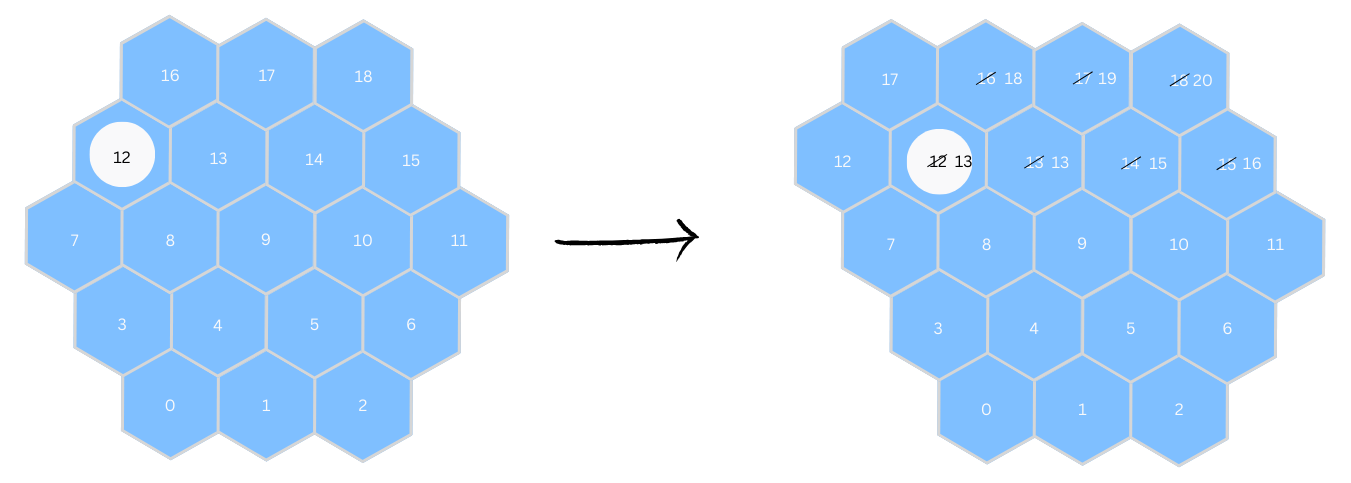}
    \captionsetup{justification=centering}
    \caption{Mapping of a growing hexagonal board - example 2}
    \label{fig:mapping_hexagonal_3}
\end{figure}

\paragraph{Mapping}
The mapping technique presented in Figures \ref{fig:mapping_hexagonal_2} and \ref{fig:mapping_hexagonal_3} applies to boards composed of hexagonal tiles.


The same principle applies to hexagonal tiles as it does to square tiles. Since the indexing of cells also depends on the direction in which the expansion occurs, newly introduced cells are tracked during a sequential traversal of the new board. A counter is incremented each time a new cell is detected, and the new index of each pre-existing cell is obtained by adding this counter value to its previous index.\\
For example, in Figure \ref{fig:mapping_hexagonal_3}, during the traversal of the new board on the right, two new cells are identified before reaching the cell that previously had index 16. As a result, this cell is now mapped to index $16+2=18$.

\subsection{Triangular}
In the case of triangular tiles, the finer and more angular topology of this configuration introduces increased complexity in managing neighbors. Indeed, each triangular cell can be in contact with up to twelve adjacent cells, due to the combination of edges contacts. Thus, when evaluating the edges, up to twelve neighboring positions are examined. Those that are not yet occupied are then generated dynamically, allowing the board to grow harmoniously.

\begin{figure}[h!]
    \centering
    \includegraphics[width=0.6\linewidth]{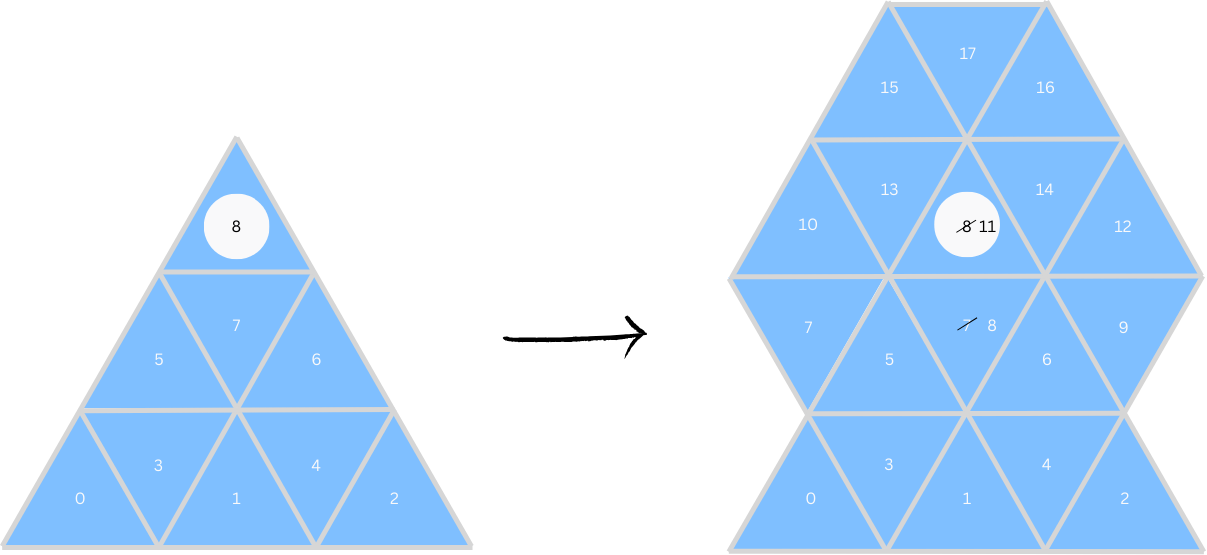}
    \captionsetup{justification=centering}
    \caption{Mapping of a growing triangular board - example 1}
    \label{fig:mapping_triangular_2}
\end{figure}

\begin{figure}[h!]
    \centering
    \includegraphics[width=0.6\linewidth]{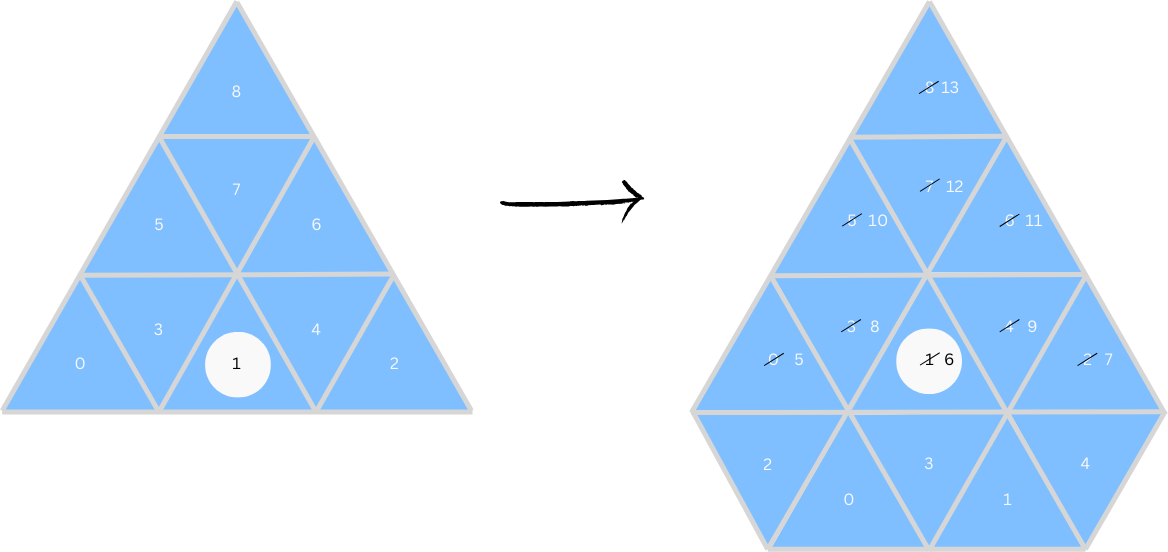}
    \captionsetup{justification=centering}
    \caption{Mapping of a growing triangular board - example 2}
    \label{fig:mapping_triangular_3}
\end{figure}

\paragraph{Mapping}
Figures \ref{fig:mapping_triangular_2} and \ref{fig:mapping_triangular_3} show the mapping technique applied to boards composed of triangular tiles.


The logic is equally applicable to triangular boards. Since the structure of the board and the indexing of its cells depend on how and where the expansion occurs, each newly added tile is identified during a linear traversal of the expanded board. A counter keeps track of how many new tiles have been inserted so far, and for every cell that existed previously, its new index is calculated by adding this count to its former index.\\
As illustrated in Figure \ref{fig:mapping_triangular_3}, five new cells are encountered before reaching the tile that had index 0 on the previous board. Accordingly, this tile is now assigned the index $0+5=5$.


\section{Move Management}
In Ludii, it is also possible to undo these moves, and this implementation was used as inspiration for the new implementation of the so-called 'forward' moves, which update the board structure when a this move is done on an edge.

\subsection{Backward}
\begin{figure}[h!]
    \centering
    \includegraphics[width=1\linewidth]{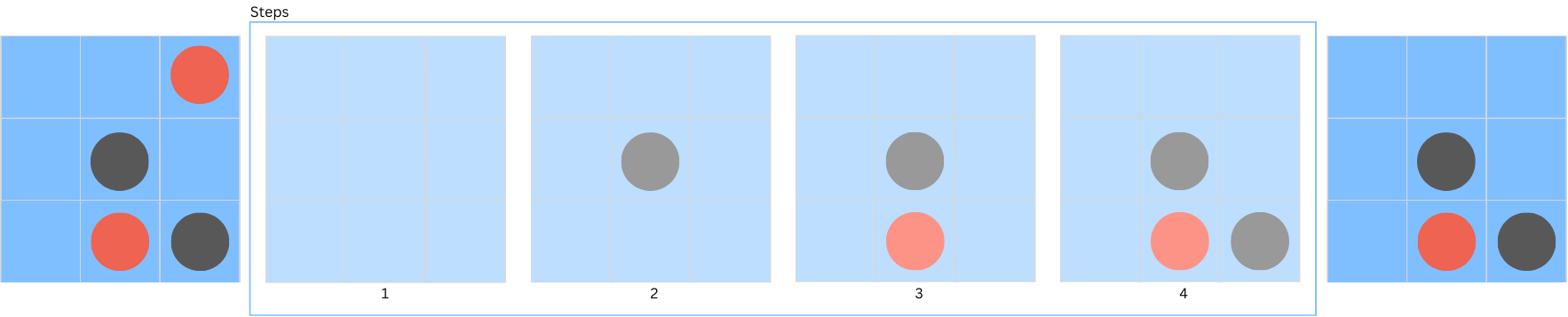}
    \captionsetup{justification=centering}
    \caption{Steps to go back from one move}
    \label{fig:going_back_1_step}
\end{figure}

When a player goes back one or more moves, the system does not simply perform a direct undo of the data structures representing the state of the board or the moves made. Instead, the game engine completely resets the state of the board, as well as all associated structures, such as those mentioned earlier, among others.\\
This approach is justified by the fact that the representation is primarily designed for AI agents that utilize a forward model rather than a backward model. Therefore, when a rollback is needed, the preferred method is to replay all moves from the beginning of the game, as demonstrated in Figure \ref{fig:going_back_1_step}, until the state just before the move to be undone is reached. This process ensures that all data structures are updated consistently, accurately reflecting the state of the game at that moment.

Thus, in the context of the new implementation, when a player goes back one move, if that move had previously resulted in an expansion of the board through the addition of new cells, this expansion will simply not be reproduced when replaying the moves from the initial state. The board will then be in a reduced state, precisely corresponding to its configuration before the last move.

\subsection{Forward}
Two approaches were explored for managing 'forward' moves (i.e., the moves played in the usual manner, one after the other, as opposed to undone moves, which involve undoing a previously played move).\\

\begin{figure}[h!]
    \centering
    \includegraphics[width=0.8\linewidth]{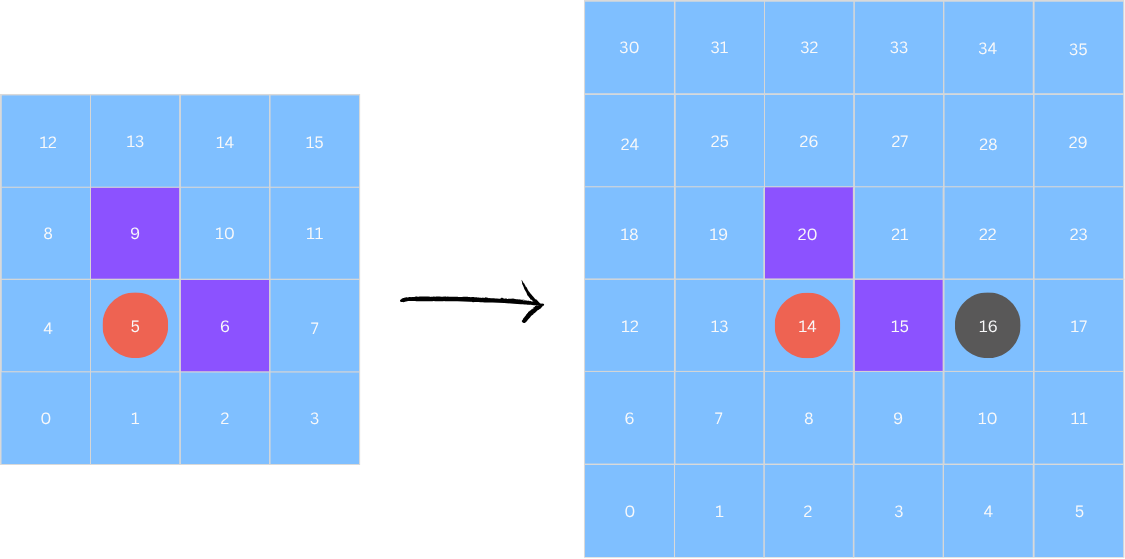}
    \captionsetup{justification=centering}
    \caption{Example of a board expansion using perimeter strategy}
    \label{fig:example_chunks}
\end{figure}

Figure \ref{fig:example_chunks} serves to illustrate the following points more clearly. In this figure, the purple tiles represent the initial tiles placed at the beginning of the game, on which players are not allowed to play. The red chip was placed by player 1, while the grey chip, played by player 2, triggered an expansion of the board.

\subsubsection{Backward inspired approach}
\begin{figure}[h!]
    \centering
    \includegraphics[width=1\linewidth]{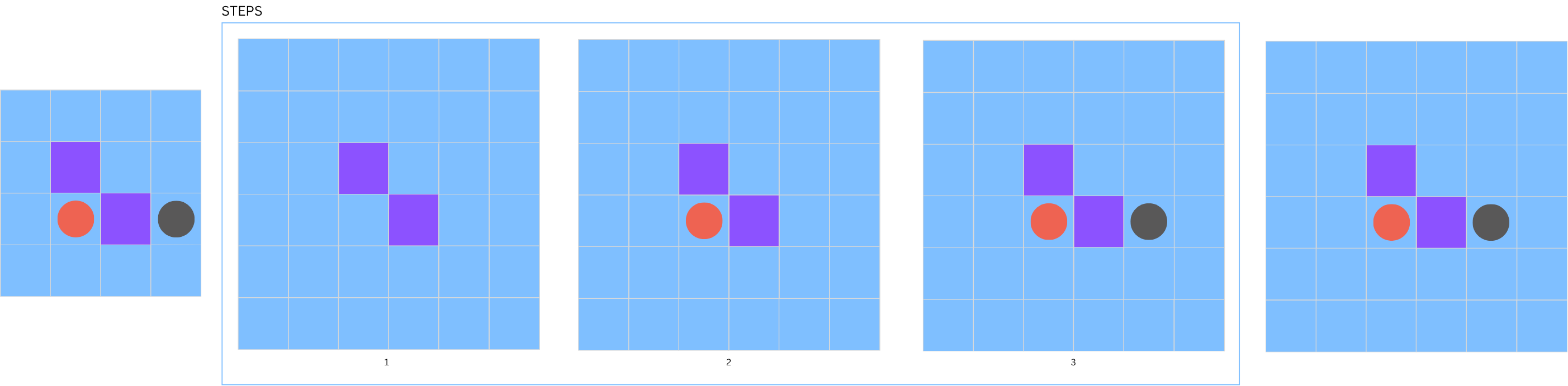}
    \captionsetup{justification=centering}
    \caption{Adding a move on an edge and reapplying all previous moves}
    \label{fig:mapping_init_to_end}
\end{figure}

The first approach is inspired by the method used for managing undone moves: it consists of resetting the board (i.e., completely clearing it), then expanding it to a larger size than the last board size. Once this reset is done, each move is reapplied one by one on the new board, so that the moves are mapped and placed onto this expanded board, as shown in Figure \ref{fig:mapping_init_to_end}.

In the current implementation, when a player undoes a move, the board is reset to the initial size defined at the start of the game. By 'reset', it should be noted that the ContainerState is completely reset. This means that the associated chunks, such as \textit{empty}, are also reset, and all cells are marked as empty, except for those with initial tiles placed before the players began the game.\\

Once this reset is performed, it is first necessary to update the game’s visual by recalculating the new graph, and, at the same time, the topology. The graph is then resized to match the new size of the board, which is now larger than the previous one, in order to account for the board expansion.\\

Following the expansion of the board, it is necessary to update the container state by incorporating the newly added sites that were absent from the initial board configuration. These new sites are, by default, unoccupied and are therefore added to the empty chunk. Since the container state is first reset, it reflects the state of the original board, which is typically mostly empty. However, if the initial board contains initial tiles, these sites will not be included in the \textit{empty} chunkset, as they will instead appear in the \textit{what}, \textit{who}, and \textit{count} chunksets.\\

Consider Figure \ref{fig:example_chunks} as an illustrative example. On the left side of the figure, the \textit{empty} chunk includes all sites except those that contain initial tiles, specifically, sites 6 and 9. Site 5 is considered empty because the red chip was added after the start of the game; thus, when the container state was reset, it did not reflect this later player move.\\
When a player makes a move that causes the board to expand (as shown on the right side of the figure), the container state is updated. This involves adding all newly created sites, namely, sites 0, 1, 2, 3, 4, 5, 6, 11, 12, 17, 18, 23, 24, 29, 30, 31, 32, 33, 34, and 35, to the \textit{empty} chunkset. However, the previous inclusion or exclusion of certain sites in this chunkset may no longer be accurate. For instance, sites 6 and 9, which were not considered empty on the original board, will now be treated as empty in the expanded configuration.\\
To handle this correctly, a remapping process is performed for each value inside the chunksets. This involves associating each site from the original board with its corresponding site in the expanded board. For example, site 0 on the original board maps to site 7 in the expanded configuration. As site 0 belonged to the \textit{empty} chunkset, site 7 will also be included in this chunkset, replacing site 0. Conversely, site 9, which was not part of the \textit{empty} chunkset, maps to site 20, which will now be excluded from this chunkset.\\
This mapping is executed prior to the addition of new sites resulting from the expansion. Doing so ensures a clear distinction between pre-existing mapped values and newly introduced ones, thereby avoiding any overlap or inconsistency in chunkset assignments.\\

Once the reset and the update of the chunks are completed, each move played since the start of the game is \textbf{reapplied} one by one on the new board. During this process, internal structures such as the container state and owned are progressively updated by each move being replayed. Additionally, before their reapplication, the \textit{from} and \textit{to} attributes of each move are mapped to the new coordinates of the expanded board, ensuring perfect consistency with the new topology.\\

It is important to note that the last move played was initially applied to the 'previous' board, triggering the board expansion. It will therefore be reapplied a second time, but this time on the new, larger board.

\subsubsection{Logical approach}
\begin{figure}[h!]
    \centering
    \includegraphics[width=0.55\linewidth]{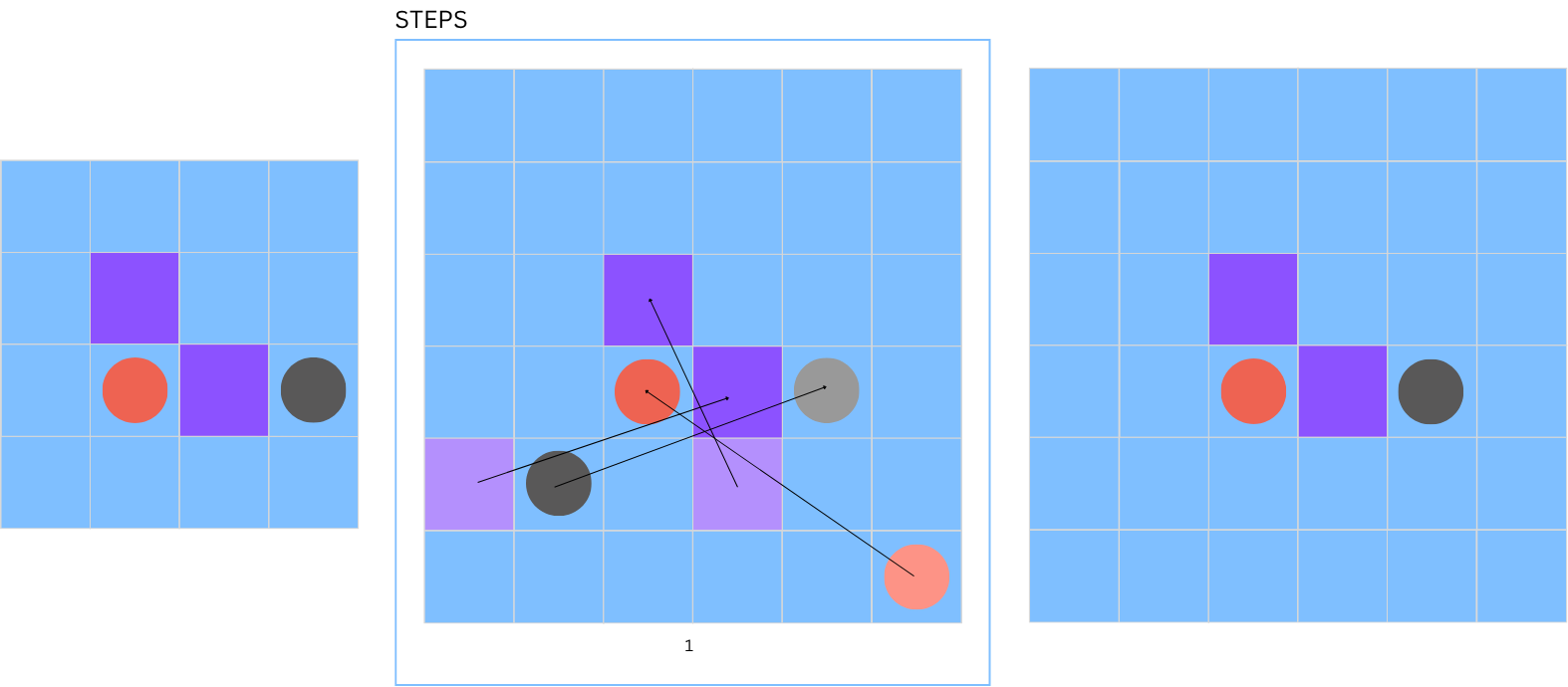}
    \captionsetup{justification=centering}
    \caption{Adding a move on an edge and mapping current game state to new one}
    \label{fig:mapping_current_to_end}
\end{figure}

The second approach, which seems more intuitive, simply involves taking the current board with all the moves played, and then mapping it onto a larger board, as shown in Figure \ref{fig:mapping_current_to_end}.


First and foremost, it is necessary to update the graph representing the board, so that the visual representation correctly reflects the new dimensions. This involves reconstructing the graph based on the new size of the board, and consequently, updating the Topology structure.\\

Once the graph and its topology have been correctly updated, the chunks within the ContainerState are remapped to align with the new board configuration. Elements that were not present on the previous board are incorporated into the empty chunkset, while existing cells are remapped in order to preserve their current state.\\
In contrast to the backward inspired strategy, the container state is not reset in this case. As a result, most sites may no longer belong to the \textit{empty} chunkset, depending on the tiles that have been placed since the beginning of the game.\\ 
As illustrated in Figure \ref{fig:example_chunks}, red chips are now taken into account. Consequently, site 5 is no longer part of the \textit{empty} chunkset. Since site 5 maps to site 14 in the expanded configuration, site 14 will then be excluded from the \textit{empty} chunkset.\\

It is important to note that the containers of other components, such as the players’ hands or the decks, do not require any updates. In fact, the expansion of the board only impacts the container specific to it; the other containers maintain their state and structure unchanged, as indexing is local and not global.\\

Ownership information must also be remapped using the owned structure to accurately reflect the positions and possessions of each player, including the "player" representing the board itself. This entity holds initially placed components as well as shared elements, which may occur in certain game configurations, among others.\\
As illustrated in the first board of Figure \ref{fig:mapping_current_to_end}, a red chip is placed on site 5 and a grey chip on site 10. Accordingly, the owned structure records that the board owns components on sites 6 and 9, player one owns the component on site 5, and player two owns the component on site 10. When the board expands, this ownership data is not reset and must therefore be remapped to the corresponding sites on the new board. This process involves iterating over each owned component and assigning it to its new index. In this example, site 5 maps to site 14, site 6 to 15, site 9 to 20, and site 10 to 21.\\

Unlike the previously mentioned method, the previous moves are not reapplied one by one. Their \textit{from} and \textit{to} attributes are mapped to match the new coordinates in the expanded board. This ensures that the move history remains valid without needing to replay all the moves.\\

Finally, once all the structures are adjusted, the last move (the one that triggered the latest expansion of the board) has its from and to attributes mapped according to the new configuration, and it is then applied to the now updated structure.\\
Thus, when a player makes a move, the system checks whether the move is made on a border of the board; if so, the move triggers the necessary update of internal structures (topology, container state, owned) to ensure integrity and consistency with the new board, before being ultimately applied to it.


\section{Initial board size}
Another important aspect is determining the initial board size. Indeed, hardcoding the same initial size for all games is not sufficient. The presence of initial tiles must be taken into account. Since the board graph is constructed before the game compilation is completed, it is difficult to determine how the initial tiles are positioned relative to one another. At this stage of the thesis, the chosen approach is to simply count the number of initial tiles and assume they are placed in a straight line, centered on the board. Initial size will different from one shape tile board to another, based on the initial numbers of tiles. All these technical choices were made to ensure that at least all the neighbors (cells) of the initial tiles are present on the board.\\

In the absence of initial tiles, the initial board size is set as follows: 3 for square tiles, 2 for hexagonal tiles, and 4 for triangular tiles. This choice was made because certain games rely on the central position of the board to determine which directions are accessible, based on the shape of the tile and its surrounding neighbors. Reducing the board to a size of 1 introduces complications, especially when the first tile to be placed can be rotated. In such cases, valid directions play a role in determining the possible orientation of the tile, and a minimal board size would not provide sufficient spatial context to support these decisions.

\chapter{Experiments}
\label{experiments}


This chapter presents an experimental evaluation of the techniques developed in this work, with the primary goal of assessing their effectiveness compared to the current implementation in Ludii. The existing implementation enforces board size limits, restricted to fixed dimensions, resulting in an incomplete representation of boardless games. In contrast, the new techniques enable a more correct boardless game modeling, eliminating arbitrary size constraints and allowing for a more accurate and faithful reproduction of the game environment.\\
The following sections detail the experimental setup, including hardware specifications, test games, and experimental protocols. Subsequently, results are presented and analyzed to assess the comparative performance of each technique.

\section{Experimental Setup}
This section presents the experimental setup, including the different techniques to be compared, the methodology used for their evaluation, and the hardware on which the tests were conducted.
\subsection{Compared Techniques}
\label{sec:techniques}

In the following comparison, five different implementation techniques for boardless games in Ludii are evaluated. The baseline implementation corresponds to the original approach provided by Ludii, and is referred to as \textbf{BASE} in the experiments. In this version, the board is static and does not expand during gameplay. Its size is fixed at the beginning of the game; a side length of 41 for square and triangular tiles, and 21 for hexagonal tiles. In addition to this baseline, four alternative implementations have been developed, categorized into two primary strategies for board expansion:
\begin{itemize}
    \item \textit{Perimeter-based expansion}: whenever a player places a tile on the edge of the current board, a new layer of cells is appended around the entire perimeter, effectively increasing the board’s dimensions uniformly in all directions.
    \item \textit{Zone expansion}: in contrast, this method only extends the board locally, by adding new tiles adjacent to the specific location where a player has just placed a tile on the boundary. This leads to a more targeted and minimal board growth.
\end{itemize}

Each of these two strategies has been implemented in two distinct ways, depending on how the system reconstructs de game state following a move played on the board’s edge:
\begin{itemize}
    \item \textit{Reconstruction via move reapplication}: the board is reset, and all previous moves are reapplied sequentially on the newly resized board. This approach mirrors Ludii’s native mechanism for undoing moves.
    \item \textit{Direct mapping without reapplication}: Instead of replaying the move history, the existing game state is directly mapped onto the new board structure, adjusting the spatial representation without altering the game’s progression.
\end{itemize}

The combination of these strategies yields four distinct variants:
\begin{enumerate}
    \item \textbf{PERI-RE}: Perimeter-based expansion with move reapplication 
    \item \textbf{PERI-MAP}: Perimeter-based expansion with direct mapping 
    \item \textbf{ZONE-RE}: Zone expansion with move reapplication
    \item \textbf{ZONE-MAP}: Zone expansion with direct mapping
\end{enumerate}
These five implementations (BASE, PERI-RE, PERI-MAP, ZONE-RE, and ZONE-MAP) form the basis for the comparative experimental analysis presented in the next section.

\subsection{Experimental Protocol}
\subsubsection{Size of a board}
To evaluate the impact of the new approaches for board expansion strategies for boardless games, a benchmark was established based on simulated game progressions up to 200 move placements. This upper bound was selected after analyzing several representative games within the genre. For instance, several well-known modern board games serve as illustrative examples for estimating an upper limit on the number of tile placements.\\
The game \textit{Carcassonne}~\cite{enwiki:1291981883}, for example, includes at least 72 tiles in its base game, and with expansions, the total number of tiles can reach up to 200.\\
\textit{Dorfromantik}~\cite{dorfromantik}, another popular contemporary title, features 104 tiles by default, with the possibility of additional tiles via expansions.\\
\textit{Kingdomino}~\cite{kingdomino}, while smaller in scale, consistently completes within 48 tile placements.\\
These examples reflect a broad range of gameplay durations across modern tile-based board games. Consequently, a ceiling of 200 moves was deemed sufficient to encompass the upper bounds of typical gameplay while allowing for comparative analysis across a wide range of possible scenarios.\\

The evaluation was based on two key plots: one tracking board size relative to the cumulative number of moves since the beginning of the game, and another measuring the percentage of unused tiles (those generated but never occupied), providing an indicator of spatial efficiency.\\

\begin{figure}[h!]
    \centering
    \begin{minipage}{0.48\textwidth}
        \centering
        \includegraphics[scale=0.13]{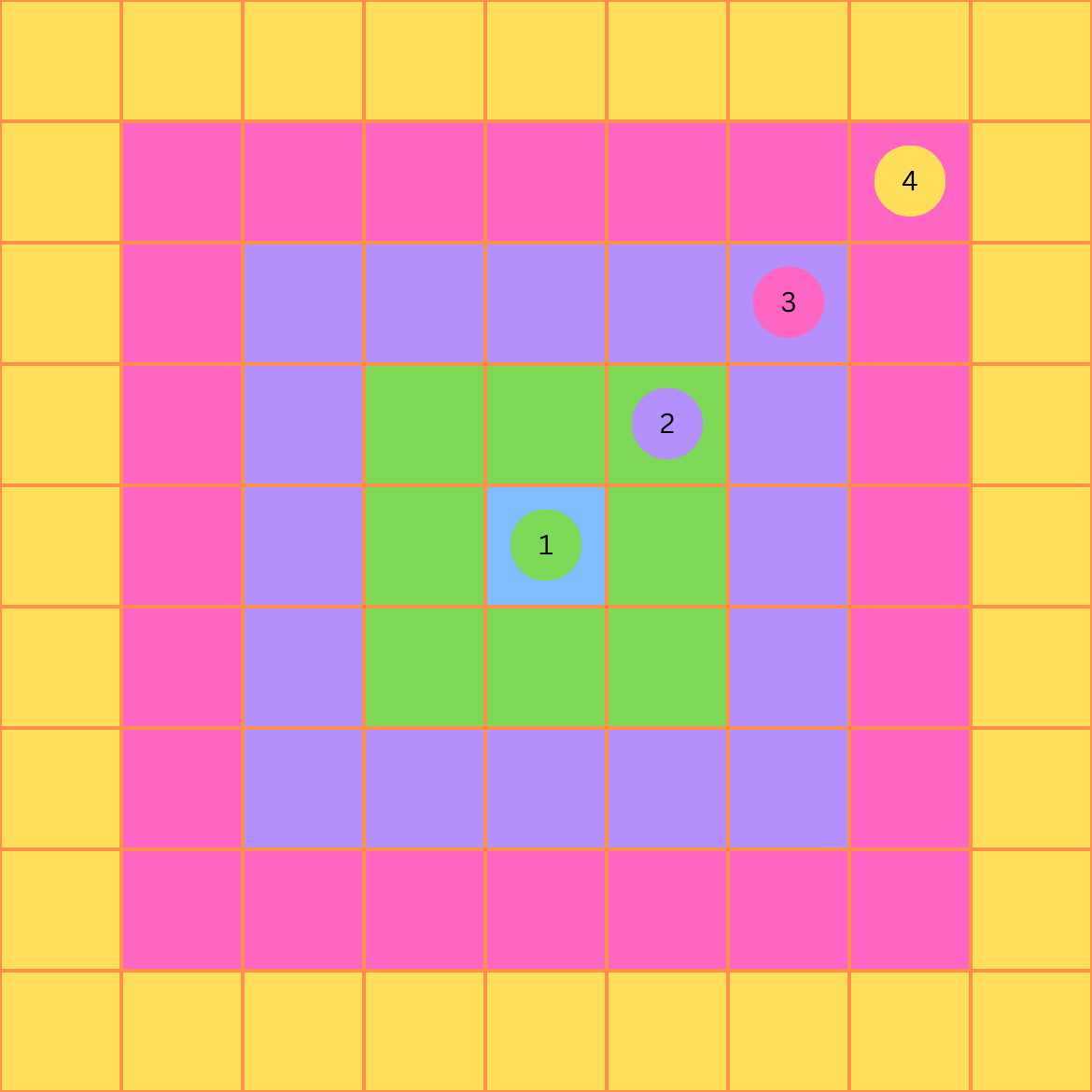}
        \captionsetup{justification=centering}
        \caption*{(a) Worst case}
    \end{minipage}
    \hfill
    \begin{minipage}{0.48\textwidth}
        \centering
        \includegraphics[scale=0.17]{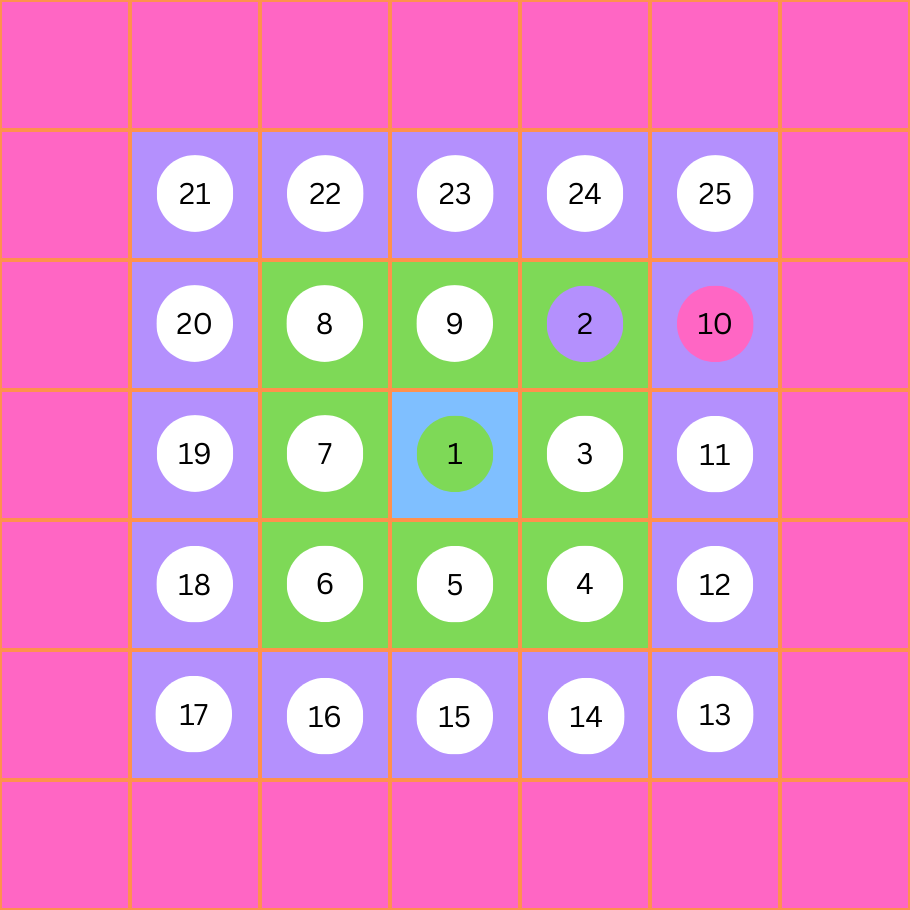}
        \captionsetup{justification=centering}
        \caption*{(b) Best case}
    \end{minipage}
    
    \captionsetup{justification=centering}
    \caption{Evolution of the board size perimeter strategy - Square tiles}
    \label{fig:board_size_contour_square}
\end{figure}

\begin{figure}[h!]
    \centering
    \begin{minipage}{0.48\textwidth}
        \centering
        \includegraphics[scale=0.2]{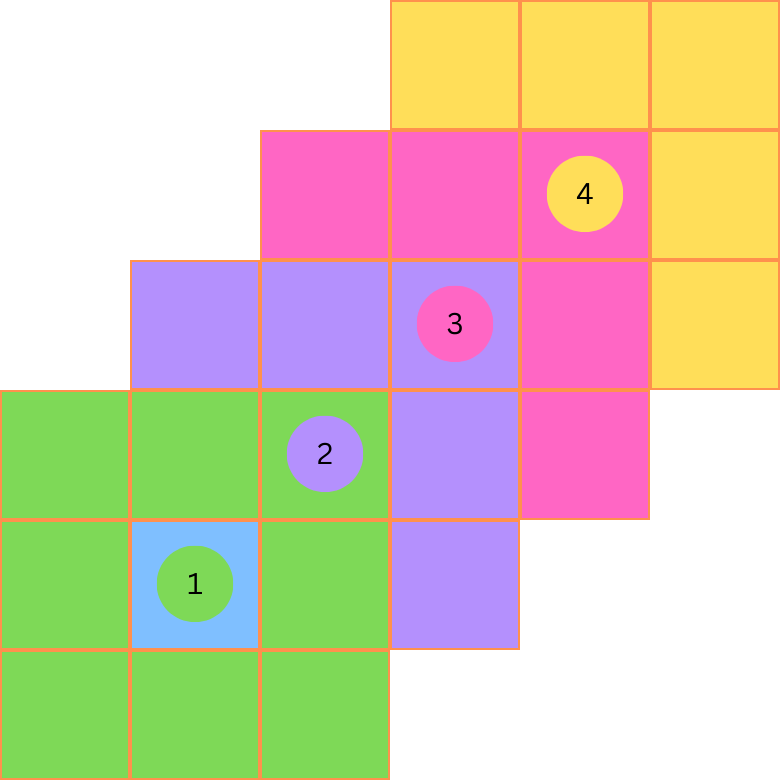}
        \captionsetup{justification=centering}
        \caption*{(a) Worst case}
    \end{minipage}
    \hfill
    \begin{minipage}{0.48\textwidth}
        \centering
        \includegraphics[scale=0.2]{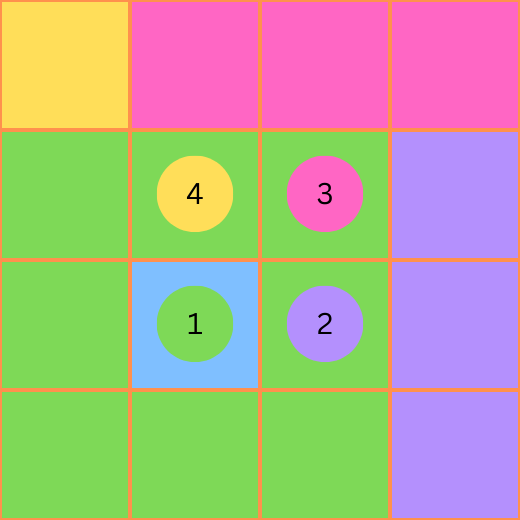}
        \captionsetup{justification=centering}
        \caption*{(b) Best case}
    \end{minipage}
    
    \captionsetup{justification=centering}
    \caption{Evolution of the board size zone strategy - Square tiles}
    \label{fig:board_size_zone_square}
\end{figure}

\begin{figure}[h!]
    \centering
    \begin{minipage}{0.48\textwidth}
        \centering
        \includegraphics[scale=0.15]{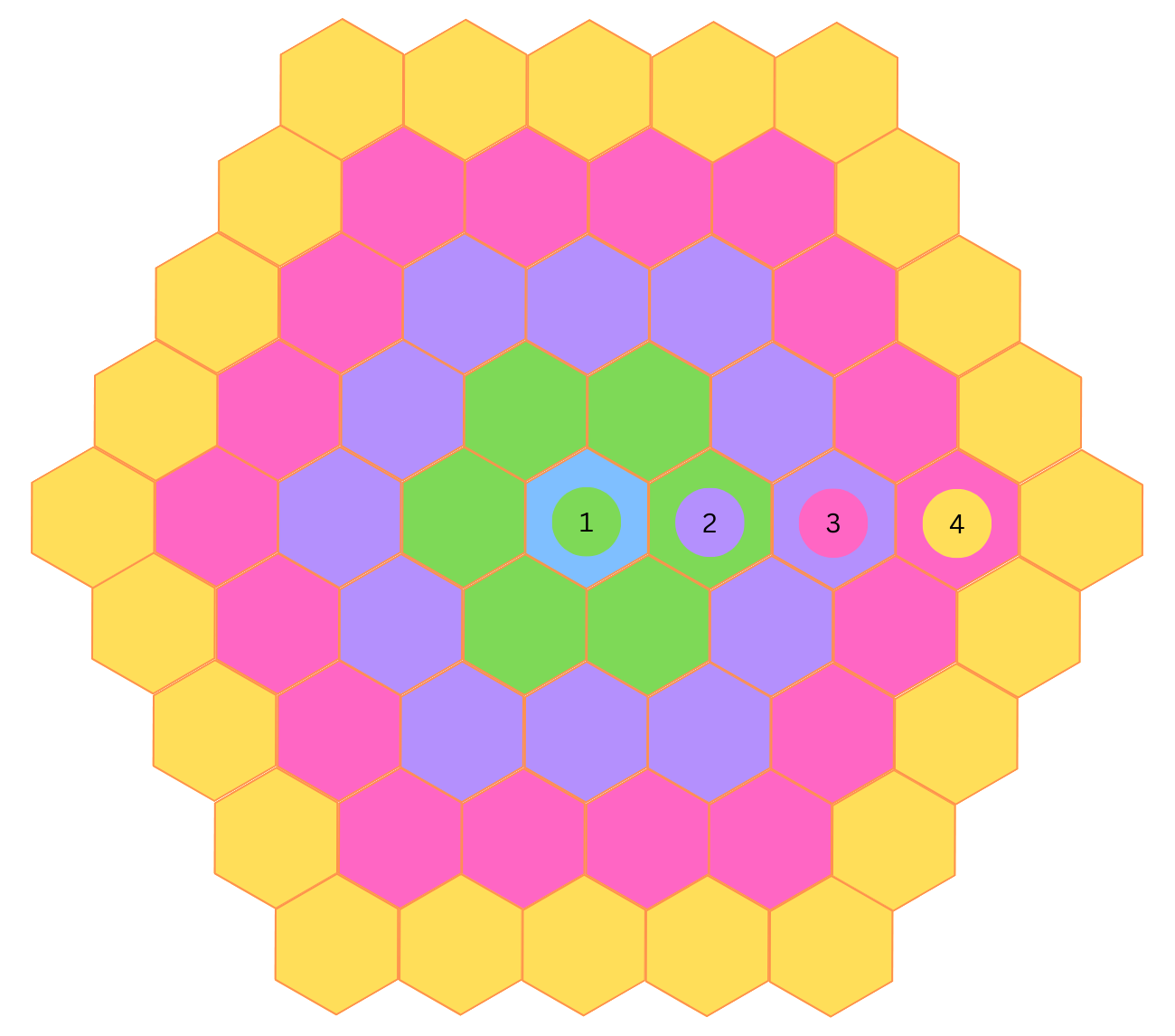}
        \captionsetup{justification=centering}
        \caption*{(a) Worst case}
    \end{minipage}
    \hfill
    \begin{minipage}{0.48\textwidth}
        \centering
        \includegraphics[scale=0.19]{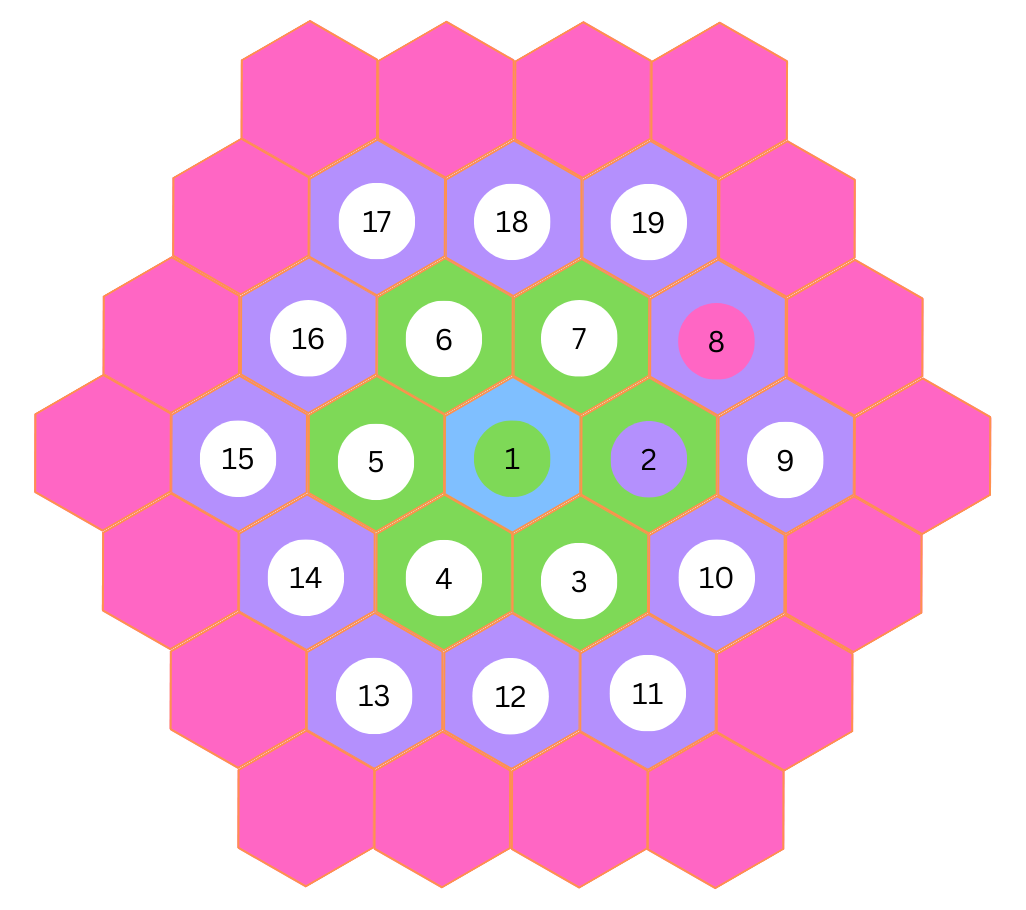}
        \captionsetup{justification=centering}
        \caption*{(b) Best case}
    \end{minipage}
    
    \captionsetup{justification=centering}
    \caption{Evolution of the board size perimeter strategy - Hexagonal tiles}
    \label{fig:board_size_contour_hexagonal}
\end{figure}

\begin{figure}[h!]
    \centering
    \begin{minipage}{0.48\textwidth}
        \centering
        \includegraphics[scale=0.2]{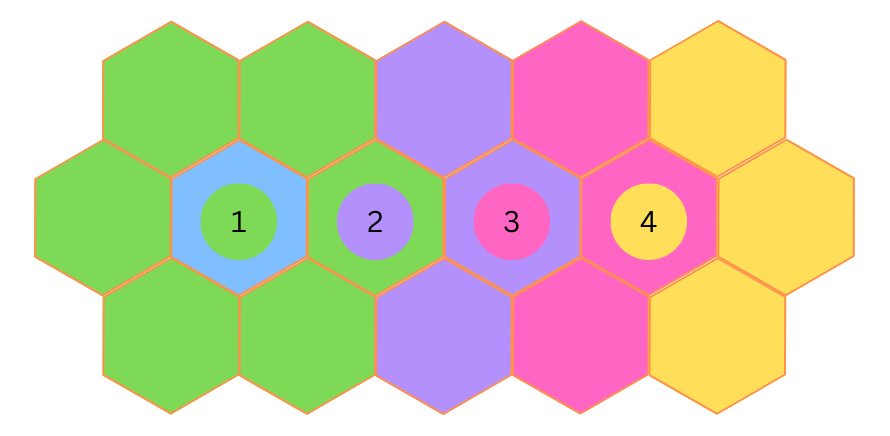}
        \captionsetup{justification=centering}
        \caption*{(a) Worst case}
    \end{minipage}
    \hfill
    \begin{minipage}{0.48\textwidth}
        \centering
        \includegraphics[scale=0.2]{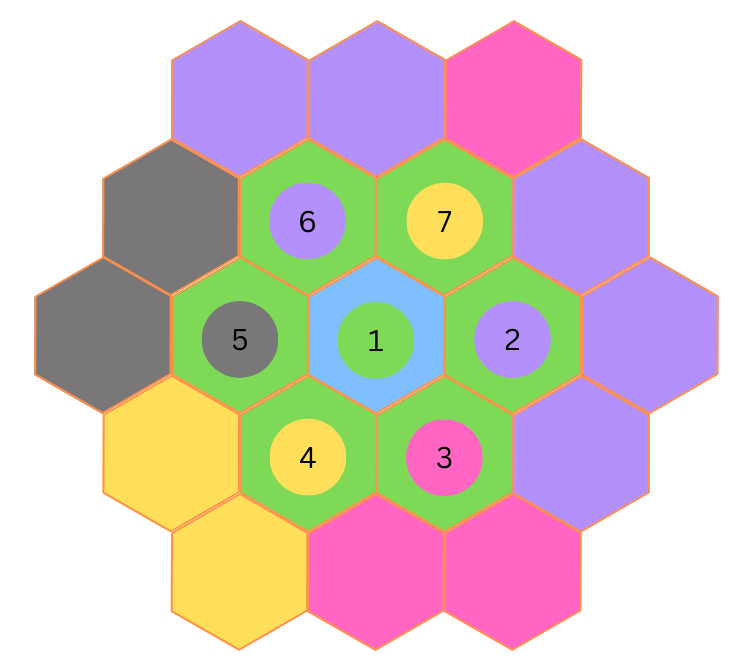}
        \captionsetup{justification=centering}
        \caption*{(b) Best case}
    \end{minipage}
    
    \captionsetup{justification=centering}
    \caption{Evolution of the board size zone strategy - Hexagonal tiles}
    \label{fig:board_size_zone_hexagonal}
\end{figure}

\begin{figure}[h!]
    \centering
    \begin{minipage}{0.48\textwidth}
        \centering
        \includegraphics[scale=0.1]{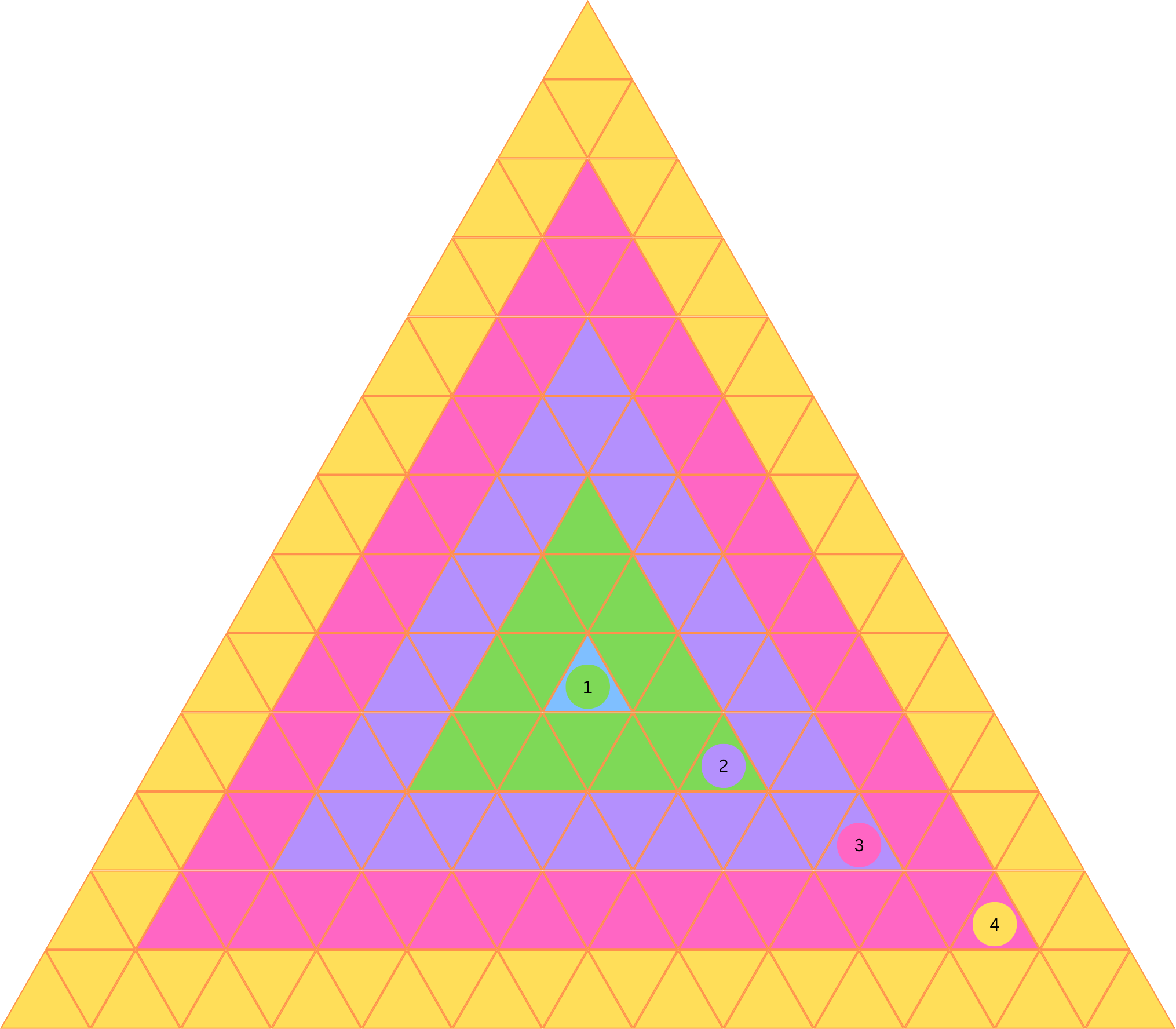}
        \captionsetup{justification=centering}
        \caption*{(a) Worst case}
    \end{minipage}
    \hfill
    \begin{minipage}{0.48\textwidth}
        \centering
        \includegraphics[scale=0.13]{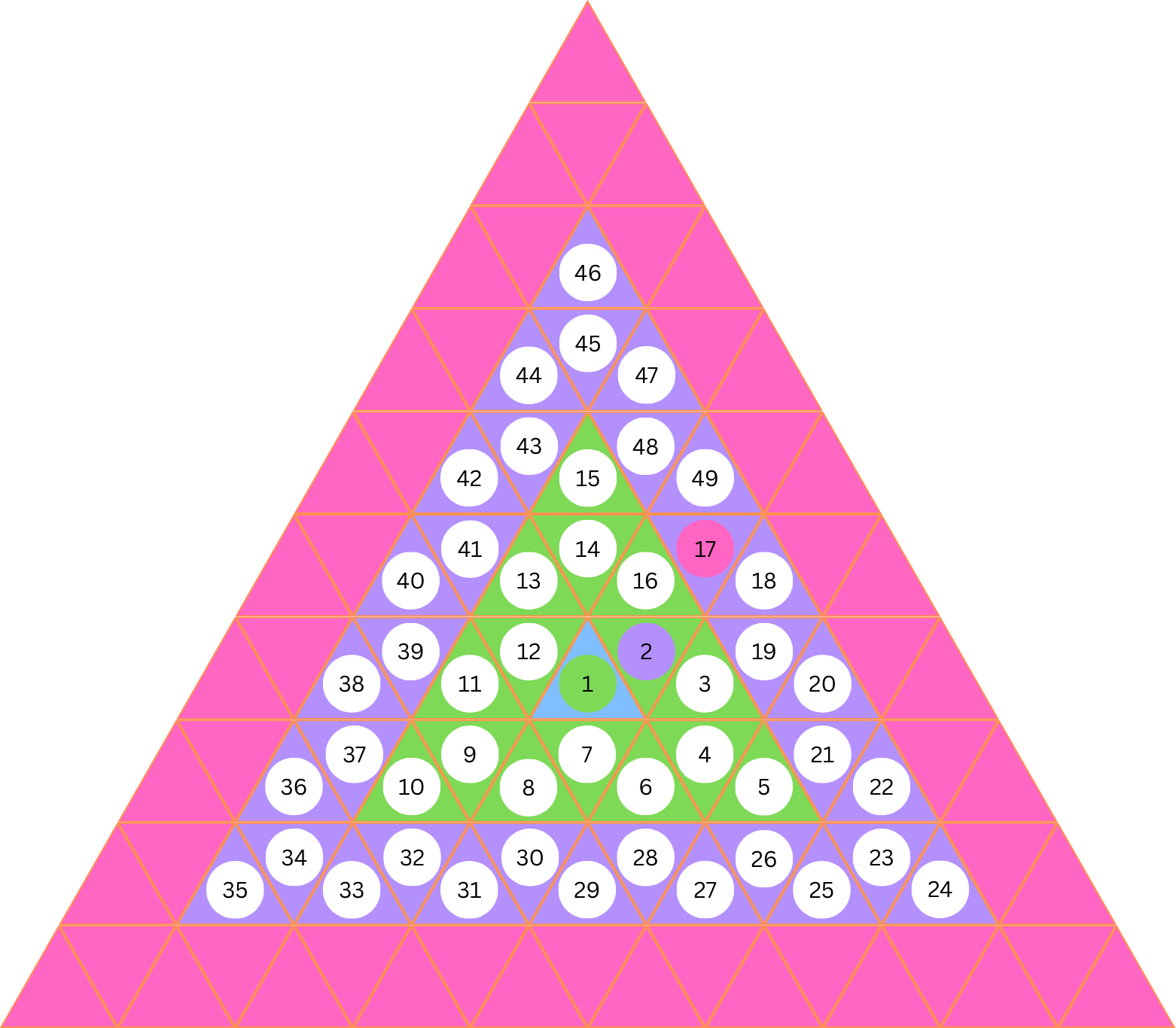}
        \captionsetup{justification=centering}
        \caption*{(b) Best case}
    \end{minipage}
    
    \captionsetup{justification=centering}
    \caption{Evolution of the board size perimeter strategy - Triangular tiles}
    \label{fig:board_size_contour_triangular}
\end{figure}

\begin{figure}[h!]
    \centering
    \begin{minipage}{0.48\textwidth}
        \centering
        \includegraphics[scale=0.2]{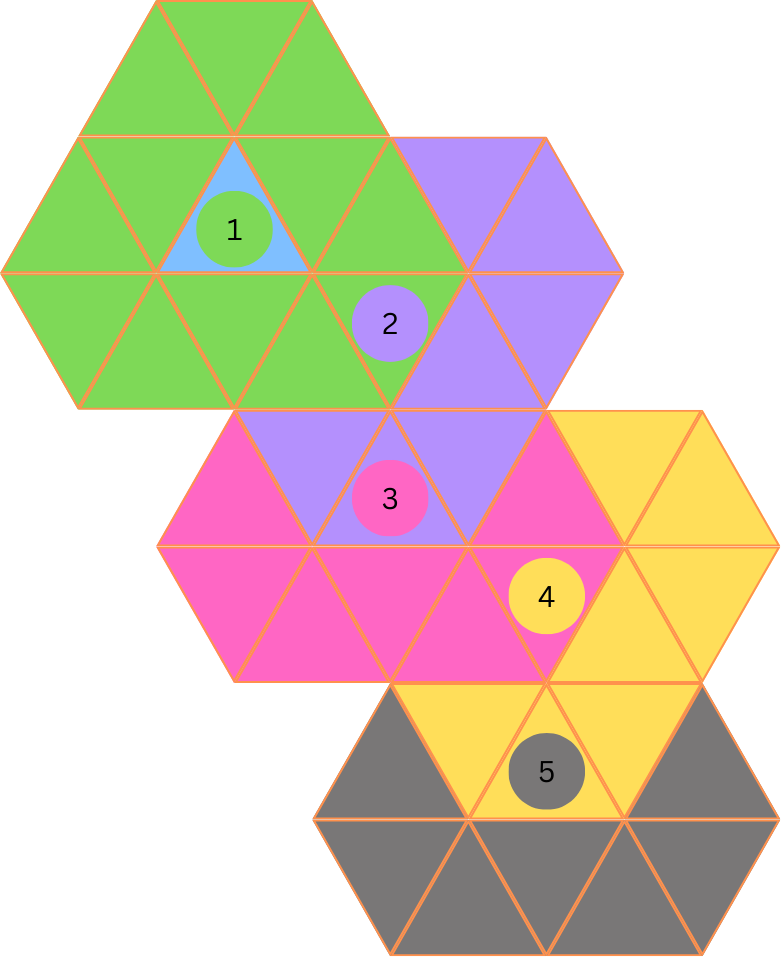}
        \captionsetup{justification=centering}
        \caption*{(a) Worst case}
    \end{minipage}
    \hfill
    \begin{minipage}{0.48\textwidth}
        \centering
        \includegraphics[scale=0.2]{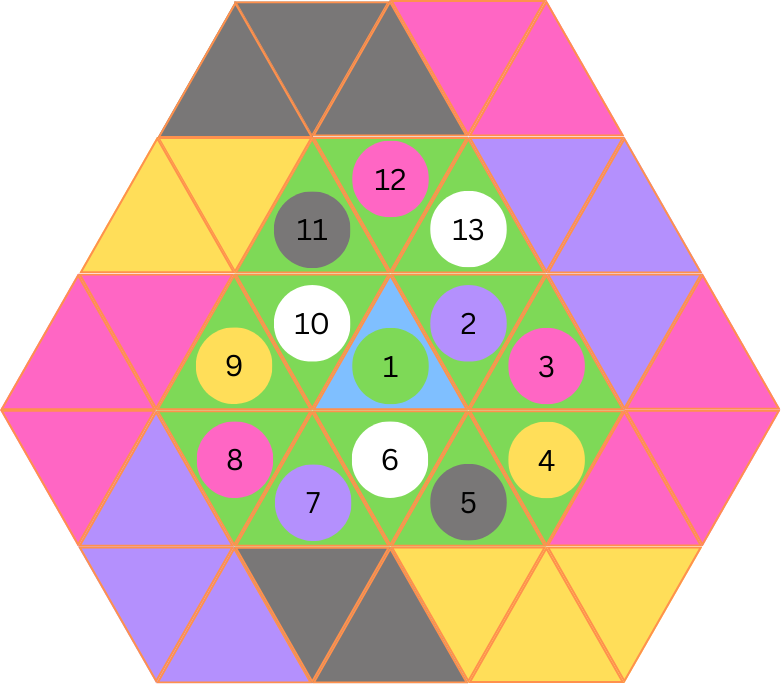}
        \captionsetup{justification=centering}
        \caption*{(b) Best case}
    \end{minipage}
    
    \captionsetup{justification=centering}
    \caption{Evolution of the board size zone strategy - Triangular tiles}
    \label{fig:board_size_zone_triangular}
\end{figure}

In order to assess the efficiency and scalability of the proposed board expansion strategies, simulations were conducted under two contrasting scenarios: a worst-case and a best-case configuration. The worst-case scenario corresponds to a sequence of moves that maximizes the number of new tiles generated at each turn, thereby stressing the system's spatial and computational resources. This scenario represents the most extreme case, causing the board to grow as much as possible with every move.\\
In contrast, the best-case scenario aims to minimize the number of tiles added, promoting a concentric and compact board expansion. This represents the most ideal growth pattern, as it minimizes spatial spread.\\
In Figures \ref{fig:board_size_contour_square}, \ref{fig:board_size_zone_square}, \ref{fig:board_size_contour_hexagonal}, \ref{fig:board_size_zone_hexagonal}, \ref{fig:board_size_contour_triangular} and \ref{fig:board_size_zone_triangular}, each move is represented by a numbered chip. Colored chips indicate that the move triggered the creation of new tiles of the same color, contributing to board expansion. White chips denote moves that resulted in no expansion. The number on each chip indicates the move order.\\

In both scenarios, the board is assumed to initially contain no tiles. Its size is thus considered to be one, representing the first and only location where a tile can be placed at the beginning of the game.

\paragraph{Square}
For square tiles, the worst-case scenario under the perimeter-based expansion strategy is trivial to reproduce: it consists of playing on any edge of the board at every turn, thereby triggering the addition of a full border to the board. This behavior is illustrated on the left side of Figure \ref{fig:board_size_contour_square}, where the board reaches a total of 81 tiles after only 4 moves.\\
In the case of the local zone expansion strategy, the position yielding the highest tile generation is a corner placement, which introduces five new adjacent tiles. This scenario is depicted on the left of Figure \ref{fig:board_size_zone_square}, with the board expanding to 24 tiles after 4 moves.

Conversely, the best-case scenario aims to minimize board growth by playing each new move adjacent to previously made ones, following a compact spiral pattern.\\
Under the perimeter expansion strategy, this approach results in a board of only 49 tiles after 25 moves, as shown on the right side of Figure \ref{fig:board_size_contour_square}, compared to 2,601 tiles in the worst-case scenario for the same number of moves.\\
Similarly, the right side of Figure \ref{fig:board_size_zone_square} shows that, under the local zone strategy, the board grows to just 16 tiles after 4 moves.

\paragraph{Hexagonal}
With hexagonal tiles, the worst-case scenario for the perimeter-based strategy is consistent with that observed for square tiles: any move placed on the board’s edge causes a full border expansion. This is shown on the left of Figure \ref{fig:board_size_contour_hexagonal}, where the board reaches 61 tiles after 4 moves.\\
As for the local zone expansion strategy, the worst-case behavior occurs when moves are made along a straight line in a fixed direction. This pattern ensures that each placement lies on the boundary and consequently triggers the addition of three new adjacent tiles. This behavior is illustrated on the left side of Figure \ref{fig:board_size_zone_hexagonal}, where the board reaches a size of 16 tiles after 4 moves.

The best-case configuration follows the same principle as with square tiles, ensuring that each move is done as close as possible to the existing structure. This is possibly achieved using a spiral growth pattern, which minimizes the number of newly generated tiles.\\
For the perimeter strategy, this compact placement yields a board of 37 tiles after 19 moves, as illustrated on the right of Figure \ref{fig:board_size_contour_hexagonal}, whereas the same number of moves would result in 1,141 tiles in the worst-case scenario.\\
In the case of the local zone strategy, this technique leads to a board of only 19 tiles after 7 moves, as illustrated on the right of Figure \ref{fig:board_size_zone_hexagonal}, whereas 7 moves played on the worst case scenario would have produced a board of 22 tiles.

\paragraph{Triangular}
In the triangular tile configuration, the worst-case scenario for the perimeter-based expansion strategy occurs when moves are made on the boundary of the board, regardless of direction, as any edge placement triggers a full-layer growth. This is illustrated on the left of Figure \ref{fig:board_size_contour_triangular}, where after only 4 moves the board reaches a size of 169 tiles.\\

For the local zone expansion strategy, the most expansive growth is observed when tiles are repeatedly placed toward the bottom-right direction of the grid. This placement pattern causes the creation of up to seven new tiles per move. An example of this behavior is shown on the left of Figure \ref{fig:board_size_zone_triangular}, where the board reaches a size of 41 tiles after 5 moves.\\

In contrast, the best-case scenario is obtained by playing each new move as close as possible to those previously made, in order to encourage a compact and centered growth.\\
Under the perimeter strategy, such compact expansion results in a board of 100 tiles after 49 moves, as shown on the right of Figure \ref{fig:board_size_contour_triangular}, in contrast to 21,904 tiles in the worst-case configuration.\\
For the zone-based approach, the board reaches only 37 tiles after 13 moves, as illustrated on the right of Figure \ref{fig:board_size_zone_triangular}, which would otherwise reach 97 tiles in the worst-case scenario.

\subsubsection{Time random playouts}

Random playouts consist of running complete game simulations where moves are selected uniformly at random from the available legal options at each step. In this study, time random playouts~\cite{ludii-user-guide} involve continuously executing full random playouts within the selected game over a fixed duration of fourty seconds. Three key metrics are recorded during this process: the average number of playouts completed per second (playouts/sec), the average number of moves processed per second (moves/sec), and the total number of playouts performed during the session. These metrics provide a comprehensive overview of the implementation’s efficiency in Ludii and serve as a baseline for comparing the performance of different techniques. Higher values in this metric indicate better performance.

\subsection{Environment}

All experiments were executed on a computing cluster running Rocky Linux 8.6 (Green Obsidian). Jobs were submitted using the Slurm workload manager, each allocated a full node with 8 CPU cores and 600 GB of RAM, and executed in exclusive mode on the batch partition. The experiments were implemented in Java with a heap size set to 512 GB. All jobs were run in exclusive mode to avoid resource contention.

\subsection{Test Datasets}
The experimental evaluation was conducted on a selection of boardless games already implemented in Ludii~\cite{boardless-games-ludii}, chosen to align with the scope and constraints of the proposed techniques. Specifically, the experiments focus on boardless games that satisfy the following criteria: they do not involve stacks, and they are composed of regular tile shapes (i.e., square, hexagonal, or triangular).\\

Based on these criteria, three games were selected: \textbf{Andantino}~\cite{andantino}, \textbf{Bravalath}~\cite{bravalath}, and \textbf{Plotto}~\cite{plotto}. Both Andantino and Bravalath are available in Ludii with support for multiple tile shapes, allowing them to be played using either square, hexagonal or triangular tiles. In contrast, Plotto is only implemented with hexagonal tiles. 




\section{Results}
This section presents the results obtained from the various experiments conducted, based on the metrics selected to compare the different techniques.

\subsection{Size of a board}
\begin{figure}[h!]
    \centering
    \includegraphics[scale=0.4]{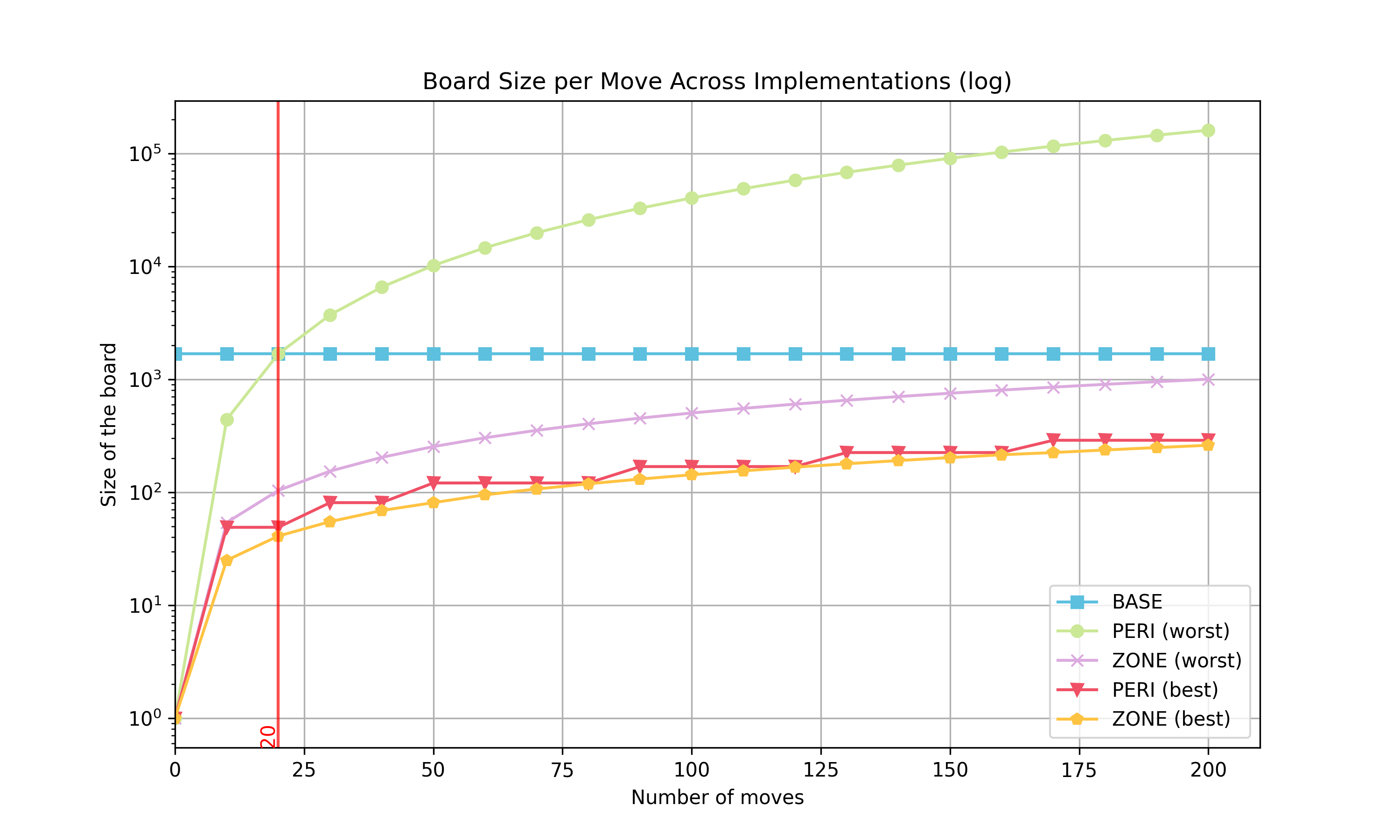}
    \captionsetup{justification=centering}
    \caption{Evolution of the board size across implementations (log) - Square tiles}
    \label{fig:board_size_per_move_square_log}
\end{figure}
\begin{figure}[h!]
    \centering
    \includegraphics[scale=0.4]{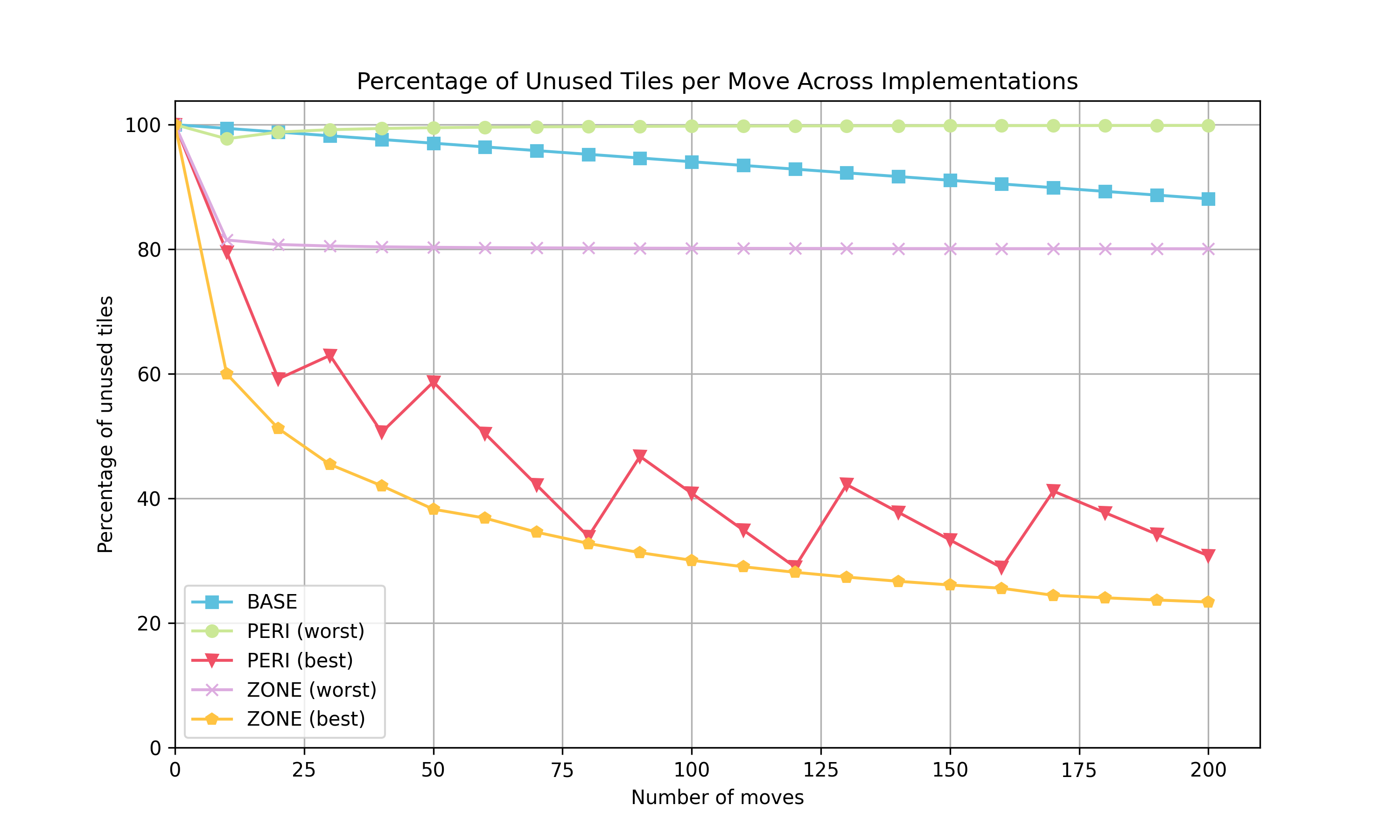}
    \captionsetup{justification=centering}
    \caption{Percentage of the unused tiles of the board - Square tiles}
    \label{fig:pct_per_move_square}
\end{figure}

\begin{figure}[h!]
    \centering
    \includegraphics[scale=0.4]{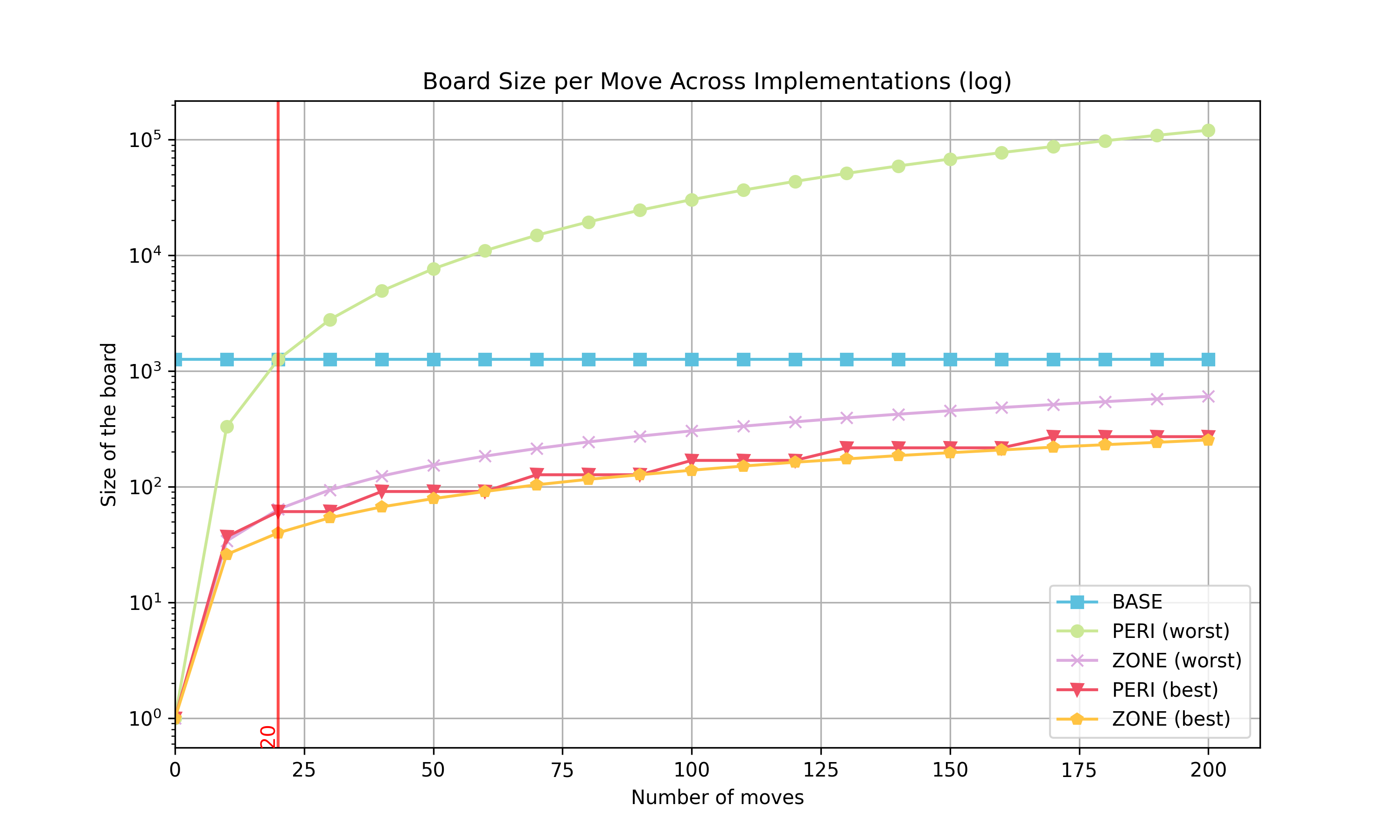}
    \captionsetup{justification=centering}
    \caption{Evolution of the board size across implementations (log) - Hexagonal tiles}
    \label{fig:board_size_per_move_hexagonal_log}
\end{figure}
\begin{figure}[h!]
    \centering
    \includegraphics[scale=0.4]{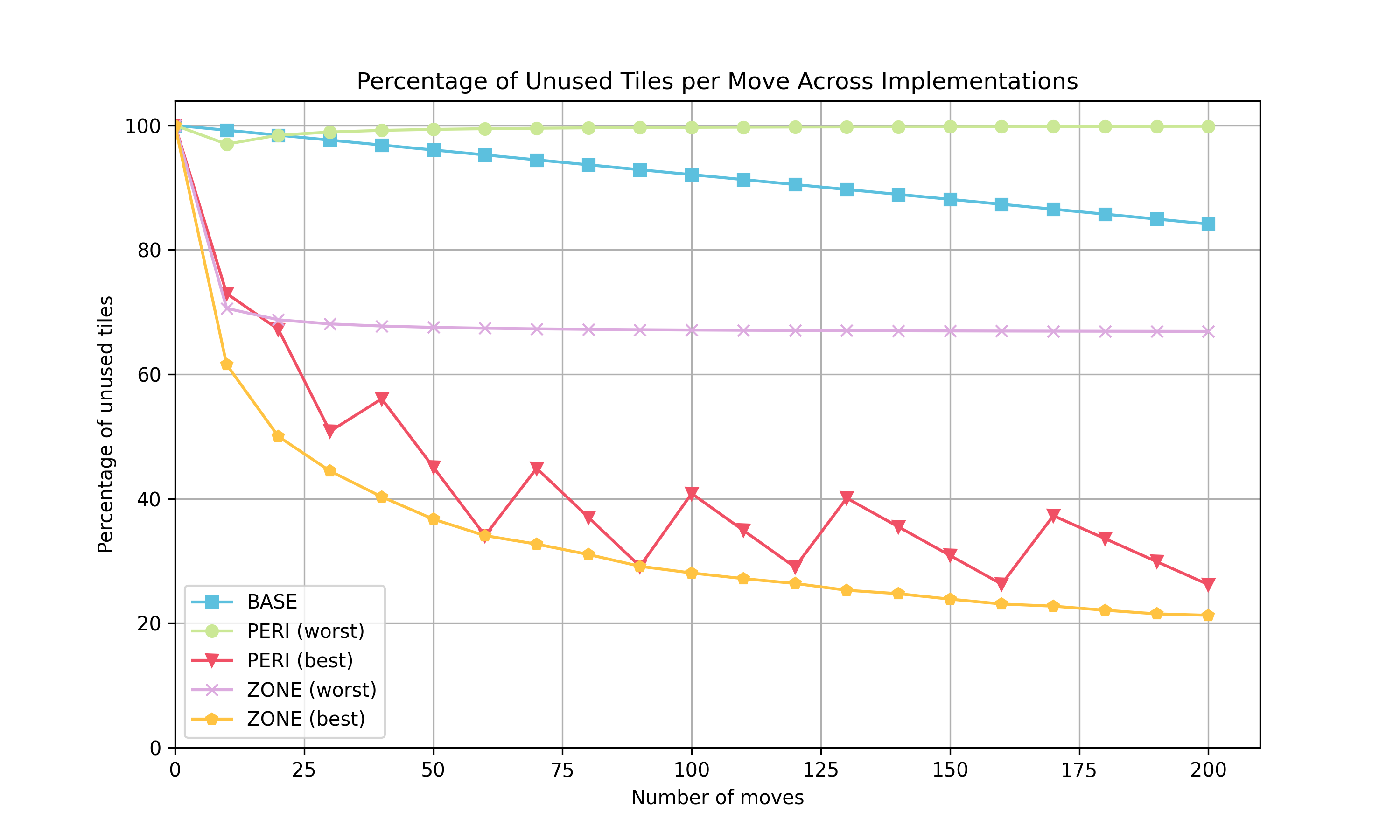}
    \captionsetup{justification=centering}
    \caption{Percentage of the unused tiles of the board - Hexagonal tiles}
    \label{fig:pct_per_move_hexagonal}
\end{figure}

\begin{figure}[h!]
    \centering
    \includegraphics[scale=0.4]{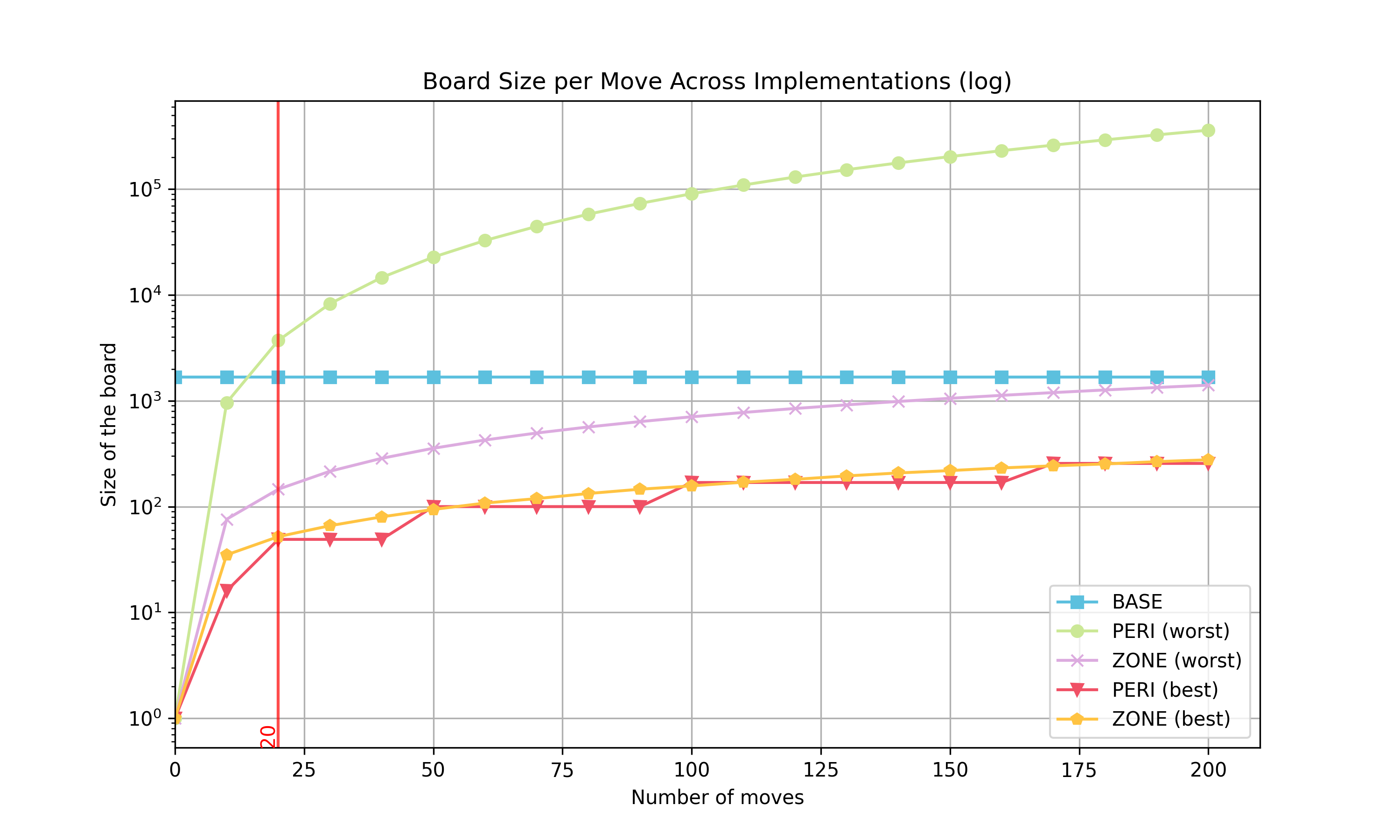}
    \captionsetup{justification=centering}
    \caption{Evolution of the board size across implementations (log) - Triangular tiles}
    \label{fig:board_size_per_move_triangular_log}
\end{figure}

\clearpage

\begin{figure}[h!]
    \centering
    \includegraphics[scale=0.4]{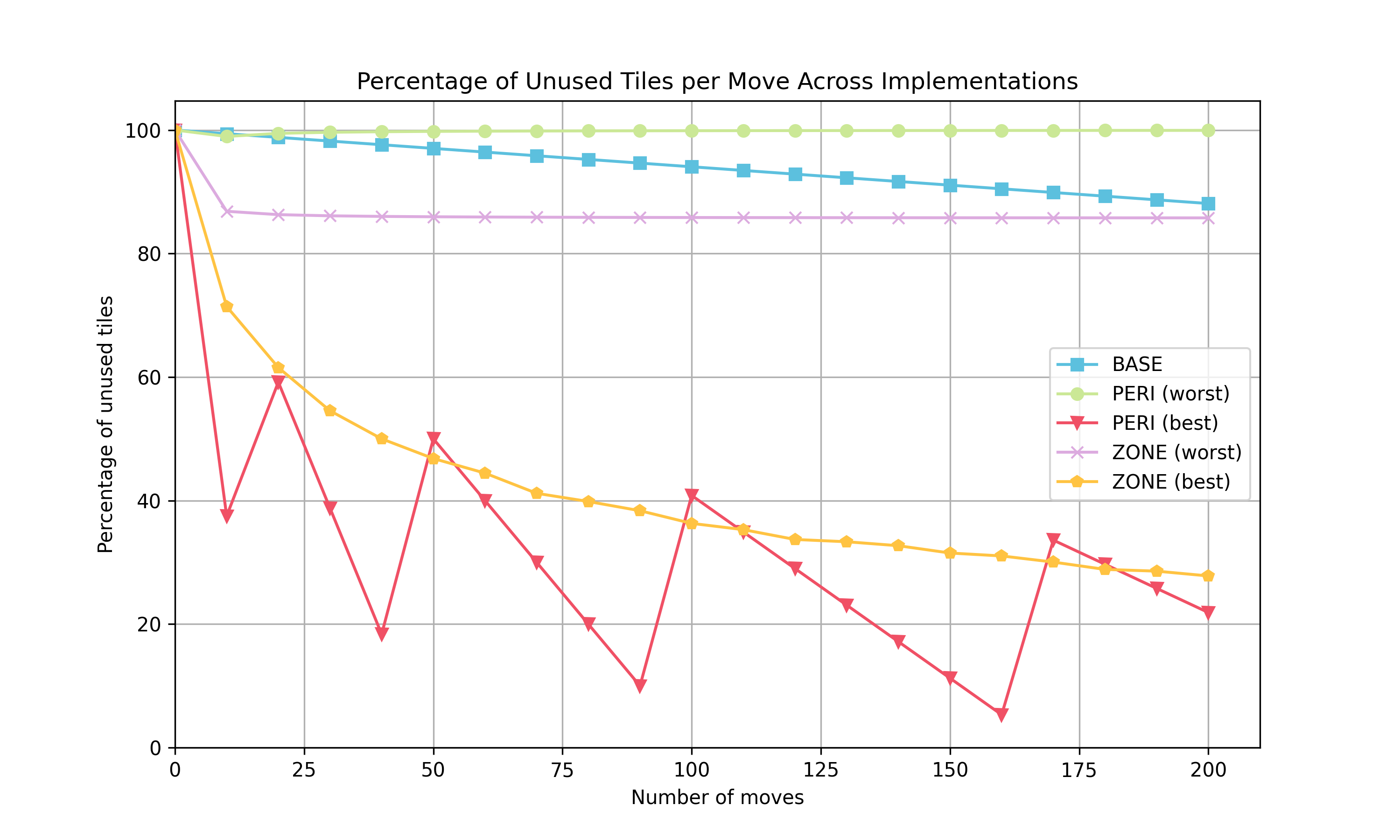}
    \captionsetup{justification=centering}
    \caption{Percentage of the unused tiles of the board - Triangular tiles}
    \label{fig:pct_per_move_triangular}
\end{figure}

\subsection{Random playout}
\begin{table}[ht]
    \centering
    \begin{tabular}{lrrrrrrrr}
        \toprule
        \textbf{Game} & \textbf{BASE} & \textbf{PERI-RE} & \textbf{PERI-MAP} & \textbf{ZONE-RE} & \textbf{ZONE-MAP} \\
        \midrule
        Andantino (square) & 0.33 & \textcolor{blue}{8.16} & 8.12 & 4.10 & 3.17 \\
        Andantino (hexagonal) & 0.30 & \textcolor{blue}{4.63} & \textcolor{blue}{4.63} & 3.65 & 1.99 \\
        Andantino (triangular) & 0.33 & 3.51 & \textcolor{blue}{3.52} & 2.39 & 2.03 \\
        Bravalath (square) & 0.34 & 10.03 & \textcolor{blue}{10.49} & 8.47 & 5.74 \\
        Bravalath (hexagonal) & 0.30 & 9.13 & \textcolor{blue}{9.70} & 8.03 & 6.30 \\
        Bravalath (triangular) & 0.32 & 4.66 & \textcolor{blue}{4.87} & 4.67 & 3.30 \\
        Plotto & 0.28 & 13.09 & \textcolor{blue}{13.70} & 7.76 & 6.63 \\
        \bottomrule
    \end{tabular}
    \captionsetup{justification=centering}
    \caption{Experimental comparison of the number of playouts per second (p/s) between the implementations}
    \label{table:playouts}
\end{table}

\begin{table}[ht]
    \centering
    \begin{tabular}{lrrrrrrrr}
        \toprule
        \textbf{Game} & \textbf{BASE} & \textbf{PERI-RE} & \textbf{PERI-MAP} & \textbf{ZONE-RE} & \textbf{ZONE-MAP} \\
        \midrule
        Andantino (square) & 9.98 & 217.18 & \textcolor{blue}{218.07} & 83.37 & 59.19 \\
        Andantino (hexagonal) & 12.17 & 157.24 & \textcolor{blue}{159.60} & 75.01 & 46.35 \\
        Andantino (triangular) & 7.05 & 95.89 & \textcolor{blue}{96.81} & 56.47 & 45.03 \\
        Bravalath (square) & 4.50 & 120.10 & \textcolor{blue}{122.36} & 101.23 & 68.28 \\
        Bravalath (hexagonal) & 3.17 & 103.44 & \textcolor{blue}{107.66} & 90.46 & 69.64 \\
        Bravalath (triangular) & 3.65 & 56.66 & \textcolor{blue}{58.73} & 58.66 & 44.65 \\
        Plotto & 4.34 & 218.15 & \textcolor{blue}{228.96} & 128.33 & 109.22 \\
        \bottomrule
    \end{tabular}
    \captionsetup{justification=centering}
    \caption{Experimental comparison of the number of moves per second (m/s) between the implementations}
    \label{table:moves}
\end{table}

\begin{table}[ht]
    \centering
    \begin{tabular}{lrrrrrrrr}
        \toprule
        \textbf{Game} & \textbf{BASE} & \textbf{PERI-RE} & \textbf{PERI-MAP} & \textbf{ZONE-RE} & \textbf{ZONE-MAP} \\
        \midrule
        Andantino (square) & 11 & \textcolor{blue}{249} & 245 & 125 & 95 \\
        Andantino (hexagonal) & 10 & 139 & \textcolor{blue}{140} & 110 & 60 \\
        Andantino (triangular) & 11 & \textcolor{blue}{107} & \textcolor{blue}{107} & 72 & 61 \\
        Bravalath (square) & 11 & 301 & \textcolor{blue}{316} & 255 & 173 \\
        Bravalath (hexagonal) & 9 & 275 & \textcolor{blue}{291} & 241 & 190 \\
        Bravalath (triangular) & 10 & 140 & \textcolor{blue}{147} & 141 & 99 \\
        Plotto & 9 & 393 & \textcolor{blue}{411} & 233 & 199 \\
        \bottomrule
    \end{tabular}
    \captionsetup{justification=centering}
    \caption{Experimental comparison of the total number of playouts done between the implementations}
    \label{table:total-playouts}
\end{table}

\section{Discussion}
The following section discusses the interpretation of the results presented in the previous section, by comparing the different techniques and analyzing their respective strengths and weaknesses.

\subsection{Size of a board}
\paragraph{Square}
Figure \ref{fig:board_size_per_move_square_log} presents the growth of the board size in logarithmic scale as a function of the number of moves. The current implementation, corresponding to a static 41×41 grid, maintains a constant board size of 1,681 tiles, regardless of the actual gameplay progression.\\
In contrast, the contour strategy, in the worst-case scenario, significantly increases the board size with each move, leading to exponential tile growth. By the 20th move, the board becomes as large as the one in the current implementation. In the best case, however, it stays consistently smaller throughout the entire sequence of moves.\\
The zone-based implementation expands the board minimally. In both the best- and worst-case scenarios, this strategy never reaches the size of the current implementation, remaining significantly below the 1,681-tile threshold throughout the 200 moves.\\

As shown in Figure \ref{fig:pct_per_move_square}, the current implementation is highly spatially inefficient, maintaining a proportion of unused tiles consistently above 90\%, as the static grid is vastly oversized relative to actual in-game activity. This percentage, however, can only decrease as the game progresses.\\
The contour approach performs even worse in its worst case, with almost 100\% of tiles remaining unused due to its aggressive perimeter-based expansion. In the best-case scenario, the percentage decreases with periodic spikes at regular intervals, corresponding to moves on the perimeter that trigger the generation of many new, initially unused tiles. However, this proportion gradually declines as gameplay continues away from the edges, stabilizing around 30-40\% unused tiles on average.\\
The zone strategy maintains the proportion of unused tiles relatively stable around 80\% in the worst case, while in the best case it demonstrates a consistently declining curve, showing an exponential decrease approaching roughly 30\% unused tiles over time, indicating efficient board usage.

\paragraph{Hexagonal}
Figure \ref{fig:board_size_per_move_hexagonal_log} illustrates the evolution of board size on a logarithmic scale relative to the number of moves for an hexagonal board. The current implementation relies on a fixed hexagon with 21 tiles per side, resulting in a static board of 1,261 tiles that remains unchanged throughout the game.\\
Similar trends are observed with the hexagonal layout as with the square one. The contour strategy once again leads to exponential board growth in the worst case, surpassing the static implementation’s size as early as the 20th move.\\
In contrast, the zone-based approach maintains significantly smaller board sizes in both scenarios, never exceeding the fixed board of 1,261 tiles used in the current implementation.\\

As in the square tile configuration, the percentage of unused tiles for the hexagonal layout reveals significant inefficiencies in the current implementation, with over 80\% of tiles remaining unoccupied throughout the game due to the disproportionate size of the static board, while it will always continue decreasing, and so go under 80\% after a while.\\
The contour strategy exhibits similar behavior: in the worst case, nearly all tiles remain unused; in the best case, the proportion decreases over time, with regular spikes corresponding to edge expansions, and stabilizes around 30-40\%.\\
The zone-based strategy remains the most efficient, with unused tile proportions around 70\% in the worst case and gradually approaching 20\% in the best-case scenario, reflecting improved spatial adaptation compared to the square configuration.

\paragraph{Triangular}
Figure \ref{fig:board_size_per_move_triangular_log} shows the evolution of the board size in logarithmic scale for triangular tiles. The overall trends closely mirror those observed for the square-tile version.\\
The current static implementation, fixed at 41×41 triangles, results in a board of 1,681 tiles, identical in count to its square counterpart, although the triangular layout differs structurally.\\
The contour-based strategy once again demonstrates exponential growth in the worst-case scenario. Notably, due to the geometry of triangular tiling which requires generating tiles in both orientations (with base facing up and down) to preserve a coherent grid, this worst-case expansion surpasses that of the square-tile version. However, the general pattern remains similar to that of the square-tile implementation, with the best-case growth still well below the static implementation throughout the move horizon implementation’s size after approximately 200 moves.\\
The zone-based strategy behaves consistently across both tile types, with a modest and controlled expansion. In the best-case scenarios, the board size remains well below the static threshold throughout all 200 moves. In the worst case, however, it gradually approaches the size of the static implementation.\\

As shown in Figure \ref{fig:pct_per_move_triangular}, the overall trends in tile usage efficiency for triangular tiling are similar to those observed with square tiles. The current implementation remains highly inefficient, with over 90\% of tiles unused throughout the game due to the oversized fixed grid.\\
The contour-based strategy, in its worst case, performs just as poorly (if not worse) maintaining nearly 100\% unused tiles as a result of its expansive perimeter growth. In the best-case scenario, the pattern resembles that of the square grid but with sharper spikes in the unused tile percentage. These spikes are caused by the need to generate nearly twice as many triangles (both upward and downward facing) when expanding the board, leading to oscillations between approximately 10\% and 40\% of unused tiles.\\
The zone-based strategy shows a stable proportion of unused tiles just under 90\% in the worst case. In the best case, the percentage follows a steadily decreasing curve similar to the square version, by converging around 30\%.

\paragraph{Overall} The ZONE strategy is the most effective for all three tile types, as it generates the fewest possible tiles by following the direction in which players tend to play. The PERI strategy can also be an acceptable option, provided it does not fall into a worst-case scenario.

\subsection{Random playout}
In Tables \ref{table:playouts}, \ref{table:moves}, and \ref{table:total-playouts}, the values highlighted in blue represent the highest values for each row. In most cases, the PERI-MAP strategy achieves the best results, with only a few exceptions where the PERI-RE strategy performs slightly better. While PERI-MAP generally demonstrates better performance than PERI-RE, the differences are minimal. Although the replay-based approach (PERI-RE) might be expected to introduce higher execution time (since the board is reset from scratch and all previous moves are re-applied one by one each time the board grows), the results indicate that both implementations produce similar performance levels. In contrast, the difference between the two zone-based strategies, ZONE-MAP and ZONE-RE, is more pronounced.\\

Contrary to expectations, the zone-based approach does not outperform the perimeter-based one. This is mainly due to inefficiencies in the current implementation of zone expansion. The perimeter approach benefits from built-in Ludii functions that enables rapid graph generation by simply specifying the board’s side length and tile type. In contrast, the zone approach cannot rely on predefined structures. Because player actions lead to irregular and unpredictable configurations, new cells must be dynamically identified and integrated into the existing graph. This requires more complex logic to determine where and how to add new cells, vertices, and edges. A redesigned implementation could potentially improve performance, making the zone strategy more competitive, as it is theoretically more localized and efficient.\\

When comparing the best-performing method, PERI-MAP, to the current implementation (BASE), the improvement is significant: on average, PERI-MAP performs approximately 25 times higher than BASE.\\

It is also important to note that, in order to run these experiments, the original method for calculating random playouts had to be revisited. Currently, the topology and graph are not implemented dynamically, only a single static instance exists. As a result, this instance must be reset each time a new instance of the same boardless game is initiated, to not play on the previous board.\\
For example, when comparing the game Andantino with square tiles, the BASE implementation reaches 20,954 playouts when the board is not reset between each playout, compared to only 11 when the board is reset (see Figure \ref{table:total-playouts}). Similarly, in the case of Andantino with hexagonal tiles, 25,307 playouts are reached without resets, versus only 10 with resets.\\
It would be unfair to compare the new methods with this initial setup, since they inherently require the topology and graph to be reset between each playout. For this reason, the current implementation is also reset between each playout in these tests, ensuring a fair comparison across approaches. This help shows that the new implementations outperform the current one.

\chapter{Discussion and Future work}
\label{future_work}


The present work focused on the modeling and implementation of dynamic board games within the Ludii system. Throughout this process, various challenges and limitations were identified, highlighting both the capabilities and current constraints of the platform. This section provides a critical discussion of the results and outlines potential directions for future research and development.

\section{Stack}
One particularly relevant direction for future work concerns the support for stack-based mechanics for boardless games, such as \textit{Dorfromantik}~\cite{dorfromantik} or \textit{Carcassonne}, where players can place pawns or smaller tiles on top of existing tiles. While stacking is already supported independently within the Ludii framework, the new implementation introduced in this study for dynamic board does not yet accommodate boardless games that require stacked components on dynamically added tiles.\\
Extending the new approach for dynamic board generation to support stacked elements requires adapting the way the previous board is mapped to the new one, as the state representation of the board varies between flat and stacked configurations. Unlike flat boards, games with stacks rely on different chunks (see Section \ref{sec:containerstate}), which must be adapted accordingly.\\
Addressing this limitation would significantly broaden the range of boardless games that could be implemented within Ludii, thus improving support for both historical and modern games relying on stacking.

\section{Other shape tile}
Another important direction for future work lies in the generalization of tile shapes used in boardless games. As a result of the work presented in this thesis, it is now possible to design boardless games in Ludii using exclusively one of the following tile types: square, hexagonal, or triangular tiles. However, many modern board games make use of other tile shapes or even combine multiple tile geometries within the same game. For example, in games such as \textit{Keythedral}~\cite{keythedral}, players place tiles that can be either octagonal or square in shape to progressively build the board. Similarly, in games like \textit{Tangram}~\cite{tangram}, players create composite shapes by assembling tiles of various forms (such as squares, parallelograms, and triangles of different sizes) into coherent figures.\\

\begin{figure}[h!]
    \centering
    \includegraphics[width=0.5\linewidth]{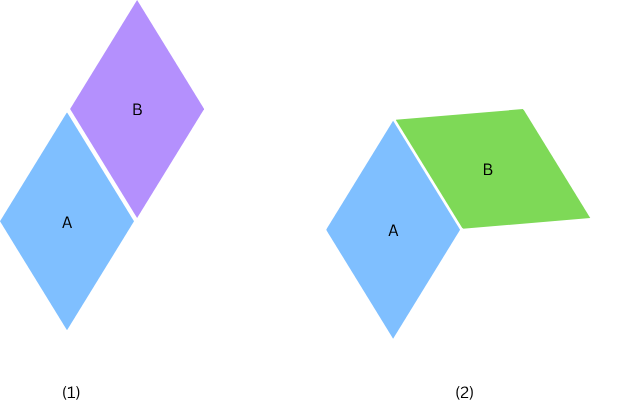}
    \captionsetup{justification=centering}
    \caption{Example of a rhombus placement tile}
    \label{fig:rhombus}
\end{figure}

Introducing multiple tile shapes, or even a single irregular tile shape, within the same game introduces new challenges in predicting the evolution of the board. When different geometries coexist, such as triangular and square tiles, placing a tile on an edge is no longer a straightforward operation. It becomes unclear how the board should expand, as players should be allowed to place either type of tile. The same issue arises in games with unconventional tile shapes, such as rhombus tiles, like the one used in \textit{Lost Valley}~\cite{lost-valley}. As illustrated in Figure\ref{fig:rhombus}, tile B can be placed in two different orientations, each leading to a different board configuration. Addressing this challenge would likely require extending Ludii’s game logic to support multiple possible board evolutions simultaneously.

\section{Rule-based expansion}
In the current implementation, two primary strategies for board expansion have been explored. The first approach consists of adding an entire perimeter around the existing board only when a player places a tile adjacent to the current boundary. The second, more localized method involves adding new board tiles exclusively around the last played position, if the move occurred on an edge.\\

\begin{figure}[h!]
    \centering
    \includegraphics[width=0.5\linewidth]{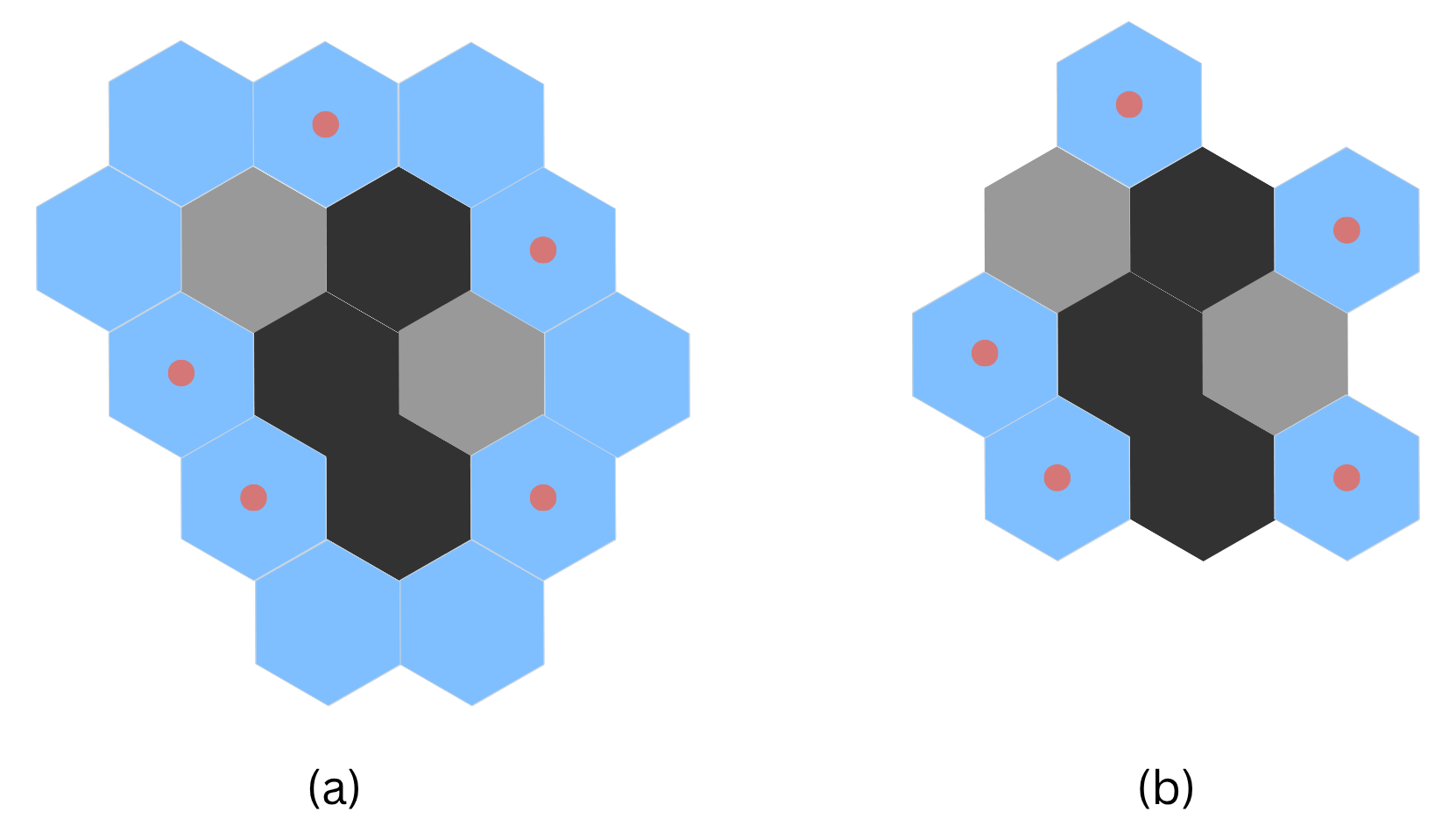}
    \captionsetup{justification=centering}
    \caption{Comparing strategies of board expansion}
    \label{fig:playable}
\end{figure}

A promising direction for future work would be to explore rule-based board expansion. Rather than expanding the board by adding all possible positions adjacent to the location of the most recent move (provided it was played on an edge), the expansion could be restricted to only those positions where a player will legally be able to place a piece in the following turn.\\
For instance, in games such as \textit{Andantino}~\cite{andantino}, a newly placed tile must be adjacent to at least two existing tiles. Applying such rule-based constraints would reduce the number of new positions generated during board expansion, resulting in a more efficient and accurate representation of the evolving board state. As illustrated in Figure\ref{fig:playable}, the board on the left shows the zone expansion strategy (see Section \ref{sec:zone}), where the board (in blue) surround all player tiles. Red dots indicate playable positions, those adjacent to at least two existing tiles. The board on the right demonstrates a rule-based expansion, generating only the necessary board tiles in playable regions, thereby avoiding the creation of irrelevant positions.

\section{Initial size of the board}
Another potential improvement concerns the initialization of the board size. Currently, the initial board dimensions are determined based on the number of tiles present at the start of the game, without considering their actual spatial arrangement. It would be more efficient to also analyze how these tiles are positioned relative to one another in order to determine the minimal required board size.\\
For instance, if nine tiles are initially placed in a 3×3 square configuration, the current implementation generates an 11×11 board, resulting in 121 sites. However, by taking the positions into account, a 5×5 board (i.e., 25 sites) would be sufficient to accommodate the same configuration while still allowing players to place tiles along the edges.\\
This approach would allow the system to generate the smallest possible board necessary for gameplay, minimizing the number of generated tiles and avoiding the creation of regions that may never be used during the game.

\section{Dynamic Topology}
An important enhancement would be to move the topology into the game state. Currently, the game state is dynamic, whereas the topology (and consequently the graph) is static. This means that each action creates a new game state, but the underlying topology remains unchanged throughout the game. However, in the context of evolving board, it would make sense for the topology to become part of the game state, as it may vary over the course of the game.\\

In the new implementation described in this thesis, the topology is overwritten at each move, without any mechanism to preserve its history. Integrating the topology into the game state would enable a more accurate representation of board evolution and allow for features such as undoing moves, analyzing past configurations, or supporting more complex AI reasoning.

\section{Multi-cell Components}
An additional limitation arises in the modeling of multi-cell components, such as those used in domino-based games, which are also boardless games; the board evolves as players successively place dominos. What's more, in the current implementation within Ludii, dominos are not represented as simple two-cell components. Instead, each of the two ends of a domino is modeled as a block of four adjacent cells, resulting in a single domino occupying eight cells in total.\\

This design choice has direct consequences for board expansion strategies. Methods such as adding a one-cell perimeter around the board, or expanding only the immediate surroundings of the latest move, become inadequate in this context. Since placing a new domino require occupying multiple cells, the board must be extended by adding 8 perimeter positions around the existing structure to ensure that all possible placements and orientations are supported.\\

To address this, future work could investigate geometry-aware board expansion, where the board is extended not based on a fixed radius, but in response to the actual spatial requirements of components about to be placed. This would allow for more precise and efficient board management, especially in games involving compound tiles composed of multiple connected sub-tiles. Such an enhancement would help support other boardless games like \textit{King Domino}~\cite{kingdomino}, \textit{King Chocolate}~\cite{king-chocolate}, or \textit{Continuo}~\cite{continuo}.

\section{Reducing Board Size}
Another potential direction for future work concerns the ability to shrink the board by removing unused regions. In the current implementation, when a tile is removed from a location (either by taking the component off the board or moving it elsewhere), the underlying board structure remains unchanged. This can lead to the maintenance of unnecessary board regions. For example, in games such as \textit{Hive}~\cite{hive}, pieces placed on the board can be moved, potentially causing the entire configuration to shift to a different area, leaving previously occupied regions empty and, potentially, irrelevant.\\

A possible extension would therefore involve introducing a mechanism for dynamically shrinking the board by removing isolated cells. Such a feature would improve performance, reduce memory usage, and benefit AI agents by limiting their search space to relevant areas of the board.

\section{One Board Per Player}
Currently, Ludii does not support games where each player has their own individual board that grows dynamically over time. For example, in games like \textit{King Domino}~\cite{kingdomino} or \textit{Ecosystem}~\cite{ecosystem}, each player must develop and expand their own board by adding tiles progressively. Implementing this type of gameplay presents unique challenges, as each player’s board needs to grow independently without interfering with the others’ boards. A promising direction for future work would be to implement a method to dynamically expand each player's board in a way that maintains the separation between players’ spaces while preserving the integrity and rules of the game. Exploring such mechanisms would significantly enhance Ludii's capacity to represent a wider range of boardless games.

\section{Implementation of Modern Boardless Games}
A key objective of this thesis was to enable the implementation of modern board games within Ludii. However, these games introduce a new dimension to the system that has not been fully addressed yet. Boardless games bring the concept that the tiles forming the board must be placed adjacent to other tiles while respecting specific constraints. In the majority of such games, each side of a tile can have different properties. Similar to a puzzle, a piece cannot be placed arbitrarily next to any other piece; rather, placement depends on matching these side-specific properties.\\
Currently, this notion of having distinct properties for each edge is not fully implemented in Ludii. For instance, the game Trax\footnote{https://ludii.games/details.php?keyword=Trax} partially incorporates this idea by relying on the concept of "paths." In the game \textit{Trax}, a tile edge featuring a path must be placed adjacent to another tile edge that also contains a path of the same color. However, in games such as \textit{Dorfromantik}, the concept of paths alone is insufficient. In this game, tiles composed of fields on a side may be placed next to tiles whose edges contain fields, forests or villages, whereas placement directly next a river or railway track is prohibited. Moreover, while some of these elements can be adjacent, it is essential to be able to count how many fields (independently of forests or villages) are in contact with one another. This necessitates a way to group types by their land type (field, forest, villages, ...), treating these as precise properties, while simultaneously specifying which elements can be placed adjacent to each other as another property.\\

The implementations developed in this thesis have opened the first door to supporting boardless games within Ludii. However, in order to accommodate modern boardless games, further developments are still required, particularly to support rule types that go beyond those found in traditional board games.




\chapter{Conclusion}
\label{conclusion}


This thesis focused on improving the representation of boardless tabletop games within the Ludii system by implementing several novel approaches aimed at improving both efficiency and adaptability. Each proposed solution showed clear advantages compared to the current Ludii implementation, addressing key limitations related to memory usage, computational cost, and the dynamic nature of scalable board games.\\

Among these approaches, one in particular, referred to as PERI-MAP in Section \ref{sec:techniques}, demonstrated outstanding performance gains and resource optimization, making it the preferred solution for representing dynamic board configurations at this stage. This method not only enhances the technical management of boardless games but also provides a robust foundation for accurately modeling the scalable expansion of the game space as game progresses.\\

By successfully integrating this advanced board expansion mechanism into Ludii, this work opens new possibilities for supporting a wide range of complex and boardless tabletop games that were previously difficult to implement. It significantly extends Ludii’s capabilities and enhances the potential for AI agents to operate more effectively within boardless game environments.\\

Overall, the contributions made through this thesis represent an important step forward in the field of GGP for boardless games. They offer a scalable and practical solution that bridges the gap between flexible game design and efficient computational representation, thereby enriching the Ludii system and its applicability to future research and development.



\thumbfalse



\bibliographystyle{unsrt}
\bibliography{bibliography/bibliography}



\includepdf[pages=-]{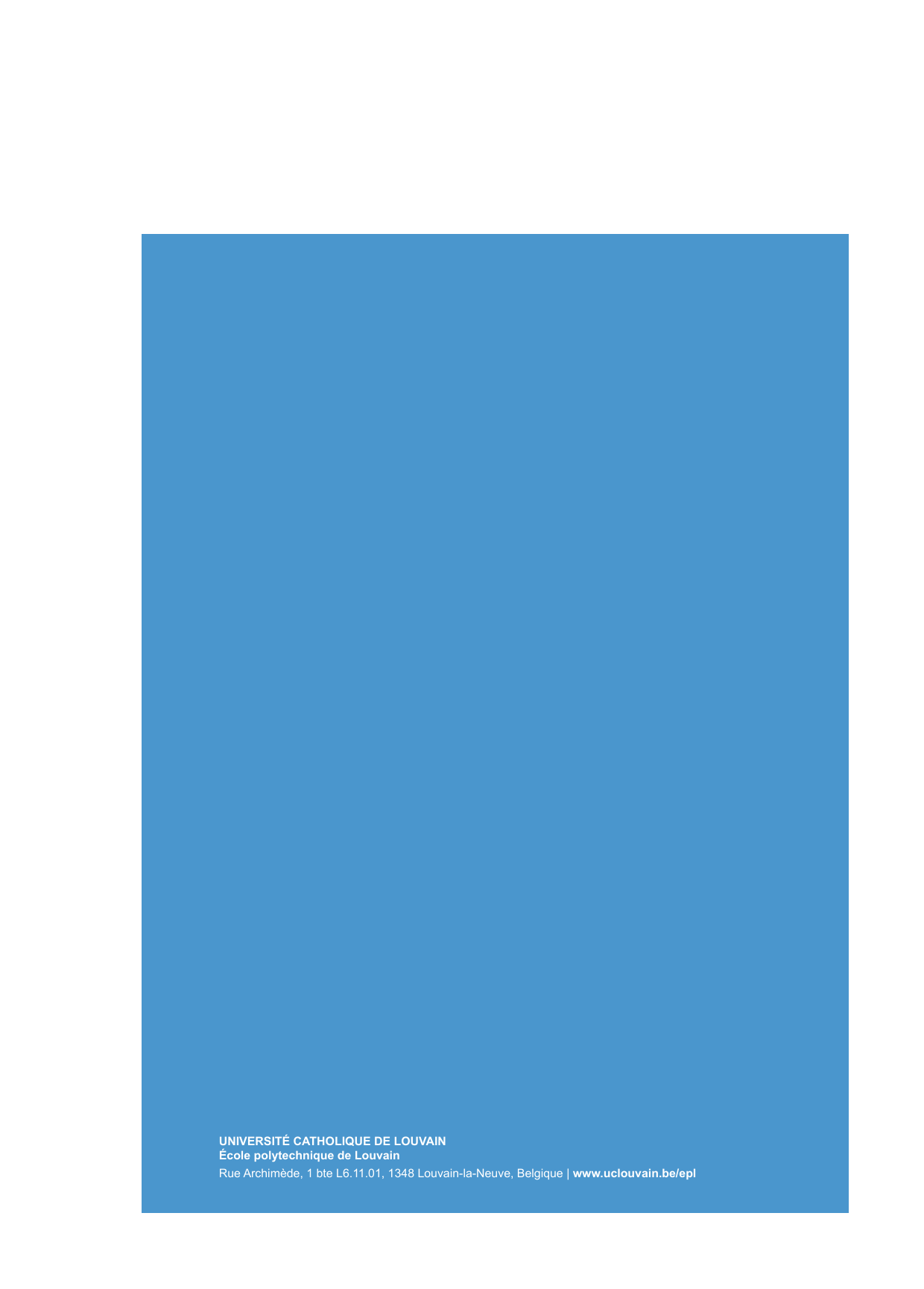}

\end{document}